\documentclass[lettersize,journal]{IEEEtran}
\usepackage{amsmath,amsfonts}
\usepackage{algorithmic}
\usepackage{algorithm}
\usepackage{array}
\usepackage{subfigure}
\usepackage{textcomp}
\usepackage{stfloats}
\usepackage{url}
\usepackage{verbatim}
\usepackage{graphicx}
\usepackage{cite}
\ifCLASSINFOpdf
\else
\usepackage[dvips]{graphicx}
\fi

\graphicspath{{figures/}}
\usepackage{booktabs}
\usepackage{multirow}
\usepackage{makecell}
\usepackage{array}
\usepackage{amssymb}
\usepackage{mathrsfs}
\usepackage{siunitx}
\usepackage{xcolor}
\usepackage{ulem} 
\usepackage{ragged2e}
\usepackage{threeparttable}

\begin{document}

\title{Robust Indoor Localization in Dynamic Environments: A Multi-source Unsupervised Domain Adaptation Framework}

\author{Jiyu Jiao, Xiaojun Wang, Chengpei Han 
\thanks{This work was supported in part by the Jiangsu Provincial Science and Technology Program Project under Grant BG2024003, in part by the Postgraduate Research\&Practice Innovation Program of Jiangsu Province under Grant SJCX24 0061, and in part by the National Key R\&D Program of China under Grant 2022YFC38010000. \textit{(Corresponding author: Xiaojun Wang)}}
\thanks{Jiyu Jiao, Chengpei Han are with National Mobile Communications Research Laboratory, School of Information science and Engineering, Southeast University, Nanjing 214135, China (e-mail: jiyu$\_$jiao@seu.edu.cn, 220230868@seu.edu.cn).}
\thanks{Xiaojun Wang is with National Mobile Communications Research Laboratory, School of Information science and Engineering, Southeast University, Nanjing 211100, China and Purple Mountain Laboratories, Nanjing 211111, China (e-mail: wxj@seu.edu.cn).}
}



\maketitle

\begin{abstract}

Fingerprint localization has gained significant attention due to its cost-effective deployment, low complexity, and high efficacy. However, traditional methods, while effective for static data, often struggle in dynamic environments where data distributions and feature spaces evolve—a common occurrence in real-world scenarios. To address the challenges of robustness and adaptability in fingerprint localization for dynamic indoor environments, this paper proposes DF-Loc, an end-to-end dynamic fingerprint localization system based on multi-source unsupervised domain adaptation (MUDA). DF-Loc leverages historical data from multiple time scales to facilitate knowledge transfer in specific feature spaces, thereby enhancing generalization capabilities in the target domain and reducing reliance on labeled data. Specifically, the system incorporates a Quality Control (QC) module for CSI data preprocessing and employs image processing techniques for CSI fingerprint feature reconstruction. Additionally, a multi-scale attention-based feature fusion backbone network is designed to extract multi-level transferable fingerprint features. Finally, a dual-stage alignment model aligns the distributions of multiple source-target domain pairs, improving regression characteristics in the target domain. Extensive experiments conducted in office and classroom environments demonstrate that DF-Loc outperforms comparative methods in terms of both localization accuracy and robustness. With 60\% of reference points used for training, DF-Loc achieves average localization errors of 0.79m and 3.72m in "same-test" scenarios, and 0.94m and 4.39m in "different-test" scenarios, respectively. This work pioneers an end-to-end multi-source transfer learning approach for fingerprint localization, providing valuable insights for future research in dynamic environments.

\end{abstract}


\section{Introduction}
\IEEEPARstart{W}{ith} the rapid advancement of 5G technology, precise and reliable location-based services have become increasingly crucial in various applications such as indoor navigation, asset tracking, and smart city management \cite{lu2024tutorial}. Traditional wireless communication-based positioning techniques, including geometric localization methods like Time of Arrival (TOA), Angle of Arrival (AOA), and hybrid approaches, have been extensively studied. Although these methods offer advantages such as explicit mathematical models and high accuracy under line-of-sight (LOS) conditions, their performance deteriorates significantly in non-line-of-sight (NLOS) scenarios and complex indoor environments due to multipath propagation and signal attenuation. Furthermore, these techniques often rely on accurate base station positioning and sophisticated hardware, which elevates deployment costs and complexity.

Alternatively, fingerprint-based localization has emerged as a promising solution, particularly in indoor settings, owing to its robustness in NLOS conditions, minimal hardware requirements, and ability to utilize existing wireless infrastructure. This approach typically involves constructing an offline database of signal characteristics—such as Received Signal Strength (RSS) or Channel State Information (CSI)—and matching online measurements to this database for position estimation. Researchers have explored various fingerprinting methods, including deterministic techniques that employ similarity measures for matching \cite{9999279}, and probabilistic models that estimate positions based on signal distributions \cite{ruan2022hi}. Recent advancements in machine learning (ML) and deep learning (DL) have further enhanced the accuracy and adaptability of fingerprint-based localization systems.

However, traditional fingerprint-based localization methods often exhibit limitations in dynamic environments, where signal fluctuations, multipath effects, and device heterogeneity can lead to diminished accuracy and reliability due to evolving signal distributions and feature spaces. Additionally, these methods often require large volumes of labeled training data, making them labor-intensive and costly to implement in large-scale or rapidly changing settings. While some studies have attempted to mitigate these challenges through data calibration to maintain an updated fingerprint database, this strategy is often impractical \cite{ji2022generating, khatab2021fingerprint}. Therefore, investigating an efficient and robust fingerprint localization method capable of performing effectively in dynamic environments holds significant theoretical and practical importance.

In recent years, indoor localization technologies—particularly fingerprinting methods designed for dynamic environments—have made significant strides. Researchers have extensively explored various technical approaches to enhance localization accuracy and adaptability\cite{khatab2021fingerprint,guo2018accurate, 9078108,8115100,8661625,9451560,9686643,khatab2021fingerprint,s20236994}. Dynamic fingerprinting techniques utilize DL and probabilistic models to optimize fingerprint matching performance\cite{khatab2021fingerprint,guo2018accurate,9078108,9451560,8115100}. For example, high-level features are extracted using autoencoders and deep extreme learning machines, and localization robustness is enhanced by integrating predictions from multiple classifiers through an Extended Candidate Location Set (ECLS)\cite{guo2018accurate}. Additionally, some studies have combined the amplitude and phase information of Channel State Information (CSI), achieving higher precision via probabilistic regression and Angle of Arrival (AoA) analysis, while effectively reducing signal labeling errors using multimodal crowdsourcing methods\cite{9078108}. However, these approaches still face challenges such as strong dependence on labeled data, high computational complexity, and limited scalability in large-scale scenarios.

To address variations in signal distributions within dynamic environments, transfer learning (TL) methods have introduced cross-domain adaptation techniques\cite{s20236994,8661625,9686643}. Examples include reducing inter-domain distribution discrepancies through Optimal Transport (OT) and preserving discriminative characteristics using Global and Local Structure Consistency Constraints (GLOSS). Further research has proposed knowledge transfer frameworks capable of adapting to short-term environmental changes and long-term feature space heterogeneity. Although TL methods effectively alleviate the need for recalibration, issues related to dependency on source-target domain relationships and high computational costs still require optimization. Therefore, despite progress in enhancing adaptability to dynamic environments, existing studies continue to encounter challenges regarding data dependency, computational complexity, and insufficient scalability.

In recent years, unsupervised domain adaptation (UDA) algorithms have been actively investigated, yet most algorithms and theoretical results have focused on single-source unsupervised domain adaptation (SUDA)\cite{long2013transfer,5640675,zhang2017joint,1011453360309}. Multi-source unsupervised domain adaptation (MUDA)\cite{xu2018deep,zhao2018multiple} offer significant advantages in addressing cross-domain transfer problems, particularly when the target domain lacks labeled data. Compared to traditional SUDA methods, MUDA can extract more comprehensive knowledge from multiple source domains, thereby mitigating the inherent biases present in single-source data. However, in practical scenarios, labeled data collected from multiple diverse sources may not only differ from the target domain but also from each other. Consequently, domain adapters from multiple sources should not be modeled identically. 

Building upon this background, we propose \textbf{DF-Loc}, an indoor \textbf{f}ingerprinting \textbf{loc}alization system based on MUDA tailored for \textbf{d}ynamic environments. To efficiently integrate information from multiple sources and enhance generalization in the target domain, DF-Loc aligns each pair of source and target domains through specific distribution alignment and regressor output alignment. This approach not only addresses the challenge of extracting domain-invariant representations within a common feature space but also effectively reduces the risk of erroneous predictions on target samples. Specifically, the system learns domain-invariant features in distinct feature spaces and aligns target sample predictions by leveraging the decision boundaries of regressors, thereby improving cross-domain regression performance. This provides a more robust solution for knowledge transfer in complex cross-domain scenarios within fingerprint localization tasks.

Given the widespread availability and detectability of indoor Wi-Fi Access Points (APs), the proposed method utilizes Channel State Information (CSI) extracted from downlink data frames of commercial Wi-Fi networks. CSI provides fine-grained wireless channel characteristics essential for precise fingerprint positioning and can be extended to 5G New Radio (NR) and other communication scenarios owing to similar MIMO-OFDM physical layer structures. This user-centric approach, implemented on the User Equipment (UE), reduces network overhead and safeguards user privacy. Additionally, integrating supplementary sensors embedded within mobile devices can further enhance positioning accuracy and reliability. From a mobile communications standpoint, this method aligns with prevailing trends in edge computing and user-centric network architectures. By leveraging existing communication infrastructure without necessitating additional hardware, it offers cost-effectiveness and scalability while achieving superior localization performance compatible with future communication standards.

In summary, the main contributions are as follows:
\begin{itemize}
\item DF-Loc addresses the challenge of data distribution discrepancies across different time scales. By employing DL techniques, it simultaneously extracts domain-invariant features from multiple source-target domain pairs, thereby enhancing generalization in the target domain while reducing the need for labeled data. To the best of our knowledge, this aspect has not been explored in previous localization research.

\item We propose a dual-stage alignment model that aligns the statistical distributions of source and target domains in multiple specific feature spaces and aligns regressor outputs using domain-specific decision boundaries. This approach facilitates knowledge transfer from multiple sources.

\item Furthermore, we design a feature learning backbone network that integrates multi-scale convolution, multi-scale channel attention Mechanism (MS-CAM) and attention-based feature fusion (AFF), enhancing the extraction of transferable features in both source and target domains. Additionally, a QC module for fingerprint data preprocessing and a CSI fingerprint reconstruction module are incorporated. Together, they constitute a comprehensive solution for dynamic indoor localization in wireless communication environments.

\item The effectiveness and robustness of DF-Loc are validated through typical indoor field experiments conducted in office and classroom environments.  We also discuss the factors and challenges that influence the localization performance of the system.
\end{itemize}

The remainder of this paper is structured as follows. Section II provides a review of significant related work. In Section III, we formally define the dynamic localization problem. A detailed description of the DF-Loc algorithm is presented in Section IV. Section V reports on the experiments and analyzes the comparative results. In Section VI, we further discuss the implications and challenges associated with DF-Loc. Finally, conclusions are drawn in Section VII.

\section{Related work}

Indoor fingerprint localization has become a research hotspot in recent years. However, due to the complex and dynamic nature of indoor environments, wireless signals are susceptible to multipath effects, environmental changes, and pedestrian movement, which can degrade localization accuracy. To enhance the robustness and accuracy of indoor fingerprint localization, researchers have proposed various methods to address the challenges posed by dynamic localization, primarily focusing on the following two aspects:

\subsection{Adaptive Fingerprint-based Indoor Localization}
Crowdsourcing, a method leveraging collective intelligence for problem-solving, has been widely adopted in indoor localization \cite{ji2022generating}. For instance, Wei et al. \cite{wei2021efficient} proposed an efficient crowdsourcing approach for fingerprint collection. In this approach, participants collect RSS data using smartphones while traversing a path, and Gaussian processes are employed for localization.

Sensor-aided indoor localization techniques leverage data collected by various sensors on mobile devices, such as Bluetooth Low Energy (BLE), Quick Response (QR) codes, and Micro-Electro-Mechanical System (MEMS) sensors \cite{yu2023intelligent}. Santos et al. \cite{santos2021crowdsourcing} employed a multimodal approach to construct highly accurate fingerprints by integrating inertial data, local magnetic fields, and Wi-Fi signals.

Data augmentation and weakly supervised learning techniques have been increasingly applied to address challenges in indoor localization, such as data sparsity and environmental dynamics \cite{njima2022dnn,khatab2021fingerprint,yan2020elm,ye2022se,kim2021unsupervised,wang2022adversarial}. These techniques aim to enhance localization accuracy through supervised learning while maintaining low online complexity.  Njima et al. \cite{njima2022dnn} addressed the issue of unlabeled and missing data by proposing a weighted semi-supervised DNN and Generative Adversarial Network (GAN)-based indoor localization method. This method improves localization accuracy and reduces data collection costs. Khatab et al. \cite{khatab2021fingerprint} introduced a novel fingerprint-based indoor localization algorithm that utilizes an autoencoder for high-level feature extraction and combines crowdsourced labeled and unlabeled data to enhance localization performance. Yan et al. \cite{yan2020elm} proposed a new indoor localization technique based on Extreme Learning Machine (ELM) and Iterative Self-Organizing Data Analysis Algorithm (ISODATA). This technique performs classification learning and feature extraction on RSSI measurements and employs semi-supervised regression learning to obtain a location regression function. Considering the security vulnerabilities of existing methods and their susceptibility to malicious attacks, Ye et al. \cite{ye2022se} proposed SE-LOC, a technique based on semi-supervised learning designed to enhance the security and resilience of fingerprint-based localization systems.  Furthermore, unsupervised learning methods, which eliminate the need for manual data labeling, have also found applications in localization. Kim et al. \cite{kim2021unsupervised} explored the use of Unsupervised View Selection Deep Learning (UVSDL) to improve the accuracy of indoor localization systems in complex environments, particularly in NLOS scenarios. Wang et al. \cite{wang2022adversarial} proposed an Adversarial Deep Learning (ADVLOC) method for indoor localization systems to enhance their resilience against adversarial attacks.

However, the aforementioned methods either assume feature homogeneity or require all calibration data to be provided in advance, thus failing to address the challenges associated with varying time scales.

\begin{table}[!t]
	\centering
	\caption{List of Notations.} 
	\label{List of Notations}
	\renewcommand{\arraystretch}{1.2} 
	\begin{tabular}{p{1.5cm} p{6.5cm}} 
		\toprule
		\textbf{Notation}   & \textbf{Definition} \\
		\midrule
		
		$\mathcal{D}_S^{\left( i \right)}$ & the source domain in time period $ t_i $  \\ 
		$\mathbf{X}_{S}$ & the source-domain data \\
		$\boldsymbol{L}_{S}$ & source labels \\
		$\mathcal{N}_S$ & The number of source samples \\
		$d_S$ & source dimensions \\
		\midrule
		
		$\mathcal{D}_{T}^{\left( j \right)}$ & the labeled target domain in time period $ t_j $ \\
		$\mathbf{X}_{T}^{\left( j \right)}$ & the target-domain data collected in time period $ t_j $\\
		$\boldsymbol{L}_{T}^{\left( j \right)}$ & target labels collected in $ t_j $\\
		$\mathcal{N}_{T}^{\left( j \right)}$ & The number of target samples in $ t_j $\\
		$d_{T}^{\left( j \right)}$ & source dimensions \\
		$\boldsymbol{A}$ & mapping function \\
		
		\midrule
		
		$\mathcal{N}$ & $\mathcal{N}=\mathcal{N}_S+\mathcal{N}_T$, the total number of samples in source and target domains \\
		$g_{r}$ & The number of training reference points \\
		$g_t$ & The number of test reference points \\
		$G$ & $ G=g_r+g_t $, the total number of reference points \\
		\bottomrule
	\end{tabular}
\end{table}

\subsection{Knowledge Transfer for Indoor Localization}
Existing indoor localization methods, whether based on purely statistical signal processing or data-driven approaches, often struggle to generalize to new environments. This limitation leads to significant time and effort wasted in recalibration and retraining \cite{xiang2022crowdsourcing,kerdjidj2024exploiting,chen2022fidora,prasad2023domain,gao2023metaloc}.

To address the performance degradation caused by changes in the propagation environment, Xiang et al. \cite{xiang2022crowdsourcing} proposed a low-complexity self-calibrating indoor crowdsourcing localization system based on multi-kernel TL.  Kerdjidj et al. \cite{kerdjidj2024exploiting} presented a TL-supported classification system that transforms one-dimensional signals into images and utilizes techniques such as spectrograms, scalograms, or Gramian Angular Fields. Chen et al. \cite{chen2022fidora} introduced FIDORA, a WiFi localization system based on domain adaptation and clustering assumptions, to address the limitations of existing WiFi fingerprinting systems when faced with varying user body types and environmental changes. Prasad et al. \cite{prasad2023domain} focused on the domain shift between offline and online RSS fingerprints caused by environmental changes, device heterogeneity, and AP variations. They proposed a novel Domain Adversarial Regression Neural Network (DANN-R) that uses an autoencoder for dimensionality reduction and employs a Gradient Reversal Layer (GRL) for adversarial learning to mitigate challenges in dynamic IoT environments.To improve the utilization of historical task data and the adaptation speed of the model to new environments, Gao et al. \cite{gao2023metaloc} proposed the MetalOC framework, which combines Model-Agnostic Meta-Learning (MAML) with DNNs. This framework utilizes historical data from well-calibrated environments for training and employs a bi-level optimization mechanism to obtain meta-parameters.

However, most of these approaches only consider knowledge transfer between a single source and target domain pair, which cannot guarantee robustness in continuously dynamic environments. Moreover, they require labels for online samples in the current stage, which are not readily available in real-time localization applications.  To address these limitations, we propose DF-Loc, an end-to-end framework that tackles both feature distribution discrepancies across multiple time scales and the cost of labeled samples.

\begin{figure*}[!t]
	\centering
	\includegraphics[width=6.5in]{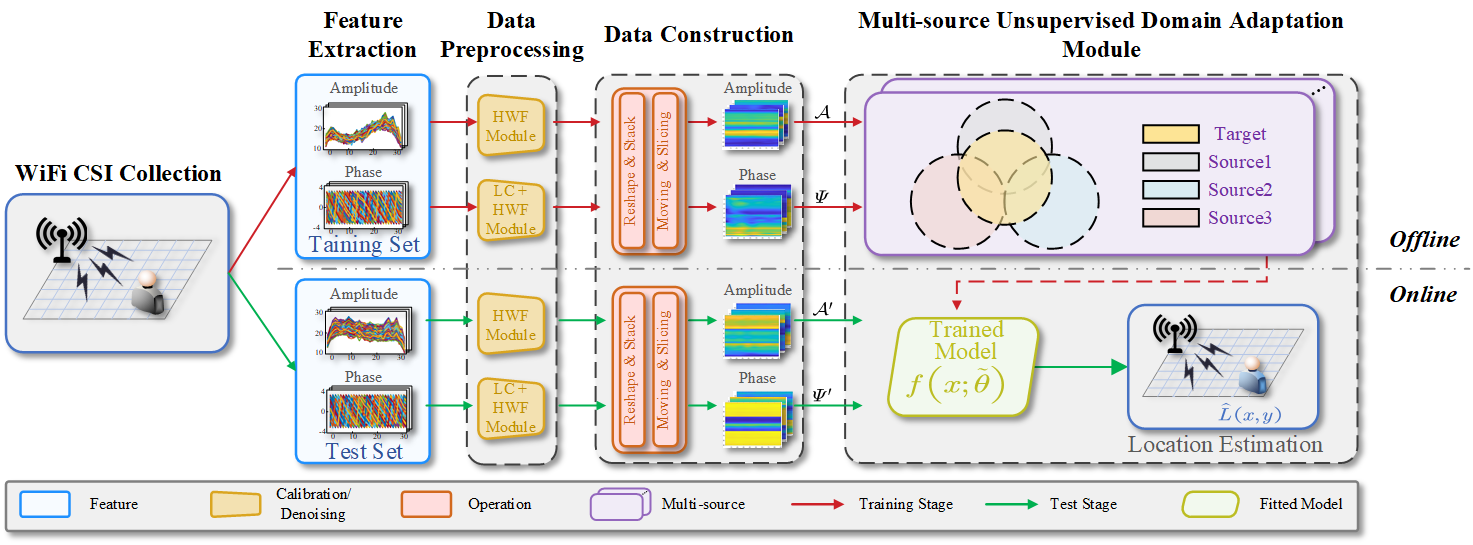}
	\caption{Architecture of DF-Loc.}
	\label{fig_1}
\end{figure*}

\begin{figure*}[!t]
	\centering
	\subfigure[ ]{\includegraphics[width=1.7in]{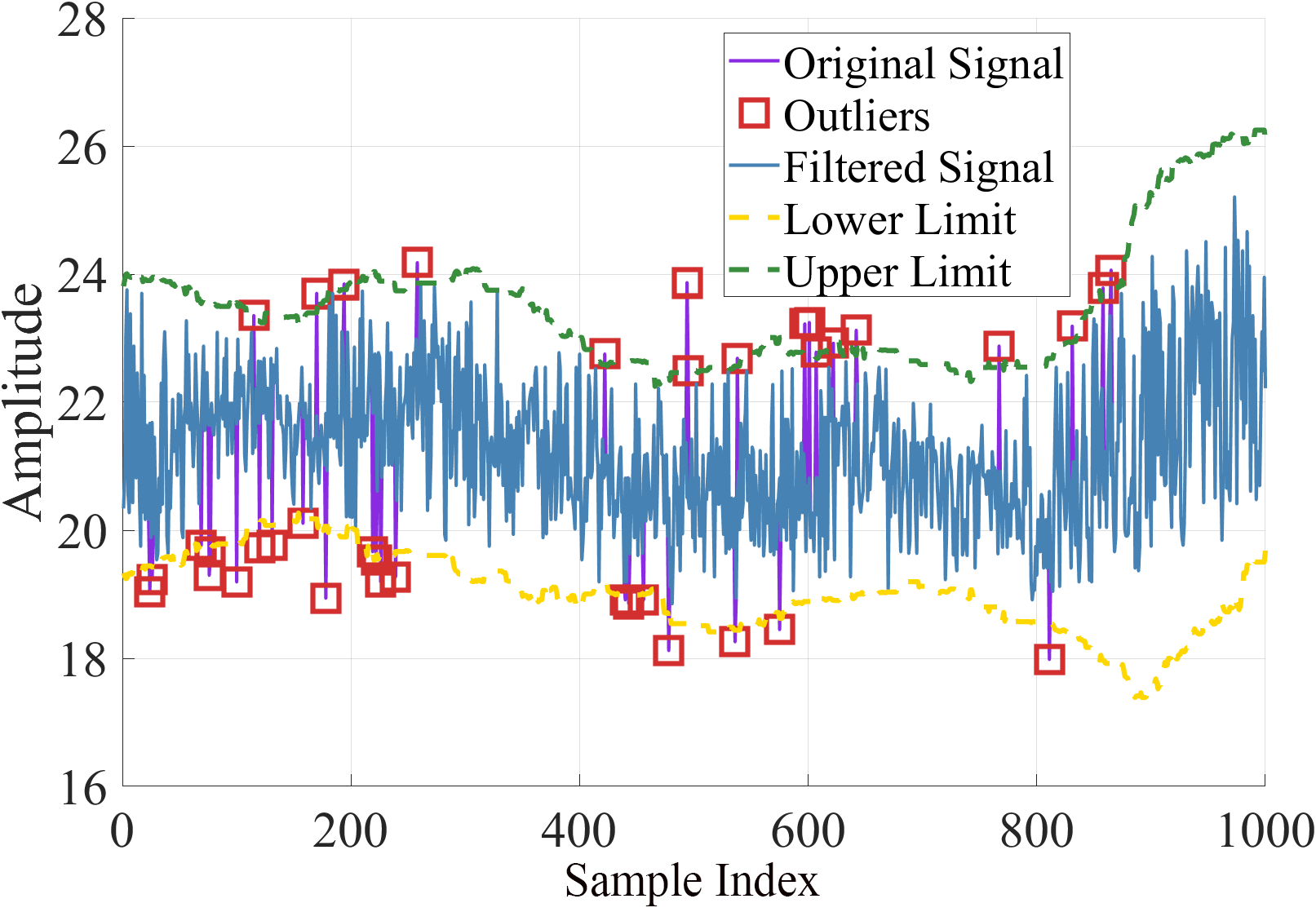}%
		\label{fig1_first_case}}
	\hfil
	\subfigure[ ]{\includegraphics[width=1.5in]{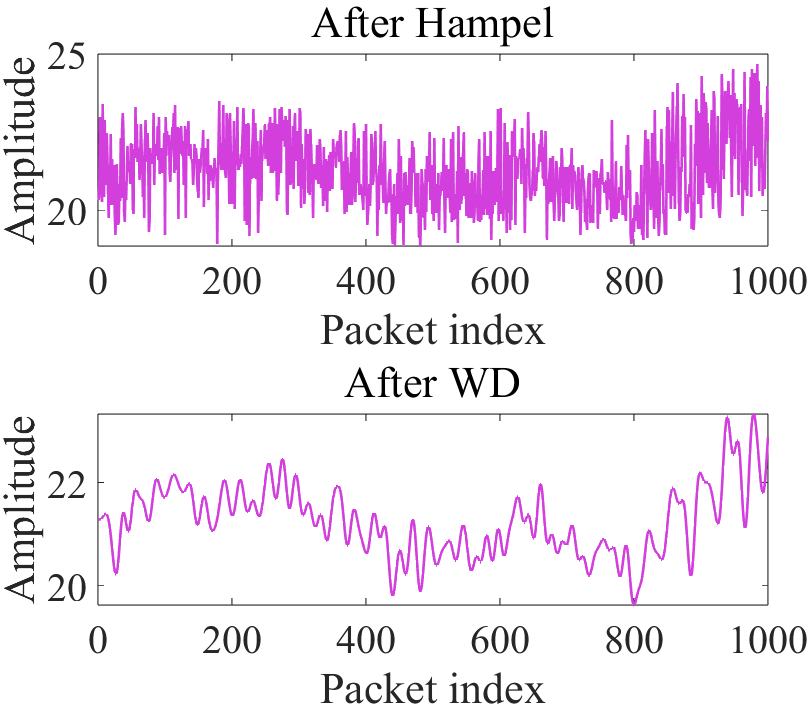}%
		\label{fig1_second_case}}
	\hfil
	\subfigure[ ]{\includegraphics[width=1.6in]{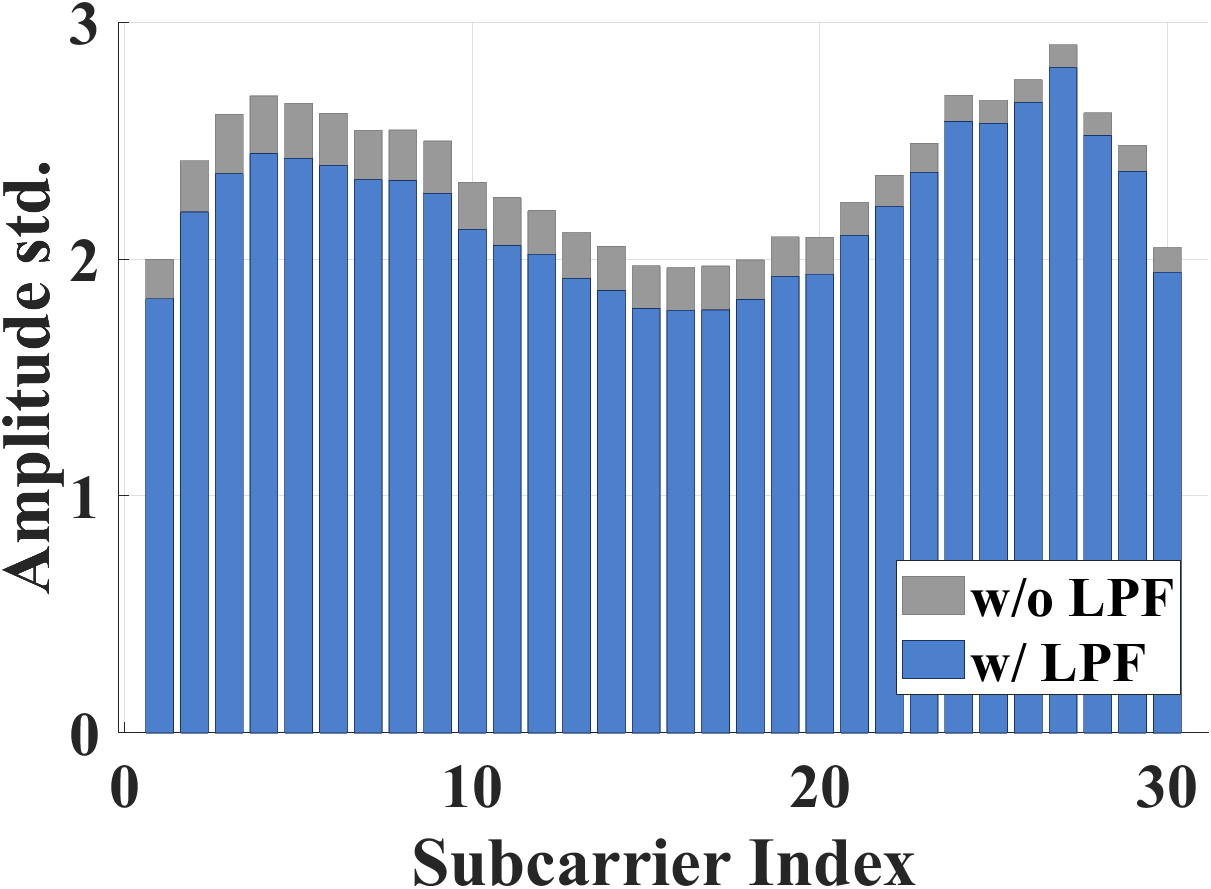}%
		\label{fig1_third_case}}
	\caption{QC-based preprocessing for amplitude. (a) Hampel identifier for coarse outliers removing. (b) Wavelet filter for smoothing sequence. (c) Butterworth low-pass filter implied outliers removing.}
	\label{fig_4}
\end{figure*}

\section{PROBLEM FORMULATION}
A mobile node collects signal measurements from surrounding Access Points (APs). We assume that there are several time periods $ \left\{ t_i \right\} $ for $ i=1,2,\cdots ,N $, during which human activities are classified into different types $ \left\{ f_i \right\} $ for $ i=0,1,\cdots ,N-1 $. Furthermore, the type of human activity remains constant within a given time period, such that the collected samples exhibit the same feature space. However, samples collected across different time periods may belong to different feature spaces, corresponding to distinct domains.

In time period $ t_1 $, we assume that the environment is divided into $ G $ grid points $ \boldsymbol{L}=\left\{ L_g=\left( x_g,y_g \right) \mid g=1,\cdots ,G \right\} $, referred to as reference points (RPs), where $ \mathcal{P} $ represents a set of Cartesian coordinates. Initially, we construct an offline fingerprint database by collecting $ K $ CSI samples from each RP and labeling them, forming a total of $ n_S $ signal-position pairs, denoted as $ \mathcal{D}_S=\left\{ \mathbf{X}_S,\boldsymbol{L}_S \right\} =\left\{ \left( csi_{S,g}^{\left( k \right)},L_{S,g}^{\left( k \right)} \right) \right\} $, where $g\in \left[ 1,...,G \right] $ and $ k\in \left[ 1,...,K \right]  $. Here, $ \mathcal{N}_S=G\times K$, $csi_{S,g}^{\left( k \right)}\in \mathbb{R}^{d_S} $ represents the $ k $-th sample with $ d_S $-dimensional features collected at the $ g $-th RP, and $ L_{S,g}^{\left( k \right)} $ is its corresponding label. In the online localization phase, we developed a multi-scale attention-based feature fusion localization model $ f_{T}^{1}\left( \cdot \right) $ to implement online localization. During the time period $ t_i $ (where $ i > 1$), we apply the same methodology to construct additional source domain fingerprint databases denoted as $ \mathscr{D}_S=\left\{ \mathcal{D}_{s_i} \right\} _{i=1}^{N_S}$, where $ N_S $ represents the number of source domains.

During the time period $ t_j $ (where $ j > i$), some received signals exhibit feature distributions different from those at the preceding time $ t_{j-1} $, and some even display heterogeneity with differing feature dimensions. This disparity renders the previous regressor $ f_{T}^{\left( j-1 \right)}\left( \cdot \right) $ obsolete. To address this issue, we propose a MUDA algorithm. We consider the calibration data as the target domain, comprising a set of labeled samples collected during online localization or shortly beforehand, denoted as $ \mathcal{D}_{T}^{\left( j \right)}=\left\{ \mathbf{X}_{T}^{\left( j \right)},\boldsymbol{L}_{T}^{\left( j \right)} \right\} =\left\{ \left( csi_{T,g}^{\left( k \right) ,\left( j \right)},L_{T,g}^{\left( k \right) ,\left( j \right)} \right) \right\} $, where $  \mathcal{N}_T^{\left( j \right)}=G^{\left( j \right)}\times K^{\left( j \right)} $, $g\in \left[ 1,...,G^{\left( j \right)} \right] $ and $k\in \left[ 1,...,K^{\left( j \right)} \right] $. Here, $ csi_{T,g}^{\left( k \right) ,\left( j \right)}\in \mathbb{R}^{d_{T}^{\left( j \right)}} $ represents the $ k $-th sample at the $ g $-th RP in the $ j $-th time period, and $ d_{T}^{\left( j \right)} $ denotes the feature dimension in the target domain. In the offline calibration phase, we aim to learn a mapping matrix $ \boldsymbol{A}\in \mathbb{R}^{d'} $ to project the data from both source and target domains into a $ {d'} $-dimensional latent feature subspace, where $ {d'}\ll \min \left( d_S,d_{T}^{\left( j \right)} \right) $. In practical scenarios, labeled data may differ not only from the target domain but also among themselves. Therefore, domain adapters from various sources should not be modeled identically. To address this issue, we introduce a MUDA algorithm that learns distinct mapping matrices $ \left\{ \boldsymbol{A}_{i}^{\left( j \right)} \right\} _{i=1}^{N_S} $ for each source, where $ N_S $ denotes the number of source domains. To further learn an effective domain-invariant space, we employ a dual-stage alignment method, aligning each pair of source and target domains through specific distribution alignment and regressor output alignment. Similarly, in the online localization phase, when presented with online samples $ \mathbf{X}_{O}^{\left( j \right)}=\left\{ csi_{t}^{\left( j \right) ,O} \right\} $, we update the weights of the model $ f_{T}^{\left( j \right)}\left( \cdot \right) $. Table \ref{List of Notations} summarizes the key notations.

\begin{figure*}[!t]
	\centering
	\subfigure[ ]{\includegraphics[width=1.7in]{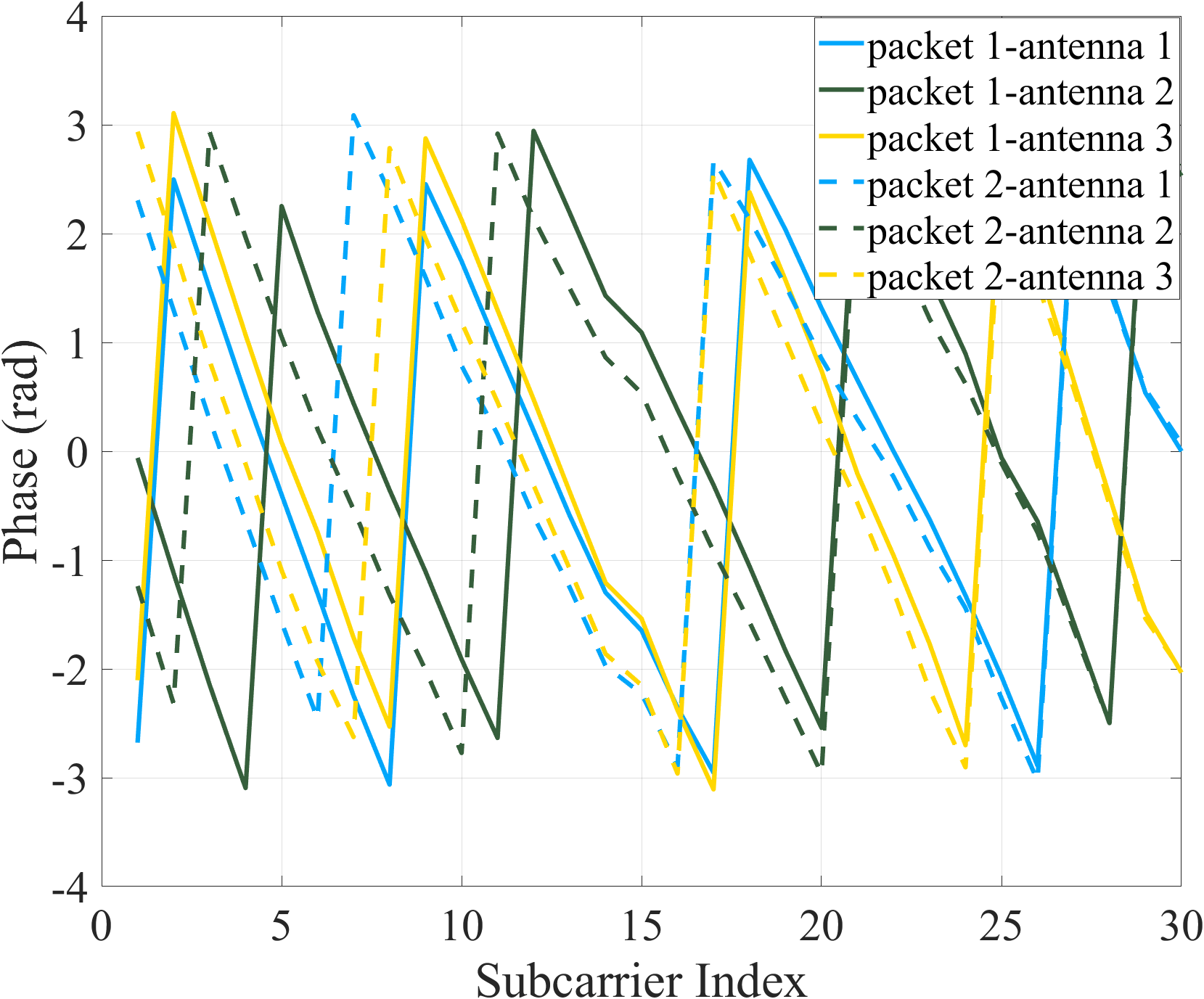}%
		\label{figLT_first_case}}
	\hfil
	\subfigure[ ]{\includegraphics[width=1.7in]{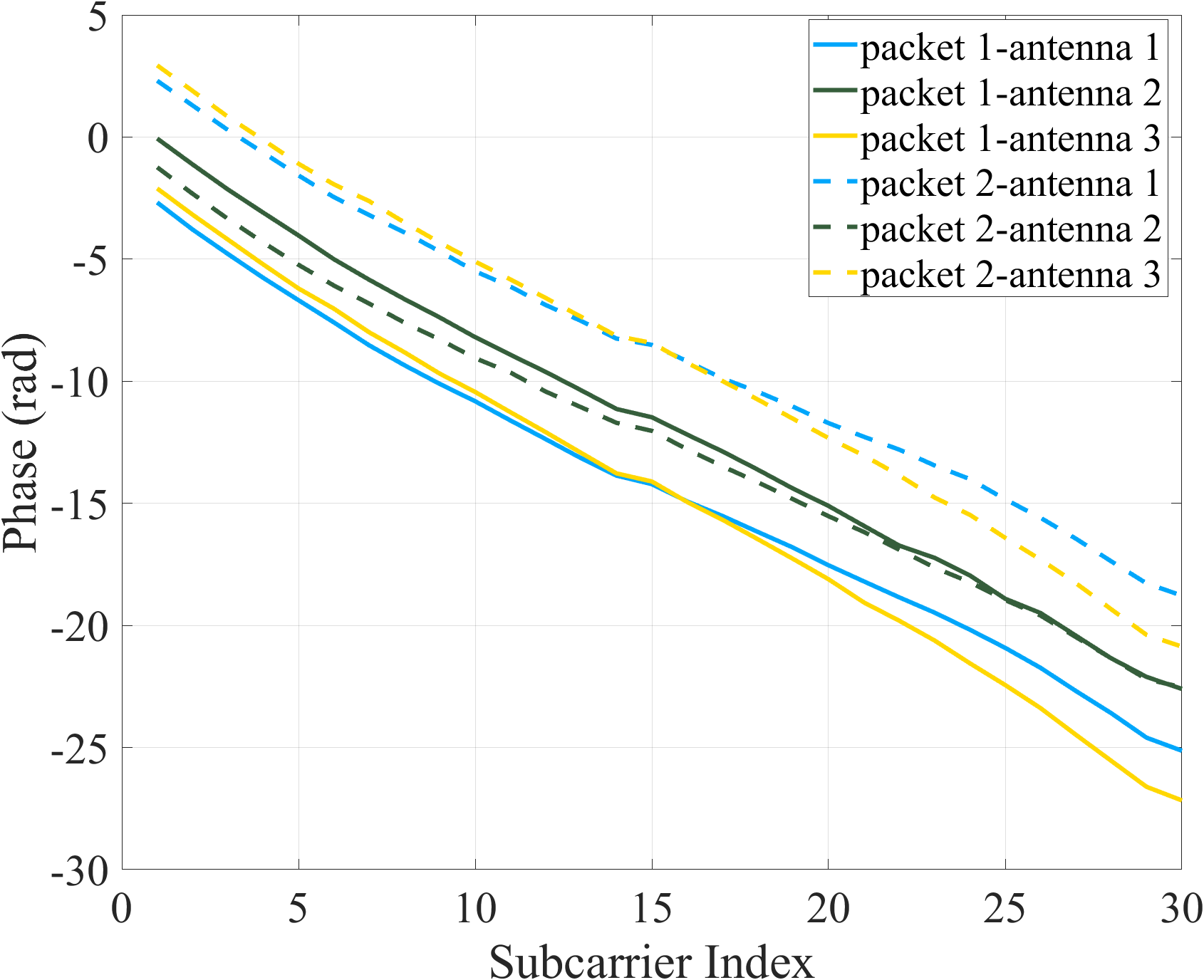}%
		\label{figLT_second_case}}
	\hfil
	\subfigure[ ]{\includegraphics[width=1.7in]{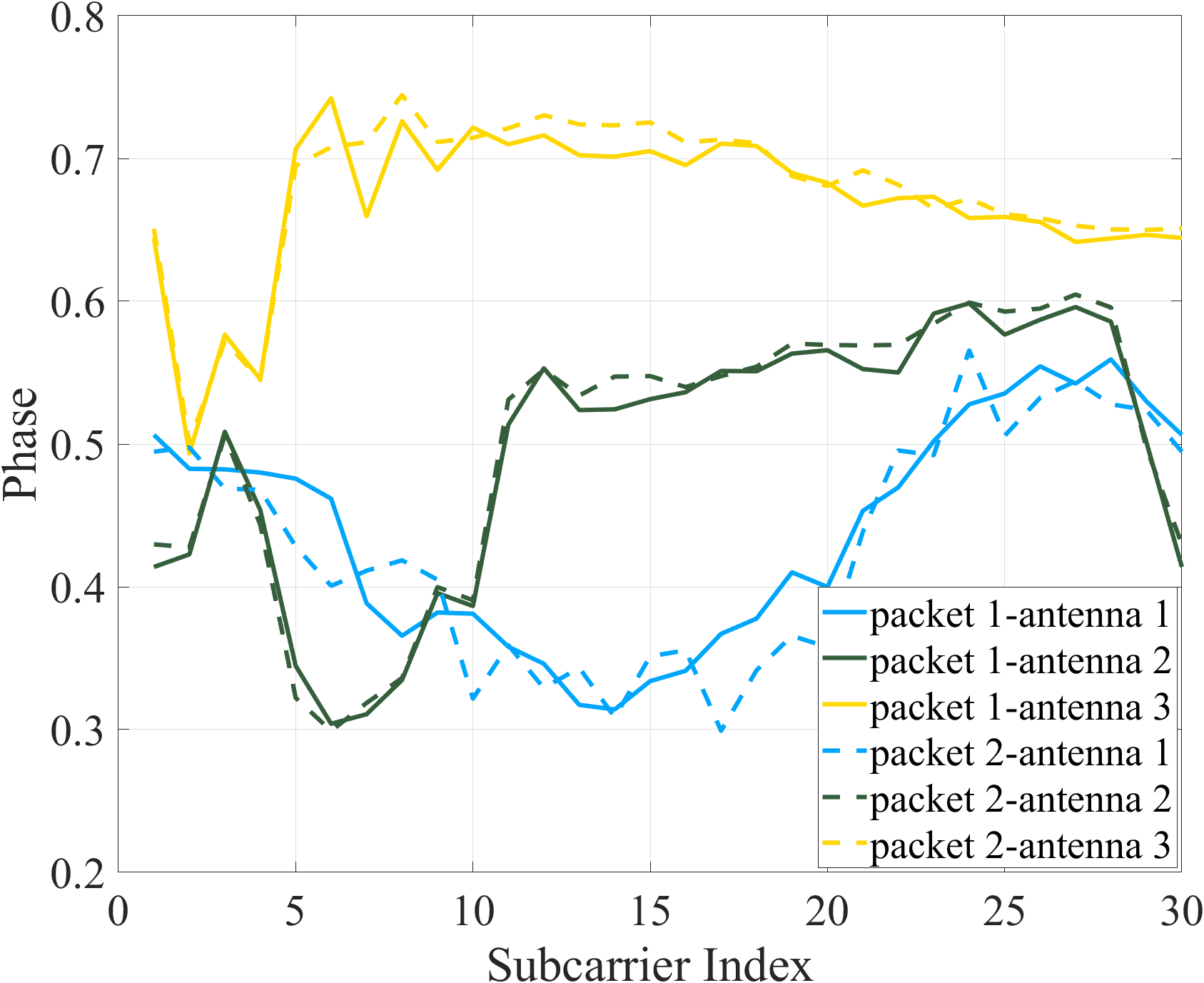}%
		\label{figLT_third_case}}
	\caption{LT for phase. (a) Measured raw phase. (b) Unwrapped phase. (c) Linear transformed phase. }
	\label{fig_LT}
\end{figure*}

\section{DF-Loc system}
Figure \ref{fig_1} illustrates the overall architecture of DF-Loc, which comprises three key components. Detailed descriptions of each component will be provided in subsequent sections.
\subsection{Preprocessing of CSI data} 

\subsubsection{HWF Module for Amplitude} \label{HWF Module for Amplitude}
Multipath effects exhibit stochastic and time-varying characteristics, significantly disrupting the transmission of indoor signals and introducing noise between collected data packets. To mitigate these disturbances, a CSI preprocessing module named QC has been developed. This module integrates a Hampel identifier, wavelet filtering techniques, and a Butterworth low-pass filter, with the objective of extracting clearer and more stable CSI fingerprints to enhance the accuracy of fingerprint positioning.
\paragraph{Hampel Identifier}The Hampel identifier \cite{liu2015contactless} is a robust outlier detection algorithm that relies on the median and Median Absolute Deviation (MAD) within a sliding window. The methodology involves setting a fixed-size sliding window, calculating the median and MAD for the data within the window, and determining outliers by applying a predefined threshold multiplier. Key parameters of the Hampel identifier include window size, threshold multiplier, and MAD scaling factor. By appropriately selecting these parameters, the Hampel identifier effectively preserves the primary trend of the data while accurately identifying and removing outliers, thereby enhancing the precision of data analysis and processing. Additionally, the Hampel identifier can replace outliers in individual subcarriers with significant temporal fluctuations using the sliding window median (The sliding window size $ N_w = 200 $), as illustrated in Figure \ref{fig1_first_case}.

\paragraph{Wavelet Filter}
To mitigate temporal jitter in samples across each subcarrier, we employed a wavelet filter based on an enhanced wavelet thresholding technique to smooth each subcarrier sequence. In this study, we developed a novel threshold function that not only further reduces the constant bias of wavelet coefficients but also ensures continuity and higher-order differentiability within the wavelet domain. The formulation is presented as follows:
\begin{equation}
f\left( X \right) =\begin{cases}
	\operatorname{sign}\left( X \right) \left( \left| X \right|-\frac{2T}{e^{\left( \frac{\mid X\mid -T}{T} \right)} +1} \right), &\left| X \right|\geqslant T\\
	0,&\left| X \right| <T		\\
\end{cases}
\label{wavelet thresholding}
\end{equation}
where $ X $ denotes the original wavelet coefficient, and $ T $ represents the threshold. Within the wavelet domain where $ \left| X \right|<T $, the coefficients are entirely set to zero, analogous to both hard and soft thresholding methods. For $ \left| X \right|>T $, when $ \left| X \right| $ is close to $ T $, the function $ f(X) $ approximates $ \operatorname{sign}\left( X \right) \left( \left| X \right|-T \right) $, thereby making Equation \ref{wavelet thresholding} similar to a soft threshold function. Conversely, when $ \left| X \right| $ significantly exceeds $ T $, $ f(X) $ approaches $ X $, resulting in an approximation to a hard threshold function. As illustrated in Figure \ref{figLT_second_case}, the smoothness of each subcarrier after wavelet filtering is substantially greater than that of unfiltered subcarriers. 
\paragraph{Butterworth Low-Pass Filter}
To further minimize errors introduced during manual data collection, a Butterworth low-pass filter is employed to denoise each subcarrier CSI data packet sequence in the time domain by setting a cutoff frequency that permits only low-frequency signals while attenuating high-frequency noise. The key parameters of this filter include the cutoff frequency, which delineates the boundary between signal and noise, and the filter order, which affects the steepness of the filter’s transition band and overall filtering performance. This filter ensures smooth and ripple-free data sequences, effectively enhancing the signal-to-noise ratio and preserving the primary characteristics of the signal. Additionally, its simple structure facilitates easy implementation, making it suitable for a wide range of application scenarios. As illustrated in Figure \ref{figLT_third_case}, the standard deviation (STD) of each subcarrier is markedly reduced following filtration compared to their unfiltered counterparts.

\begin{figure}[!t]
	\centerline{\includegraphics[width=3.5in]{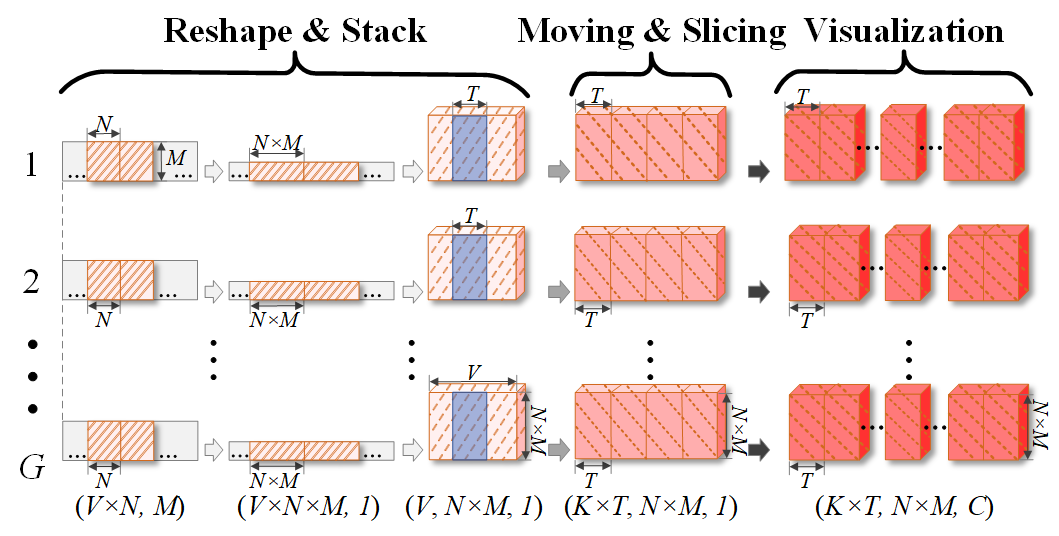}}
	\caption{The mechanism of CSI fingerprint construction involves several key parameters: $ N $, representing the number of subcarriers; $ M $, the number of antennas; $ V $, the number of data packets; $ T $, the size of the sliding window; $ K $, the number of reconstructed fingerprint samples in each RP; and $ G $, the number of RPs.}
	\label{fig2}
\end{figure}

\subsubsection{LC + HWF Module for Phase}
In practice, clock asynchrony between transmitters and receivers complicates the measurement of Time of Flight (ToF) parameter variations. In addition to Sampling Time Offset (STO), each WiFi transmitter-receiver pair experiences Sampling Frequency Offset (SFO) and Carrier Frequency Offset (CFO). These offsets introduce additional noise into cross-packet ToF estimations, resulting in phase shift errors between adjacent subcarriers. Consequently, the similarity of raw phase fingerprints from different locations is insufficient for direct fingerprint-based positioning. To extract CSI phase fingerprint features, we designed a Linear Calibration(LC) module for phase calibration. Specifically, the raw phase (as shown in Figure \ref{figLT_first_case}) is first unwrapped, and the unwrapped phase (illustrated in Figure \ref{figLT_second_case}) is processed using Equations (\ref{LT_1}),(\ref{LT_2}) \cite{kotaru2015spotfi} and (\ref{LT_3}) to mitigate the effects of STO, SFO, and CFO.
\begin{equation}
\widehat{\tau_{s, i}}=\underset{\rho}{\arg \min } \sum_{m, n=1}^{M, N}\left(\psi_{i}(m, n)+2 \pi f_{\delta}(n-1) \rho+\beta\right)^{2}
\label{LT_1}
\end{equation}
\begin{equation}
	\widehat{\psi_{i}}(m, n)=\psi_{i}(m, n)+2 \pi f_{\delta}(n-1) \widehat{\tau_{s, i}}
	\label{LT_2}
\end{equation}
\begin{equation}
	\tilde{\psi}\left( n \right) =\widehat{\psi _i}\left( n \right) -\frac{\widehat{\psi _i}\left( N \right) -\widehat{\psi _i}\left( 1 \right)}{d\left( N \right) -d\left( 1 \right)}\times d\left( n \right) -\frac{1}{N}\sum_{n=1}^N{\widehat{\psi _i}}\left( n \right)
	\label{LT_3}
\end{equation}
where $ \psi _i\left( m,n \right) $ denotes the unwrapped phase response of the $ i $-th packet, with $ m $ representing the antenna index and $ n $ the subcarrier index. $ \tau _{s,i} $ refers to the STO of the $ i $-th packet, while $ f_{\delta} $ is the frequency spacing between two adjacent subcarriers. $ \rho $ and $ \beta $ are the coefficients in the least squares method. $ \tilde{\psi} $ represents the transformed phase, while $ \widehat{\psi} $ denotes the phase after STO sanitizing. $ d(k) $ refers to the pilot subcarrier index in the OFDM symbol. As shown in Figure \ref{figLT_third_case}, the raw phase is processed through the LC module. Subsequently, the phase is further refined using the QC module, following a process similar to the amplitude sequence processing described in Section \ref{HWF Module for Amplitude}.

\begin{figure}[!t]
	\centering
		\subfigure[Amplitude at location 1]{
		\includegraphics[width=0.2\textwidth]{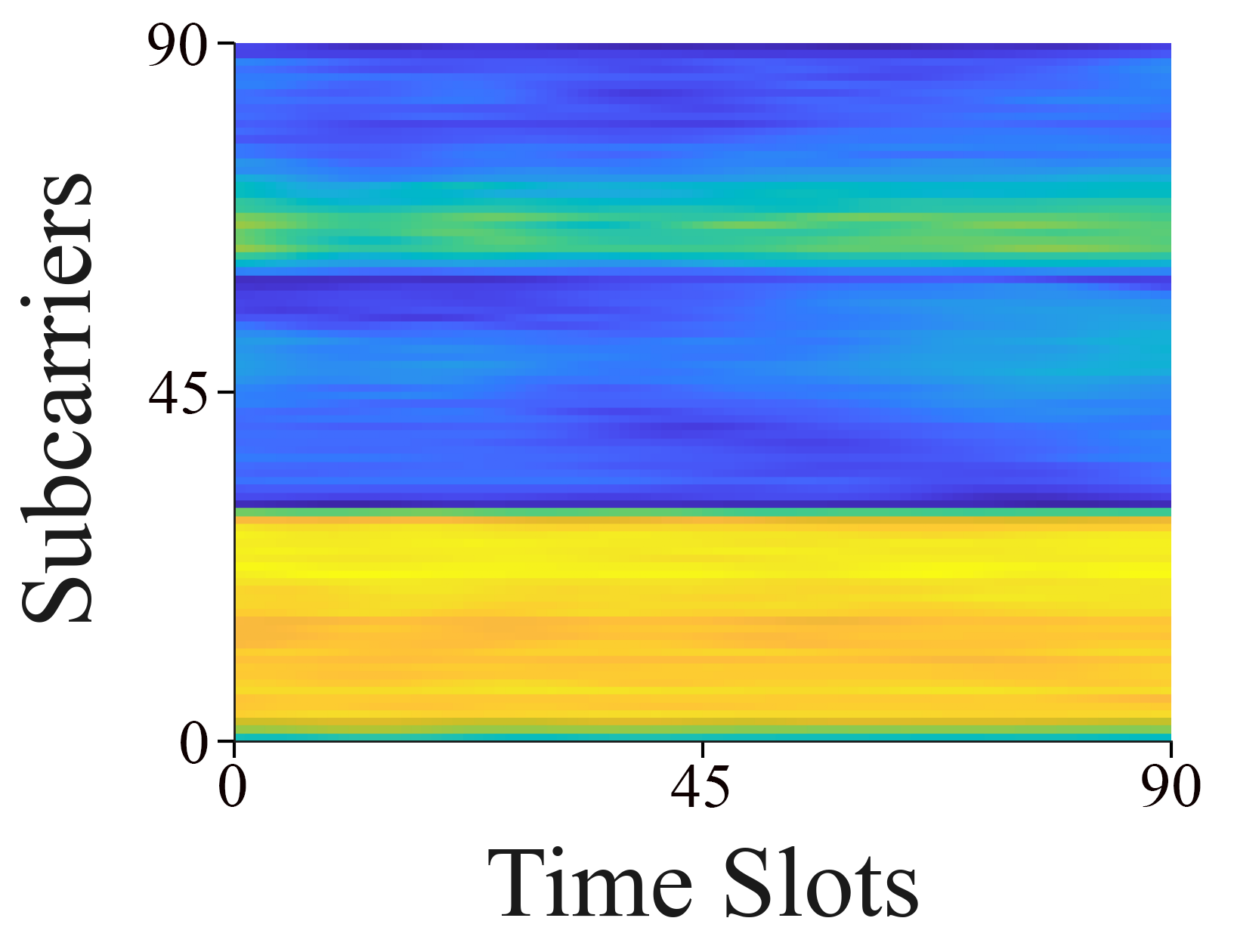}
		\label{csiimage_first_case}} 
	\hfil
	\subfigure[Amplitude at location 2]{ 
		\includegraphics[width=0.2\textwidth]{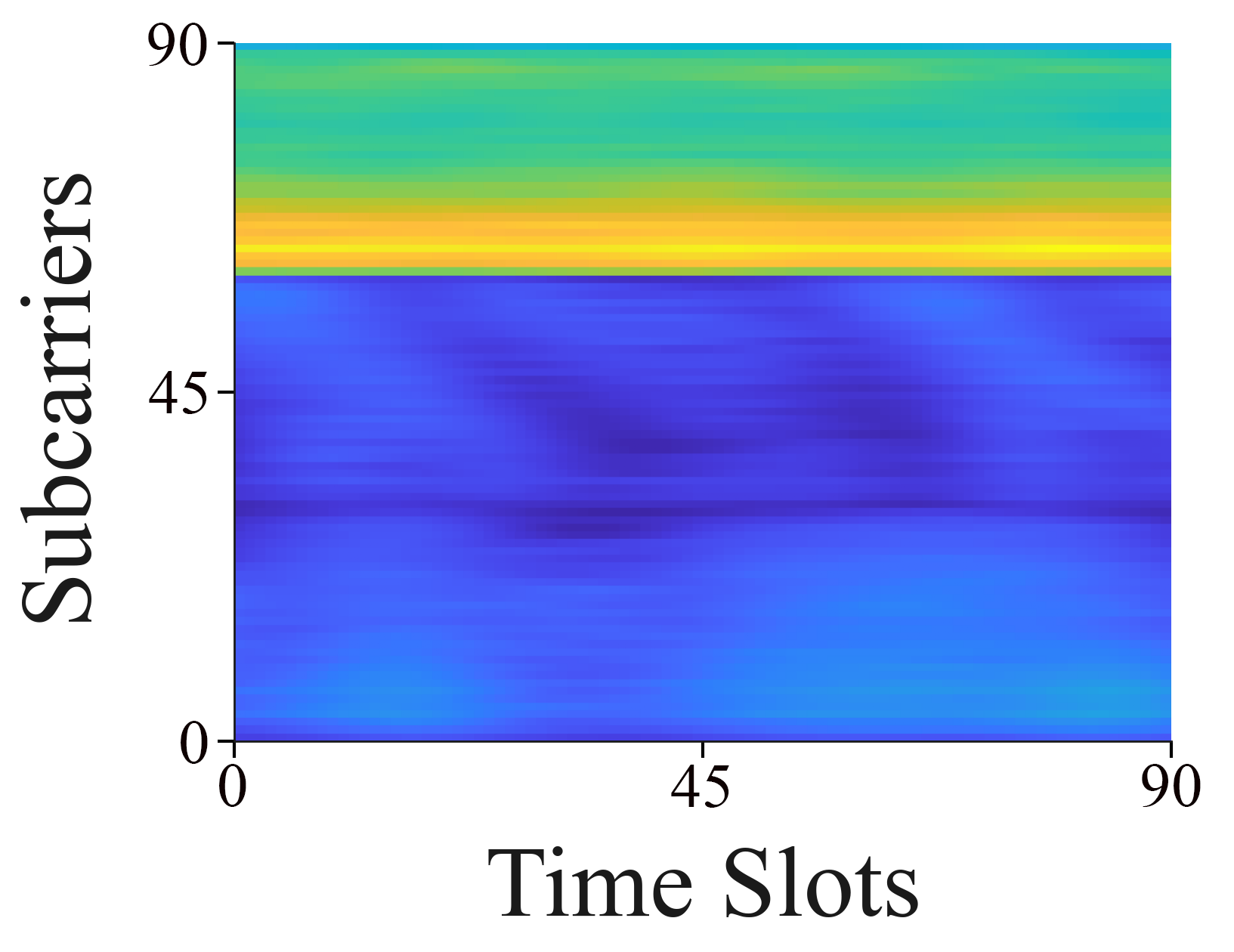}
		\label{csiimage_second_case}} 
	\\
		\subfigure[Phase at location 1]{
		\includegraphics[width=0.2\textwidth]{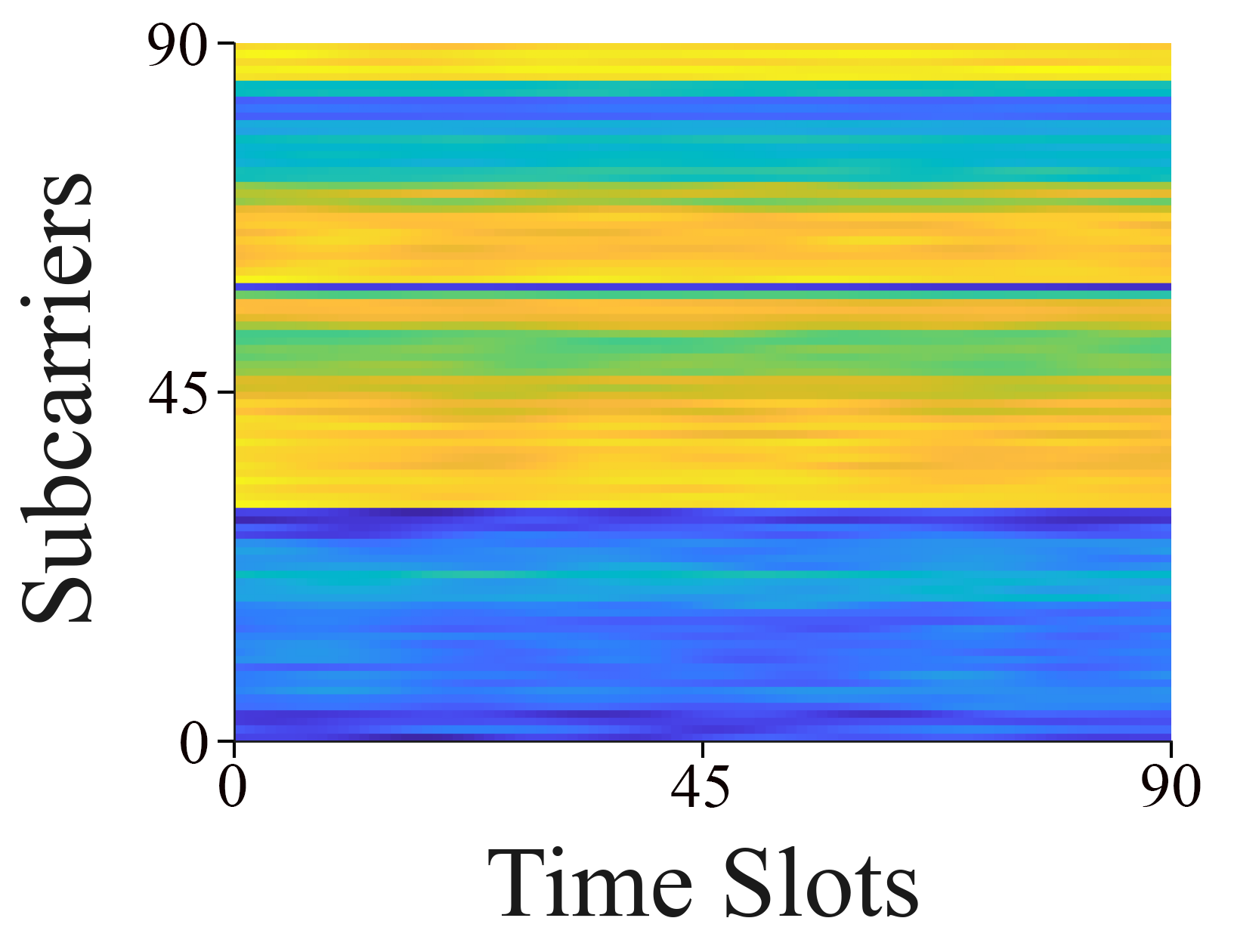} 
		\label{csiimage_third_case}} 
	\hfil
	\subfigure[Phase at location 2]{ 
		\includegraphics[width=0.2\textwidth]{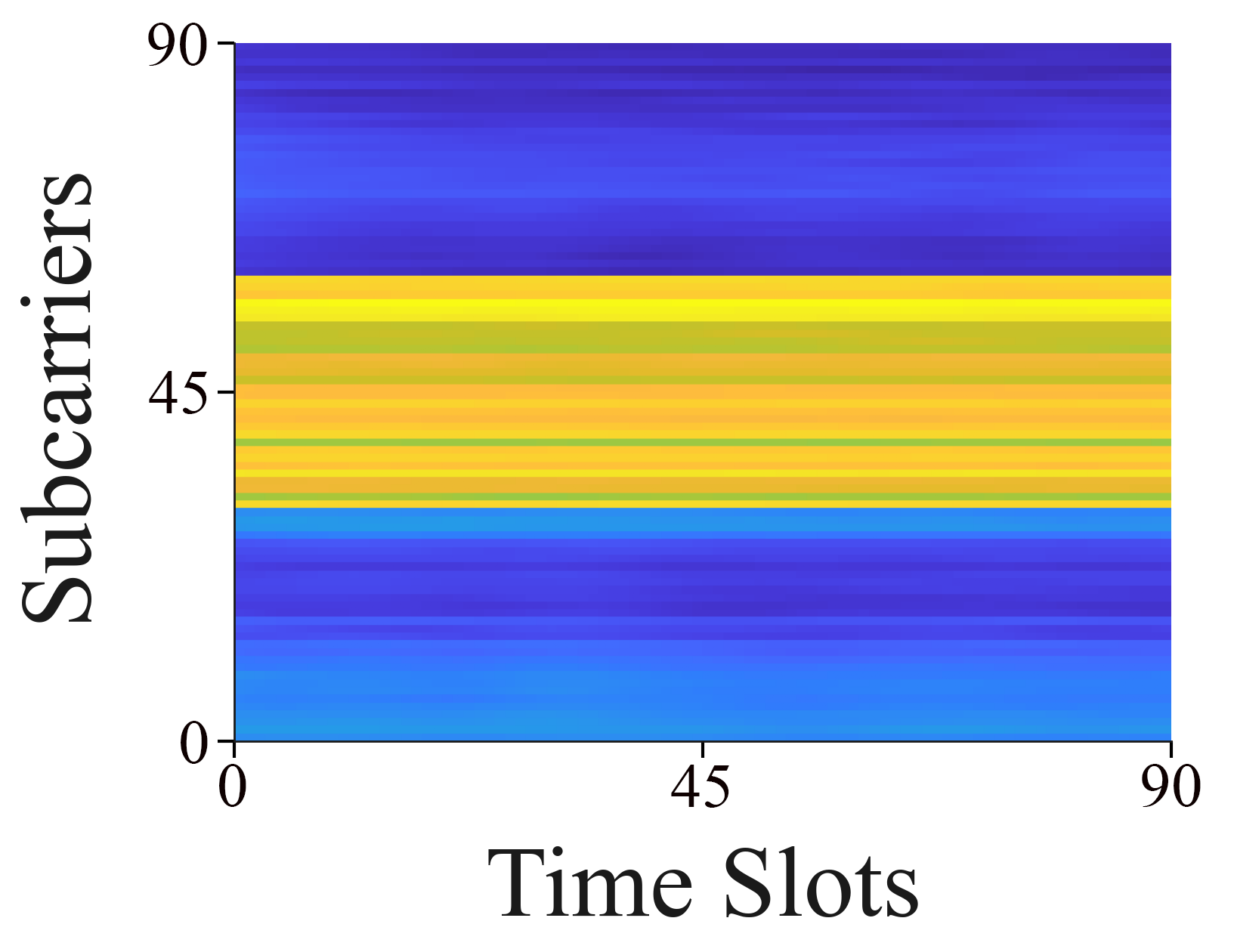} 
		\label{csiimage_fouth_case}} 

	\caption{The newly designed CSI fingerprints.} 
	\label{CSI images} 
\end{figure}
\subsection{CSI Fingerprint Design}
Orthogonal Frequency Division Multiplexing (OFDM) is a widely adopted technology in modern wireless communication standards. The Channel State Information (CSI) extracted from OFDM receivers reveals the multipath characteristics of the wireless channel. Typically, CSI is defined as $ H_n=\left| H_n \right|e^{j\sin \left( \angle H_n \right)} $, where $ \left| H_n \right| $ and $ \angle H_n $ represent the amplitude and phase of the $ n $ -th subcarrier, respectively. By providing information about different fading or scattering paths across multiple subcarriers, CSI enables the construction of robust fingerprints for each location, making accurate positioning systems feasible. In this paper, we extract amplitude and phase information from multiple CSI packets obtained from APs. As shown in Figure \ref{fig2}, the CSI heatmap is constructed through three processing steps, with the resulting heatmap illustrated in Figure \ref{CSI images}. Here, $ N=30 $, $ M=3 $, $ V=1000 $, $ T=90 $, $ V=K \times T $. Converting CSI data into heatmap images facilitates the recognition of fingerprint similarities and data distributions at different locations. This transformation enhances the application and design of subsequent image processing algorithms.

\subsection{MUDA Approach for fingerprinting} \label{Multi-source domain adaption Approach}

Assume $ \mathcal{D} \in \mathbb{R}^d $ is an input measurable space of dimension $ d $, and let $ \mathbb{C} $ denote the set of possible labels. Denote by $ P\left( \mathcal{D} \right) $ the collection of all probability distributions over $ \mathcal{D} $. In this paper, $ {csi}_S\sim P\left( \mathcal{D}_S \right) $ and $ {csi}_T\sim P\left( \mathcal{D}_T \right) $ represent samples from the source and target domains, respectively. Distinct from existing research that assumes differences in either marginal or conditional distributions between domains, this work addresses a more general scenario where both distributions differ. Specifically, $ P\left( {csi}_S \right) \ne P\left( {csi}_T \right) $, $ P\left( {L}_S\mid {csi}_S \right) \ne P\left( {L}_T\mid {csi}_T \right) $. Our objective is to learn a transferable regressor $ f $ that minimizes the risk on the target domain:
\begin{equation}
\epsilon_T=\min P_{(csi,L)\sim\mathcal{D}_T}(f(csi)\neq L).
\end{equation}

To tackle the aforementioned problem, it can be formulated and solved using the principle of Structural Risk Minimization (SRM) in statistical ML \cite{vapnik1998statistical}. Consequently, the predictive function
$ f $ can be formulated as:
\begin{equation}
f = \underset{f \in \mathcal{H}_K, \ (csi, L) \sim \mathcal{D}_l}{\arg\min} \ J(f(csi), L) + \lambda R(f),
\label{2}
\end{equation}
where the first term $ J\left( \cdot ,\cdot \right) $ denotes the loss function, the second term represents the regularization term. $ \mathcal{H}_K $ is the Hilbert space induced by the kernel function $ K\left( \cdot ,\cdot \right) $, the parameter $ \lambda $ serves as the trade-off parameter, and $ \mathcal{D}_l $ denotes the domain with labeled data.

\begin{figure}[!t]
	\centerline{\includegraphics[width=3.5in]{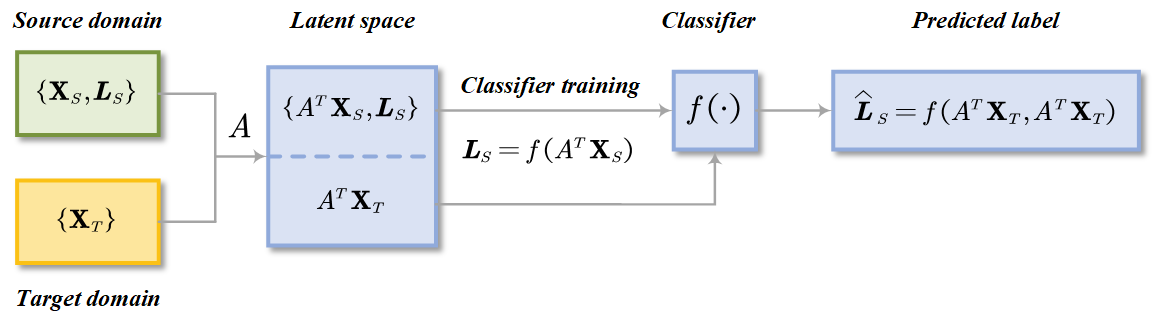}}
	\caption{The overview of the traditional tranfer learning based localization framework.}
	\label{traditional tranfer learning}
\end{figure}

To effectively address the differing distributions between $ \mathcal{D}_S $ and $ mathcal{D}_T $, the regularization term can be further decomposed as:
\begin{equation}
	R(f) = \lambda \overline{D_f}(\mathcal{D}_S, \mathcal{D}_T) + \rho R_f(\mathcal{D}_S, \mathcal{D}_T),
\end{equation}
where $ \overline{D_f}\left( \cdot ,\cdot \right) $ quantifies the distributional discrepancy between $ \mathcal{D}_S $ and $ \mathcal{D}_T $. The parameters $ \lambda $ and $ \rho $ serve as trade-off coefficients, and $ R_f\left( \cdot ,\cdot \right) $ represents additional regularization terms.

As illustrated in Figure \ref{Multi-source domain adaption}, MUDA first conducts feature learning to acquire more transferable representations. Subsequently, it performs alignment of both marginal and conditional distributions. By iteratively and alternately optimizing these two steps multiple times, a domain-invariant regressor $ f $ can be learned. Combining these two steps based on the SRM principle in Equation \ref{2} yields:
\begin{equation}
	\begin{aligned}
	f = \underset{f \in \sum_{i=1}^n \mathcal{H}_K}{\arg\min} & J(f(g(\mathbf{x}_i)), y_i) + \eta \|f\|_{K}^2 \\ 
	& + \lambda \overline{D_f}(\mathcal{D}_S, \mathcal{D}_T) + \rho R_f(\mathcal{D}_S, \mathcal{D}_T),
	\end{aligned}
\end{equation}
where $ g\left( \cdot \right) $ represents the feature learning function, and $ \|f\|_{K}^2 $ denotes the squared norm of $ f $. The term $ \overline{D_f}\left( \cdot ,\cdot \right) $ quantifies the distribution adaptation between domains, $ R_f\left( \cdot ,\cdot \right) $ is the Laplacian regularization component that promotes smoothness. The parameters
$ \eta $, $ \lambda $ and $ \rho $ serve as regularization coefficients.

The aim of distribution adaptation in domain adaptation is to align both the marginal distribution $ P $ and the conditional distribution $ Q $. The distribution alignment term $ \overline{D_f} $ is defined as:
\begin{equation}
\overline{D_f}(\mathcal{D}_S, \mathcal{D}_T) = D_f(P_S, P_T) + \sum_{c=1}^C D_f^{(c)}(Q_S, Q_T),
\label{D_f}
\end{equation}
where $ c\in \left\{ 1,...,C \right\} $ serves as the class indicator. The term $ D_f(P_S, P_T) $ denotes the alignment of the marginal distributions between the source and target domains, $ D_f^{(c)}(Q_S, Q_T) $ represents the alignment of the conditional distributions for each class $ c $.

\subsubsection{Marginal distribution adaptation}

To measure the distributional differences between the two domains, we employ a non-parametric metric known as the Maximum Mean Discrepancy (MMD) \cite{ben2006analysis}. Alternative methods such as the Kullback-Leibler divergence and cross-entropy can also assess inter-domain distributional discrepancies, but MMD offers greater effectiveness \cite{gretton2012kernel}. Accordingly, the MMD distance between the distributions $ \mathcal{P} $ and $ \mathcal{Q} $ is computed as\cite{gretton2012kernel}: 
\begin{equation}
MMD\left(\mathcal{H}_{k},\mathcal{P},\mathcal{Q}\right):=\sup _{\|f\|_{\mathcal{H}_{k}} \leq 1} \mathbf{E}_{X \sim \mathcal{P}} f(X)-\mathbf{E}_{Y \sim \mathcal{Q}} f(Y),
\end{equation}
where $ \|f\|_{\mathcal{H}_{k}} $ represents the unit norm ball in the Hilbert space $ \mathcal{H}_{k} $, and $ \mathbf{E}\left[ \cdot \right] $  denotes the mean of the embedded samples.

To facilitate computation, the empirical expectations over samples $ X $ and $ Y $ are used to replace the population expectations, resulting in a biased empirical estimate of the MMD:
\begin{equation}
MMD_b(\mathcal{H}_k, \mathcal{P}, \mathcal{Q}) = \sup_{\|f\|_{\mathcal{H}_k} \leq 1} \left( \frac{1}{m} \sum_{i=1}^m f(X_i) - \frac{1}{n} \sum_{i=1}^n f(Y_i) \right)
\label{MMD2}
\end{equation}
where $ m $ and $ n $ represent the number of samples in $ \mathcal{P} $ and $ \mathcal{Q} $, respectively.

As illustrated in Figure \ref{traditional tranfer learning}, the divergence in marginal distributions and the divergence in conditional distributions within traditional TL can be calculated using equations \ref{marginal distributions} and \ref{conditional distributions}, respectively \cite{5640675,wang2017balanced}.
\begin{equation}
\begin{aligned}
	D_f\left( P_S,P_T \right) &=\lVert \frac{1}{n_S}\sum_{i=1}^{n_S}{A^T}g\left( csi_i \right) -\frac{1}{n_T}\sum_{j=n_S+1}^n{A^T}g\left( csi_j \right) \rVert ^2 \\
	&=\operatorname{Tr}\left( \boldsymbol{A}^T\mathbf{X}'\,\mathbf{M}_0\,\mathbf{X}'^T\boldsymbol{A} \right)
\end{aligned}
\label{marginal distributions}
\end{equation}
where $ \mathbf{X}'=\left[ \mathbf{X}_S,\mathbf{X}_T \right] $, $ n=n_S+n_t $, $ \operatorname{Tr} $ denote the trace operator, and $ g\left( \cdot \right) $ denote the feature learning function. $ \mathbf{M}_0 $ represents the MMD matrix ,which is defined as follows:
\begin{equation}
\left( \mathbf{M}_0 \right) _{ij}=
 \begin{cases}
	\frac{1}{n_{S}^{2}},&		csi_i,csi_j\in \mathcal{D}_S\\
	\frac{1}{n_{T}^{2}},&		csi_i,csi_j\in \mathcal{D}_T\\
	\frac{-1}{n_Sn_T},&		 otherwise.\\
\end{cases}. 
\label{M0}
\end{equation}

\subsubsection{Conditional distribution adaptation}
The conditional distribution discrepancy serves as another critical metric for evaluating differences between distributions. Here, the class-conditional distribution discrepancy is minimized based on the true positional labels of the samples in the target domain, as detailed below:
\begin{equation}
	\begin{aligned}
	D_{f}^{\left( c \right)}\left( Q_S,Q_T \right) &=\lVert \frac{1}{n_{S}^{\left( c \right)}}\sum_{csi_i\in \mathbf{X}_{S}^{\left( c \right)}}{f\left( A^Tg\left( csi_i \right) \right)} \\
	&-\frac{1}{n_{T}^{\left( c \right)}}\sum_{csi_j\in \mathbf{X}_{T}^{\left( c \right)}}{f\left( A^Tg\left( csi_j \right) \right)} \rVert ^2 \\
	&=\operatorname{Tr}\left( \boldsymbol{A}^T\mathbf{X}' \mathbf{M}_c\,\mathbf{X}'^T\boldsymbol{A} \right)
	\end{aligned} 
\label{conditional distributions}
\end{equation}
where $ c\in \left\{ 1,...,C \right\} $ represents the positional label of the RPs. The MMD matrix $ \mathbf{M}_c $ is defined as follows \cite{long2013transfer}:
\begin{equation}
\left( \mathbf{M}_c \right) _{ij}=
\begin{cases}
	\frac{1}{n_{S}^{\left( c \right)}n_S^{\left( c \right)}},&		csi_i,csi_j\in \mathbf{X}_{S}^{\left( c \right)}\\
	\frac{1}{n_{T}^{\left( c \right)}n_T^{\left( c \right)}},&		csi_i,csi_j\in \mathbf{X}_{T}^{\left( c \right)}\\
	-\frac{1}{n_{S}^{\left( c \right)}n_{T}^{\left( c \right)}},&		csi_i\in \mathbf{X}_{S}^{\left( c \right)},\,csi_j\in \mathbf{X}_{T}^{\left( c \right)}\\
	0,&		otherwise.\\
\end{cases}.
\label{MC}
\end{equation}
where $ \mathbf{X}_{S}^{\left( c \right)} $ and $ \mathbf{X}_{T}^{\left( c \right)} $ represents samples from class $ c $ in $ \mathcal{D}_{S} $ and $ \mathcal{D}_{T} $. $ n_{S}^{\left( c \right)} $ and $ n_{T}^{\left( c \right)} $ denote the number of samples belonging to $ \mathbf{X}_{S}^{\left( c \right)} $ and $ \mathbf{X}_{T}^{\left( c \right)} $, respectively.

Therefore, Equation \ref{D_f} can be expressed as:
\begin{equation}
	\begin{aligned}
	 \sum_{c=1}^C&{\lVert \frac{1}{n_{S}^{\left( c \right)}}\sum_{csi_i\in \mathbf{X}_{S}^{\left( c \right)}}{f\left( A^T\mathbf{z}_i \right)}-
	 	\frac{1}{n_{T}^{\left( c \right)}}\sum_{csi_j\in \mathbf{X}_{T}^{\left( c \right)}}{f\left( A^T\mathbf{z}_j \right)} \rVert ^2} + \\
	& \lVert \frac{1}{n_S}\sum_{i=1}^{n_S}{A^T}\mathbf{z}_i -\frac{1}{n_T}\sum_{j=n_S+1}^n{A^T}\mathbf{z}_j \rVert ^2 \\
	&=\operatorname{Tr}\left( \boldsymbol{A}^T\mathbf{X}'\mathbf{M}\mathbf{X}'^T\boldsymbol{A} \right)
\end{aligned} 
\end{equation}
where $ \mathbf{z}_i=g\left( csi_i \right) $, $ \mathbf{z}_j=g\left( csi_j \right) $ , and the matrix $ \mathbf{M} $ can be obtained as follows:
\begin{equation}
\mathbf{M}=\mathbf{M}_{0}+ \sum_{c=1}^{C} \mathbf{M}_{c} .
\end{equation}

\subsubsection{Two-stage alignment model}
In recent years, Xu et al. \cite{xu2018deep} and Zhao et al. \cite{zhao2018multiple} have integrated DL with multi-source domain adaptation by learning common domain-invariant representations within a shared feature space across all domains. This is accomplished by minimizing the distance between each source domain and the target domain. The formulation is as follows: 
\begin{equation}
	\begin{array}{l}
		\min _{g, f} \sum_{j=1}^{N} \mathbf{E}_{csi \sim \mathbf{X}_{S j}} J\left(f\left(g\left(csi_{i}^{S j}\right)\right), L_{i}^{S j}\right) \\
		\quad+\lambda \sum_{j=1}^{N} {D}\left(g\left(\mathbf{X}_{S j}\right), g\left(\mathbf{X}_{T}\right)\right),
	\end{array}
\end{equation}
where $ {D}\left( \cdot ,\cdot \right) $ represents an estimator of the discrepancy between two domains, $ {g}\left( \cdot \right) $ serves as a feature extractor mapping each domain into a common feature space, and $ f\left( \cdot \right) $ denotes a regressor. The symbol $ N $ indicates the number of distinct underlying source distributions, and $ i $ refers to the sample index.

However, these approaches primarily focus on learning a unified domain-invariant representation across all domains, without addressing domain-specific decision boundaries. For target samples that lie close to such domain-specific’s decision boundaries, different regressors may produce varying coordinate predictions. To address this issue, our DF-Loc method learns a domain adapter for each source-target domain pair, and subsequently aligns the regressor outputs for the target samples.

\begin{figure}[!t]
	\centerline{\includegraphics[width=3.5in]{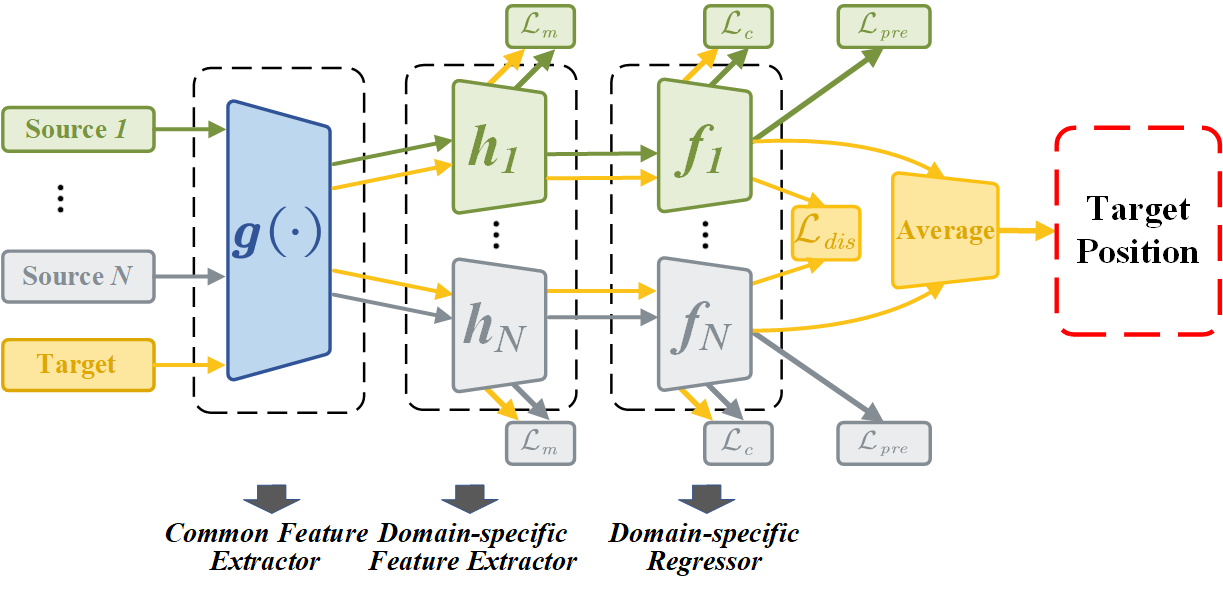}}
	\caption{Two-stage alignment model for fingerprinting.}
	\label{Multi-source domain adaption}
\end{figure}

The dual-stage alignment model aligns the distributions of source-target domain pairs, facilitating consistent predictions across different environments, while enabling simultaneous end-to-end learning of the feature extraction function $ g\left( \cdot \right) $ and the regressor $ f $. As illustrated in Figure \ref{Multi-source domain adaption}, DF-LocNet leverages advanced CNNs to effectively learn the ability to extract representative features. Specifically, a backbone network is employed to learn transferable feature representations, while domain adaptation is achieved through distribution alignment.

Initially, batches of images from the source domain across multiple time periods, $ \left\{ csi_{Sj} \right\} _{j=1}^{N} $, along with batches of images from the target domain,  $ csi_T $ , are provided as inputs to DF-LocNet. Subsequently, the proposed backbone network (illustrated in blue) is designed to project these CSI images from their original feature space into a shared feature space. Following this, $ N $ domain-specific feature extractors, $ \left\{ h_j\left( \cdot \right) \right\} _{j=1}^{N} $, which are not shared among domains, receive the shared features $ g\left( csi_{Sj} \right) $ and $ g\left( csi_T \right) $. These extractors map each pair of source and target data into distinct latent spaces by minimizing the MMD to achieve alignment of marginal distributions. Next, $ N $ domain-specific predictors, $ \left\{ f_j \right\} _{j=1}^{N}  $, composed of fully connected layers, process the domain-invariant features obtained from the domain-specific feature extractor $ h_j\left( g\left( csi \right) \right) $ for $ j $-th source domain, while concurrently aligning the conditional distributions of the source and target domains. Finally, the decision boundaries are aligned through multiple domain-specific regressors to mitigate the inherent bias associated with single-source data and to enhance cross-domain predictive performance.

\paragraph{Domain-specific marginal and Conditional Distribution Alignment}

Based on Equation \ref{D_f}, the learning objective of DF-LocNet can be represented as follows:
\begin{equation}
	\begin{aligned}
	f = \underset{\mathbf{\Theta}}{\min} \sum_{j=1}^{N} \Bigg\{ & \mathbf{E}_{csi \sim \mathbf{X}_{Sj}} J\left( f_j\left( h_j\left( g\left( csi_{i}^{Sj} \right) \right) \right), L_{i}^{Sj} \right) \\
	& + \lambda  \overline{D_f}\left( g\left( \mathbf{X}_{Sj} \right), g\left( \mathbf{X}_T \right) \right) + \rho R_f\left( g\left( \mathbf{X}_T \right) \right)  \Bigg\} 
	\end{aligned}
\end{equation}
where $ J\left( \cdot ,\cdot \right) $ represents the mean squared error (MSE) loss function, and $ \mathbf{\Theta }=\left\{ w,b \right\} $ denotes the set of network parameters, including weights and biases. Position prediction can be formulated either as a classification or a regression problem, with the main difference stemming from the design of the activation function in the network’s final layer. In this study, we treat it as a regression task and approximate conditional distribution alignment by minimizing the MMD between source and target domains within the regressor.

Since DF-LocNet is based on a DNN architecture, its training process employs mini-batch stochastic gradient descent (SGD) rather than using the domain data. This approach ensures that distribution adaptation is calculated only between batches. Such a design is particularly practical and efficient for real-world applications like fingerprint-based localization, where data is often received in a streaming manner.

Traditional MMD-based TL methods are generally based on Equation \ref{MMD2}, which relies on pairwise similarity and exhibits a quadratic time complexity. Moreover, these methods are often simplified to a linear kernel, as shown in Equations \ref{M0} and \ref{MC}. However, in domain adaptation approaches based on DL, such computations tend to be more time-consuming. To address this issue, we adopt the unbiased linear-time approximation of MMD proposed in \cite{gretton2012optimal}, which significantly reduces computational complexity.
\begin{equation}
MMD^2_l(S, T) = \frac{2}{n} \sum_{i=1}^{n/2} h_l(\mathbf{z}_i)
\label{MMDl}
\end{equation}
where $ \mathbf{z}_i=\left[ csi_{2i-1}^{S},csi_{2i}^{S},csi_{2j-1}^{T},csi_{2j}^{T} \right] $, $ h_l $ is an operator that defined on a quad-tuple $ \mathbf{z}_i $ and is defined as follows:
\begin{equation}
	\begin{aligned}
	h_l(z_i) =& k\left(x_{2i-1}^s, x_{2i}^s\right) + k\left(x_{2j-1}^t, x_{2j}^t\right) - k\left(x_{2i-1}^s, x_{2j}^t\right) 
	\\ &- k\left(x_{2i}^s, x_{2j-1}^t\right)
	\end{aligned}
\end{equation}
where $ k $ denotes a characteristic kernel. Thus, the summation approximation of Equation (28) is suitable for gradient computation using mini-batch processing.

The gradients of the parameters are defined as follows:
\begin{equation}
\Delta_{\Theta} = \frac{\partial J(\cdot, \cdot)}{\partial \Theta} + \lambda \frac{\partial \overline{D_f}(\cdot, \cdot)}{\partial \Theta} + \rho \frac{\partial R_f(\cdot, \cdot)}{\partial \Theta}.
\end{equation}

In accordance with Equation \ref{MMDl}, each domain-specific feature extractor aligns the marginal distributions for every source-target domain pair by minimizing Equation \ref{Loss1}:
\begin{equation}
	\mathcal{L}_{m}=\frac{1}{N}\sum_{j=1}^N{{D}}\left( h_j\left( g\left( csi_{Sj} \right) \right) ,h_j\left( g\left( csi_T \right) \right) \right)
	\label{Loss1}
\end{equation}

Similarly, the alignment of conditional distributions is achieved by minimizing the following MMD loss:
\begin{equation}
\mathcal{L}_c=\frac{1}{N}\sum_{j=1}^N{D}\left( f_j\left( h_j\left( g\left( csi_{Sj} \right) \right) \right) ,f_j\left( h_j\left( g\left( csi_T \right) \right) \right) \right)
\end{equation}

\begin{figure*}[!t]
	\centering
	\subfigure[]{\includegraphics[width=3.0in]{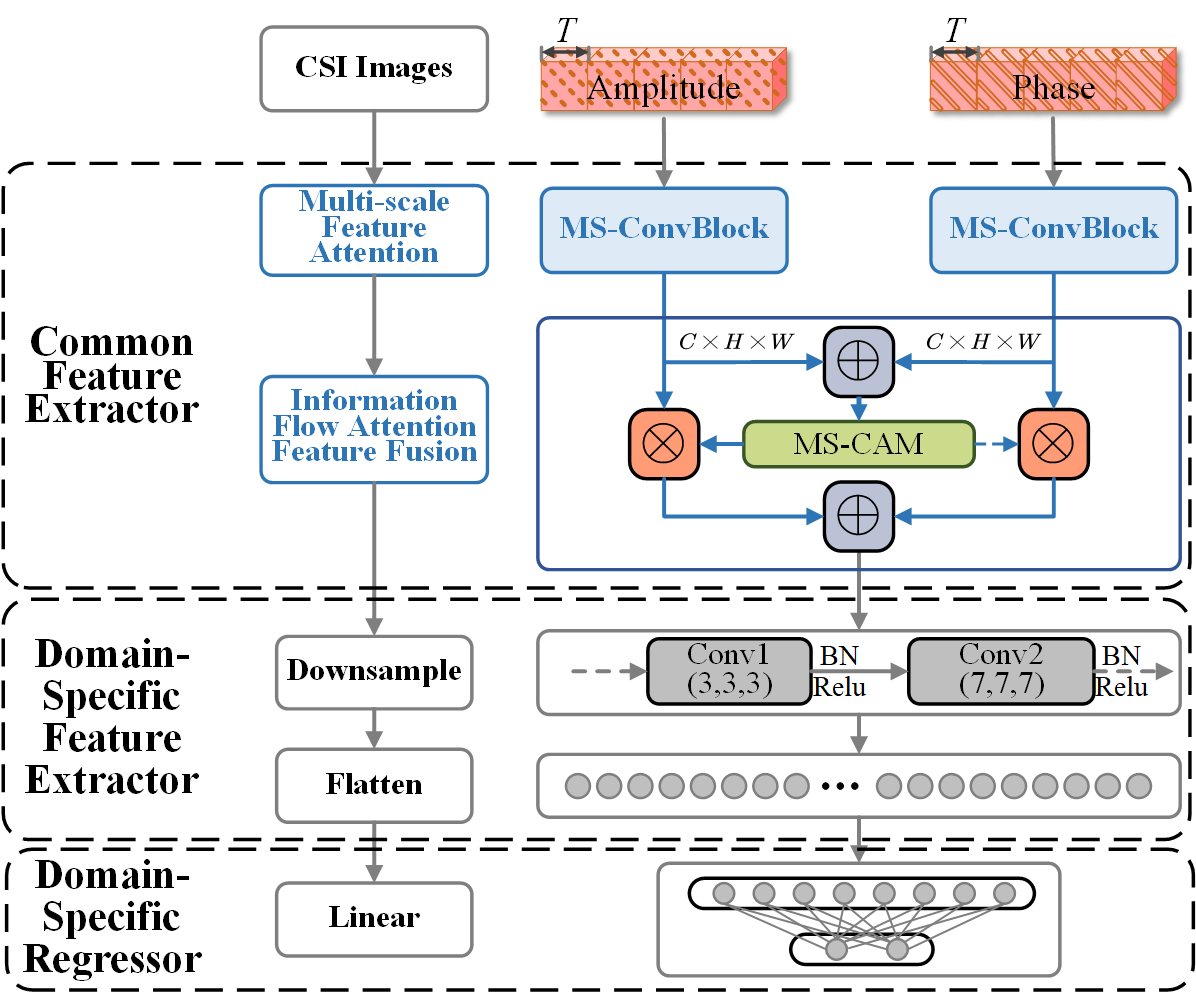}%
		\label{DF-LocNet_first_case}}
	\hfil
	\subfigure[]{\includegraphics[width=2.1in]{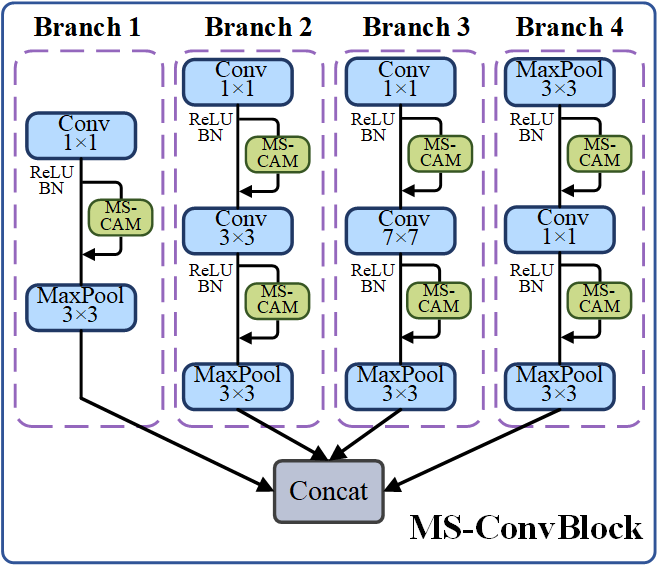}%
		\label{DF-LocNet_second_case}}
	\hfil
	\subfigure[]{\includegraphics[width=1.6in]{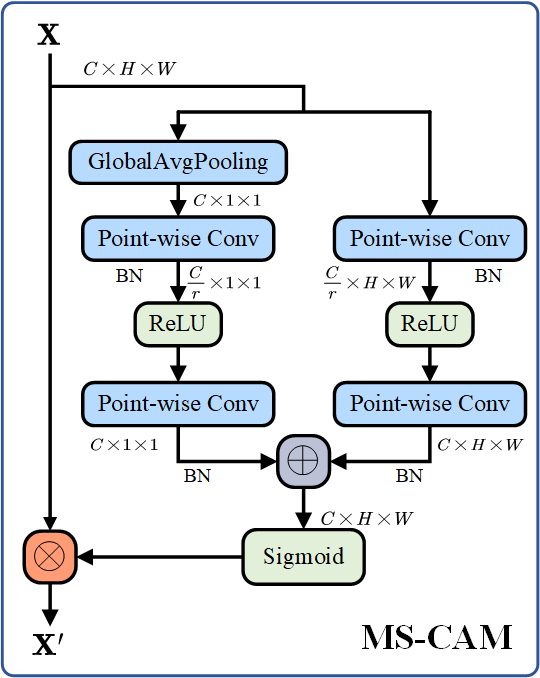}%
		\label{DF-LocNet_third_case}}
	\caption{The DF-LocNet workflow for dynamic positioning utilizes MUDA. (a) Overall Architecture of DF-LocNet. (b) Structure of MS-ConvBlock. (c) Structure of MS-CAM.}
	\label{DF-LocNet}
\end{figure*}

\paragraph{Domain-specific Regressor Alignment}

Target samples approaching regression boundaries are more prone to estimation errors by regressors trained on source domains. These regressors originate from diverse source domains and may generate inconsistent predictions for the same target sample, especially within boundary regions. Ideally, each regressor sould produce similar outputs for identical samples. Consequently, the second stage of alignment focuses on minimizing the discrepancies among regressors. In this study, the Euclidean distance between the outputs of all regressors on target domain data is utilized as the loss function:
\begin{equation}
	\begin{aligned}
		\mathcal{L}_{dis} &= \frac{2}{N \times (N-1)} \sum_{j=1}^{N-1} \sum_{i=j+1}^N \mathbf{E}_{csi_{T} \sim \mathbf{X}_T} \Big[ \\
		& \quad \lVert f_i \left( h_i \left( g \left( csi_T \right) \right) \right) - f_j \left( h_j \left( g \left( csi_T \right) \right) \right) \rVert_2 \Big]
	\end{aligned}
\label{Regressor Alignment}
\end{equation}
By minimizing Equation \ref{Regressor Alignment}, the outputs of the regression models are aligned. 

Furthermore, for each regressor, a prediction loss is calculated using MSE, as defined by the following equation:
\begin{equation}
	\mathcal{L}_{pre}=\sum_{j=1}^N{\mathbf{E}_{csi\sim \mathbf{X}_{Sj}}}J\left( f_j\left( h_j\left( g\left( csi_{i}^{Sj} \right) \right) \right) ,L_{i}^{Sj} \right)
\end{equation}
where $ i $ denotes sample index. In summary, DF-Loc comprises two alignment phases: the learning of source-specific domain-invariant representations and the alignment of regressor outputs for target samples. The loss function of this approach encompasses prediction loss, marginal distribution discrepancy loss, conditional distribution discrepancy loss, and regressor alignment loss. Specifically, the network enhances the predictive accuracy of source domain data by minimizing regression errors, facilitates the learning of domain-invariant representations by reducing MMD loss, and aligns the outputs of individual regressors by decreasing discrepancy loss. The overall loss is formulated as follows:
\begin{equation}
\mathcal{L}_{total}=\mathcal{L}_{pre}+\lambda \left( \mathcal{L}_m+\mathcal{L}_c \right) +\rho \mathcal{L}_{dis}.
\end{equation}

The training process primarily adheres to the standard mini-batch SGD algorithm and sequentially trains the source-specific networks. Ultimately, the predicted value for a target sample is determined by averaging the outputs of all regression models:
\begin{equation}
\hat{L}=\frac{1}{N}\sum_{j=1}^N{\hat{L}_j}
\end{equation}
where $ j $ denotes the index of the regression model.

\subsection{Multi-Scale AFF Network for Feature Extraction}
The constructed CSI images for each location exhibit significant distinguishability in the feature dimension, as shown in Figure 5, but are less distinct in the sample dimension and contain two streams of information: amplitude and phase. To address this, we designed an fingerprint positioning network named DF-LocNet, which is based on multi-scale convolutional AFF. This network captures multi-level fingerprint features (using Multi-Scale Convolutional Blocks, MS-ConvBlock), highlights key features while suppressing irrelevant information (using MS-CAM), and efficiently fuses features to enhance the model's overall comprehension (using AFF).

Specifically, to adapt to the MUDA method proposed in Section \ref{Multi-source domain adaption Approach}, DF-LocNet is divided into three components: common feature extractor, domain-specific feature extractors, and domain-specific regressors. The common feature extractor employs two MS-ConvBlocks to learn critical local information from amplitude and phase fingerprints, and these are integrated through the AFF module, thereby improving the model's holistic understanding. The domain-specific feature extractors reduce the dimensionality of feature maps and perform domain alignment. Finally, the domain-specific regressors predict the two-dimensional coordinates of the input fingerprint samples through fully connected layers. The architecture of DF-LocNet is presented in Figure \ref{DF-LocNet_first_case}, with implementation details outlined below.

\subsubsection{Multi-scale Convolutional Feature Learning}
As illustrated in Figure \ref{DF-LocNet_second_case} and inspired by the Inception network architecture \cite{szegedy2015going}, the backbone network is constructed using parallel multi-scale feature extraction combined with the MC-CAM design. This configuration allows the network to focus on detailed features while simultaneously capturing global context, thereby enhancing its ability to adapt flexibly to diverse input feature distributions and improving the model's representational capacity. Consequently, it achieves a balance between network depth and width for this task. Specifically, the reconstructed amplitude and phase features, with dimensions $ \mathbb{R}^{C\times H\times W} $, are input into parallel two-dimensional convolutional neural networks (CNN) for feature extraction. To mitigate overfitting and vanishing gradient issues during optimization, $ 1\times1 $ convolutions are employed for dimensionality reduction, thereby reducing redundant features and enhancing computational efficiency, and Rectified Linear Units (ReLU) are utilized as activation functions. The output of the CNN is formalized as:
\begin{equation}
\hbar ^{\left( l \right)}=\delta\left( \omega^{\left( l \right)}\circledast csi^{\left( l-1 \right)}+b^{\left( l \right)} \right)
\end{equation}
where $ \delta $ represents the ReLU, $ \omega^{\left( l \right)}\in \mathbb{R}^{\text{N}_k} $ represents the learnable convolution kernels, $ b^{\left( l \right)}\in \mathbb{R}^{\text{N}_f} $ denotes the learnable biases, $ \text{N}_k $ is the size of the convolution kernels, $ \text{N}_f $ is the number of filters, and $ \circledast $ signifies the convolution operation.

\subsubsection{MS-CAM}
The MS-CAM addresses the challenge of insufficient feature consistency across different scales by integrating both global and local contextual information, as illustrated in Figure \ref{DF-LocNet_third_case}. By varying the spatial pooling sizes, MS-CAM aggregates multi-scale contextual features along the channel dimension, enabling it to simultaneously emphasize the globally distributed features of large objects and the locally distributed features of small objects. The local channel context $ \mathscr{L}\left( \mathbf{X} \right) \in \mathbb{R}^{C\times H\times W} $, comprising $ C $ channels and feature maps of size $ H\times W $, is computed using a bottleneck structure, as described below:
\begin{equation}
\mathscr{L}(\mathbf{X})=\mathcal{B}\left(\mathrm{PWConv}_{2}\left(\delta\left(\mathcal{B}\left(\mathrm{PWConv}_{1}(\mathbf{X})\right)\right)\right)\right)
\end{equation}
where $ \mathcal{B} $ represents Batch Normalization (BN), and the $ \mathrm{PWConv} $ denotes Point-Wise Convolution. The convolutional kernel dimensions of $ \mathrm{PWConv}_{1} $ and $ \mathrm{PWConv}_{2} $ are $ \frac{C}{r}\times C\times 1\times 1 $ and $ C\times \frac{C}{r}\times 1\times 1 $, respectively, where $ r $ denotes the channel reduction ratio, thereby conserving parameters. The tensor $ \mathscr{L}(\mathbf{X}) $ maintains the same shape as the input features. Given the global channel context $ \mathscr{G}\left( \mathbf{X} \right) $ and the local channel context $ \mathscr{L}(\mathbf{X}) $, attention weights $ \mathbf{M}(\mathbf{X}) \in \mathbb{R}^{C\times H\times W} $ are computed using the Sigmoid function, and the optimized feature $ \mathbf{X}'\in \mathbb{R}^{C\times H\times W} $ is generated according to the following equation:
\begin{equation}
\mathbf{X}^{\prime}=\mathbf{X} \otimes \mathbf{M}(\mathbf{X})=\mathbf{X} \otimes \sigma(\mathscr{L}(\mathbf{X}) \oplus \mathscr{G}(\mathbf{X}))
\end{equation}
where $ \sigma $ represents the sigmoid function, $ \otimes $ denotes element-wise multiplication, and $ \oplus $ signifies broadcast addition.

\subsubsection{AFF}
Based on the MS-CAM module $ \mathbf{M} $ and as illustrated in Figure \ref{DF-LocNet_first_case}, the network is capable of performing soft selection or weighted averaging between $ \mathbf{X} $ and $ \mathbf{Y} $. Consequently, the designed network can simultaneously fuse amplitude fingerprint information and phase fingerprint information while incorporating contextual information beyond simple initial fusion. The AFF is formulated as follows:  
\begin{equation}
\mathbf{Z}=\mathbf{M}(\mathbf{X} \uplus \mathbf{Y}) \otimes \mathbf{X}+(1-\mathbf{M}(\mathbf{X} \uplus \mathbf{Y})) \otimes \mathbf{Y}
\end{equation}
where the feature maps $ \mathbf{X},\mathbf{Y}\in \mathbb{R}^{C\times H\times W} $ and $ \mathbf{Z}\in \mathbb{R}^{C\times H\times W} $ represent the fused features, and $ \uplus $ denotes initial feature fusion. It is important to note that the sum of the weights applied to the two feature maps must equal 1. Specifically, the fusion weights $ \mathbf{M}(\mathbf{X} \uplus \mathbf{Y}) $ and $ 1-\mathbf{M}(\mathbf{X} \uplus \mathbf{Y}) $ are real numbers within the range $ \left[ 0,1 \right] $.

\begin{figure}[!t]
	\centering
	\subfigure[]{\includegraphics[width=3.5in]{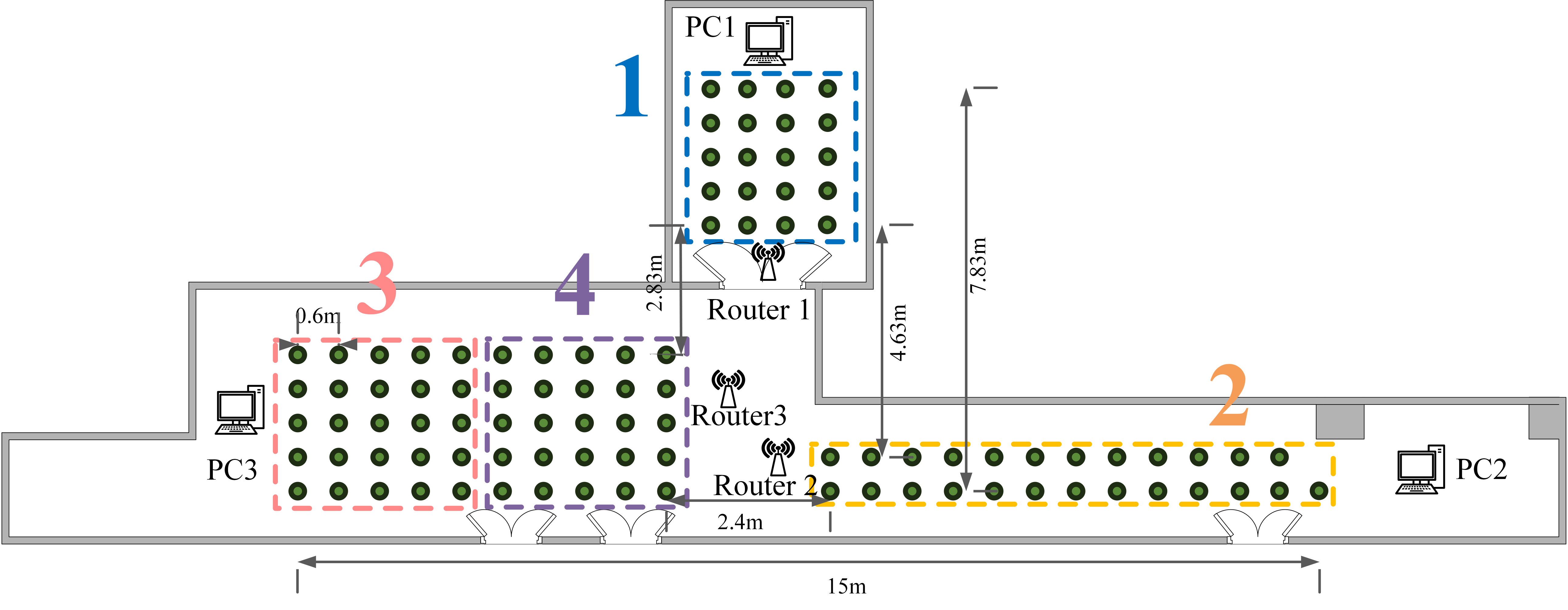}  
		\label{Scenario_1_case}}	
	\subfigure[]{\includegraphics[width=2.0in]{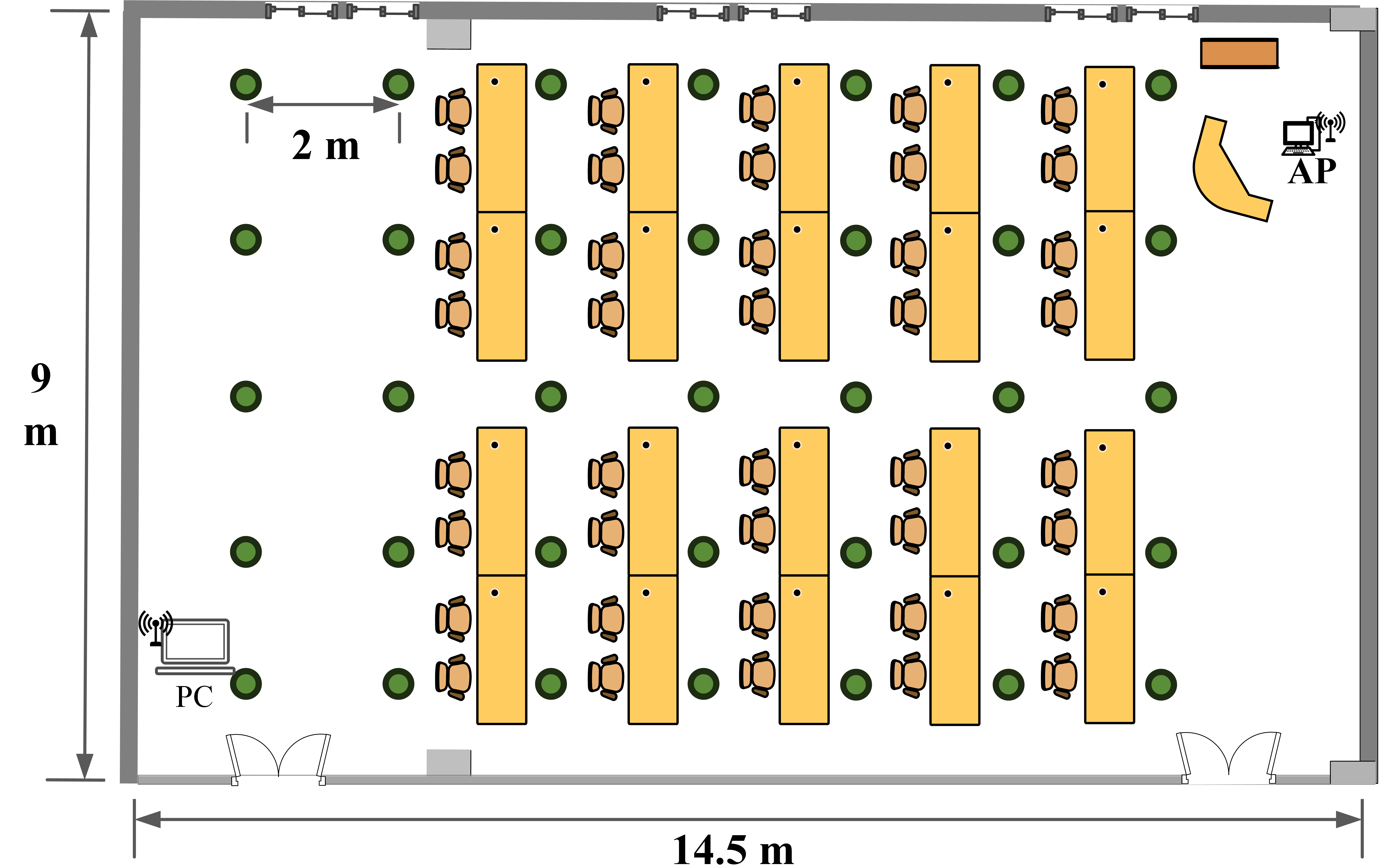}  
		\label{Scenario_2_case}}
	\subfigure[]{\includegraphics[width=1.8in]{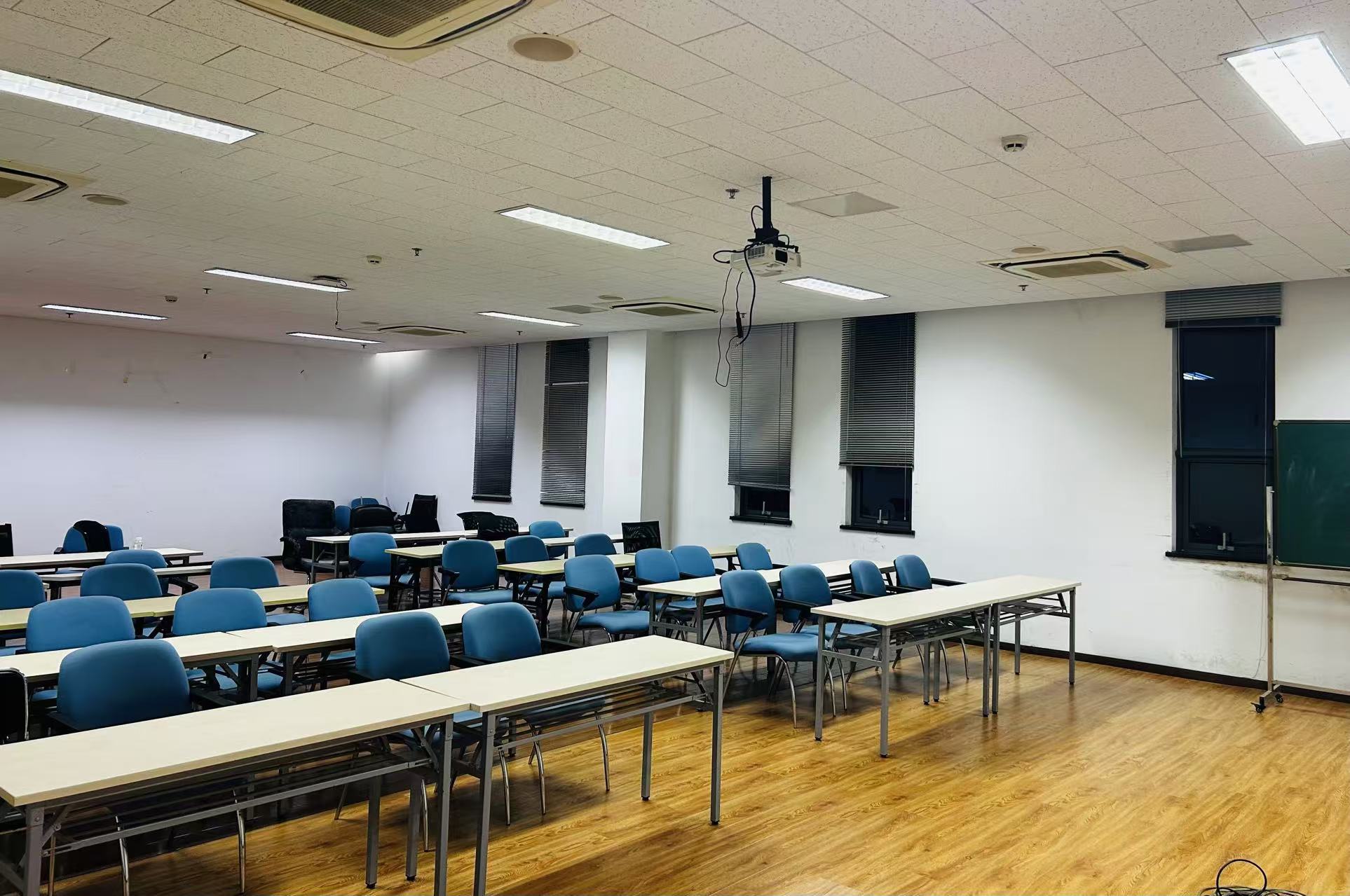}
		\label{Scenario_22_case}}
	\caption{Experimental Scenario Layout(1000 packets). (a) RPs Locations in the Office. (b) RPs Locations in the Classroom. (c)Classroom on-site photograph }
	\label{fig_Scenario12}
\end{figure}

\subsubsection{Fully Connected Regression Positioning}
Regression-based localization is achieved by integrating the outputs from domain-specific feature extractors through a fully connected network. Specifically, the downsampled output $ \chi $ is flattened into a one-dimensional tensor $ \bar{\chi} $ and serves as input to the fully connected network. The final predicted coordinates are expressed as follows:  
\begin{equation}
\hat{L}\left( x,y \right) =\boldsymbol{f}\left( \bar{\chi}\cdot W_{\chi l}+b_l \right) 
\end{equation}
where the function $ \boldsymbol{f}\left( \cdot \right) $ constitutes the linear regression component of the fully connected neural network, with $ W_{\chi l} $ and $ b_l $ representing its weights and biases, respectively. Furthermore, MSE is employed as the loss function, formulated as follows: 
\begin{equation}
	\mathcal{L}_{pre}=\sum_{g=1}^G{\lVert \sum_{k=1}^K{\hat{L}_g}\left( x\left( k \right) ,y\left( k \right) \right) -L_g\left( x,y \right) \rVert _{2}^{2}}
\end{equation}
where $ \lVert \cdot \rVert _2 $ denotes the Euclidean norm, and $ L_g\left( x,y \right) $ represents the true coordinates of the $ g $-th RP.

During the training phase of the localization model based on MUDA, the model's generalization was enhanced and convergence was accelerated by incorporating an L2 regularization term, multiple Dropout layers, and ReLU activation functions. The entire model was trained using an SGD optimizer combined with an adaptive learning rate adjustment strategy. Additionally, to prevent overfitting, an early stopping mechanism and a save-best-only policy were implemented to monitor and retain the optimal model throughout the training process.

\section{EXPERIMENTS AND RESULTS}
To validate the positioning performance of the proposed DF-Loc in indoor dynamic environments utilizing commercial signal access, field tests were conducted in two prominent large-scale indoor settings: a single-story office at Nanjing University of Posts and Telecommunications \cite{wei2023meta} and a classroom within the National Key Laboratory of Mobile Communications at Southeast University. The subsequent sections will provide a comprehensive description of the experimental bench, experimental scenarios, and experimental results.

\subsection{Experimental Bench}
In this study, a testing platform for signal sampling, recording, and CSI acquisition was established. Data collection was performed using a commercial wireless access point (AP), TP-Link TL-WR886N, and a receiver, Thinkpad X201. The receiver operated on a 64-bit Ubuntu 12.04 LTS operating system, capturing $ 1000 $ CSI data packets at a sampling rate of $ 100 $ Hz at each RP over a duration of 10 seconds. The entire system operated at 5 GHz with a bandwidth of 20 MHz to ensure high-quality wireless channels. The distance between adjacent antennas was $ d=2.68 $ cm. 

CSI data preprocessing and image fingerprint construction were conducted using MATLAB. The DF-Loc system was developed on Southeast University's Ascend computing platform, utilizing Python and the PyTorch framework. During the training phase, acceleration was achieved 
with one KUNPENG CPU (24 cores, 128 GB RAM) and one Ascend 910 NPU (312 T computational power, 64 GB memory). 

\subsection{Field Testing Scenarios}

\subsubsection{Office Scenario}Figure \ref{Scenario_1_case} illustrates the experimental setup, comprising three distinct zones: a corridor, a laboratory, and a hall. The corridor is predominantly open, functioning as a pure line-of-sight (LOS) environment, although pedestrian movement is inevitable. In contrast, the laboratory and hall represent NLOS environments due to the presence of numerous obstacles, such as tables, computers, and sofas, which obstruct LOS transmission. These zones are further segmented into areas 1 through 4, containing 20, 25, 25, and 25 RPs respectively \cite{wei2023meta}. At each RP, five unique human activities, including standing, squatting, and walking, were recorded, resulting in a total of 20 indoor localization tasks. The RPs were spaced 0.6 m apart.

	%
\setlength{\tabcolsep}{1.8pt}
\begin{table}[!t]  
	\centering
	\caption{TEST CASE DEFINITION IN THE OFFICE AND CLASSROOM SCENARIOS(OST:Tested in Office under the Same Conditions)}
	\label{tab1tab1}
	\begin{threeparttable}
		\begin{tabular}{c|ccc|ccc} 
			\toprule
			Case  & \multicolumn{3}{c|}{Source Sets} & \multicolumn{3}{c}{Target Set\tnote{1}}\\
			\midrule
			OST  \hfill&\hfill 60\% green dots $\textcolor[rgb]{0.352,0.553,0.220}{\bullet}$& & subset 123   &\hfill 40\% green dots $\textcolor[rgb]{0.352,0.553,0.220}{\bullet}$ & & subset 1  \\ 
			ODT \hfill& 60\% green dots $\textcolor[rgb]{0.352,0.553,0.220}{\bullet}$& & subset 123  &\hfill 40\% green dots $\textcolor[rgb]{0.352,0.553,0.220}{\bullet}$& & subset \textcolor{red}{4}   \\
			CST \hfill & 60\% green dots $\textcolor[rgb]{0.352,0.553,0.220}{\bullet}$& & subset 123   &\hfill 40\% green dots $\textcolor[rgb]{0.352,0.553,0.220}{\bullet}$& & subset 1   \\
			CDT \hfill & 60\% green dots $\textcolor[rgb]{0.352,0.553,0.220}{\bullet}$& & subset 123     &\hfill 40\% green dots $\textcolor[rgb]{0.352,0.553,0.220}{\bullet}$& & subset \textcolor{red}{4} \\
			OWT-$ 1 $ \hfill &\hfill Whole dots $\textcolor[rgb]{0.352,0.553,0.220}{\bullet} \textcolor[rgb]{1, 0, 0}{\bullet}$& & subset 123 &\hfill Whole dots $\textcolor[rgb]{0.352,0.553,0.220}{\bullet} \textcolor[rgb]{1, 0, 0}{\bullet}$ & & subset 1 \\
			CWT-$ 1 $  \hfill&\hfill Whole dots $\textcolor[rgb]{0.352,0.553,0.220}{\bullet} \textcolor[rgb]{1, 0, 0}{\bullet}$& & subset 123 &\hfill Whole dots $\textcolor[rgb]{0.352,0.553,0.220}{\bullet}$$\textcolor[rgb]{1, 0, 0}{\bullet}$& & subset 1 \\
			OWT-$ 2 $  \hfill&\hfill Whole dots $\textcolor[rgb]{0.352,0.553,0.220}{\bullet} \textcolor[rgb]{1, 0, 0}{\bullet}$& & subset 123 &\hfill Whole dots $\textcolor[rgb]{0.352,0.553,0.220}{\bullet}$$\textcolor[rgb]{1, 0, 0}{\bullet}$& & subset \textcolor{red}{4} \\
			CWT-$ 2 $  \hfill&\hfill Whole dots $\textcolor[rgb]{0.352,0.553,0.220}{\bullet} \textcolor[rgb]{1, 0, 0}{\bullet}$& & subset 123 &\hfill Whole dots $\textcolor[rgb]{0.352,0.553,0.220}{\bullet}$$\textcolor[rgb]{1, 0, 0}{\bullet}$& & subset \textcolor{red}{4} \\
			
			\bottomrule
		\end{tabular}
		\begin{tablenotes}
			\footnotesize
			\item[1] By default, 70\% of the target domain dataset is used for training, and the remaining 30
		\end{tablenotes}
	\end{threeparttable}
\end{table}
 
\subsubsection{Classroom Scenario}Figure \ref{Scenario_2_case} illustrates the second experimental setup, a 14.5 m $ \times  $ 9 m classroom within the China Wireless Valley office building in Nanjing, comprising 35 RPs. The indoor environment is furnished with numerous chairs and desks, with the AP positioned below these furnishings, thereby classifying the test area as a NLOS environment. Consistent with the office scenario, data collection was conducted at a uniform height across all RPs. At each RP, five distinct human postures—including standing, walking, squatting, and multiple individuals standing—were recorded, resulting in five separate tasks, with RPs spaced 2 m apart.

\subsubsection{Test Case}
The following nomenclature is used to denote the test cases in the two test scenarios: 1) Office Same Test (OST); 2) Office Different Test (ODT); 3) Classroom Same Test (CST); 4) Classroom Different Test (CDT); 5) Office Whole Test (OWT); 6) Classroom Whole Test (CWT). The training and test sets for each case are outlined in Table \ref{tab1tab1}.
 
\subsection{Baseline Models/Methods}
To evaluate the performance of DF-Loc, we benchmarked it against state-of-the-art fingerprint-based indoor localization models/methods, including end-to-end approaches such as ML techniques like KNN, RF regression (RFR), and support vector regression (SVR), as well as DL methods like CiFi and Hi-Loc. Additionally, we compared DF-Loc with TL methods such as TCA and JDA. 
\begin{itemize}
	\item KNN \cite{youssef2005horus}: In the signal space, we employ the Euclidean distance metric and select the $ K=10 $ nearest RPs for the test point. The estimated location is then determined by averaging the positions of the selected RPs.
	
	\item RFR \cite{wang2018wifi}: The number of decision trees in the ensemble was set to 200, and the number of leaf nodes for each tree was set to 5.
	
	\item SVR \cite{jondhale2022support}: The optimal parameters $ c $ and $ g $ for the support vector machine (SVM) model were determined using particle swarm optimization (PSO) within the range of $ \left[ -8,8 \right] $. The kernel function employed was the standard Gaussian radial basis function (RBF).
	
	\item CiFi \cite{8468057}: The dataset and training strategy were kept consistent with those used in DF-Loc, and the network hyperparameters were adopted from \cite{8468057}. For OST and CST, data from the same environment were used for both training and testing. For ODT and CDT, data from a historical environment served as the training set, while data collected in a new environment were used for testing.
	
	\item Hi-Loc \cite{ruan2022hi}: We followed the same comparison strategy as CiFi.
	
	\item TCA \cite{yin2021localization}: We selected RFR as the prediction function from among KNN, RFR, and SVR. The kernel function employed was the standard Gaussian RBF, and the root mean squared error (RMSE) was used as the loss function.  The remaining settings were consistent with those in \cite{yin2021localization}.
	
	\item JDA \cite{guo2019transferred}: Similarly, we selected SVR as the prediction function.  The kernel function was the standard Gaussian RBF, and RMSE was used as the loss function. The remaining settings were kept consistent with \cite{guo2019transferred}.
\end{itemize}
The CNN in DF-Loc employed filter sizes $ N_f $ of 32, with kernel sizes $ N_k $ of 3 and 7. The initial learning rate $ \alpha  $ for the SGD optimizer was set to 0.002, and batch sizes $ B $  of 40 and 70 were used. The testing RP ratio $ \gamma$ is set to 0.4. A detailed discussion of these parameter settings is provided in Section \ref{Parameters}.

\begin{figure}[!t]
	\centerline{\includegraphics[width=1.8in]{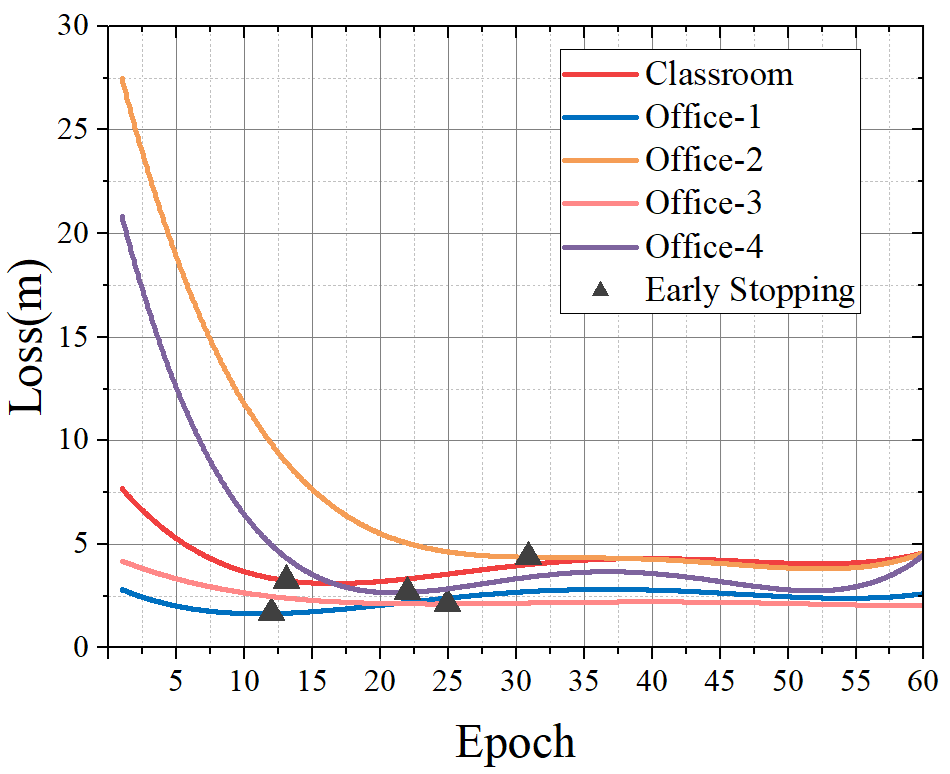}}
	\caption{Training loss for the office and Classroom experiments.}
	\label{fig_Training loss}
\end{figure}
\begin{figure}[!t]
	\centering
	\subfigure[]{\includegraphics[width=1.5in]{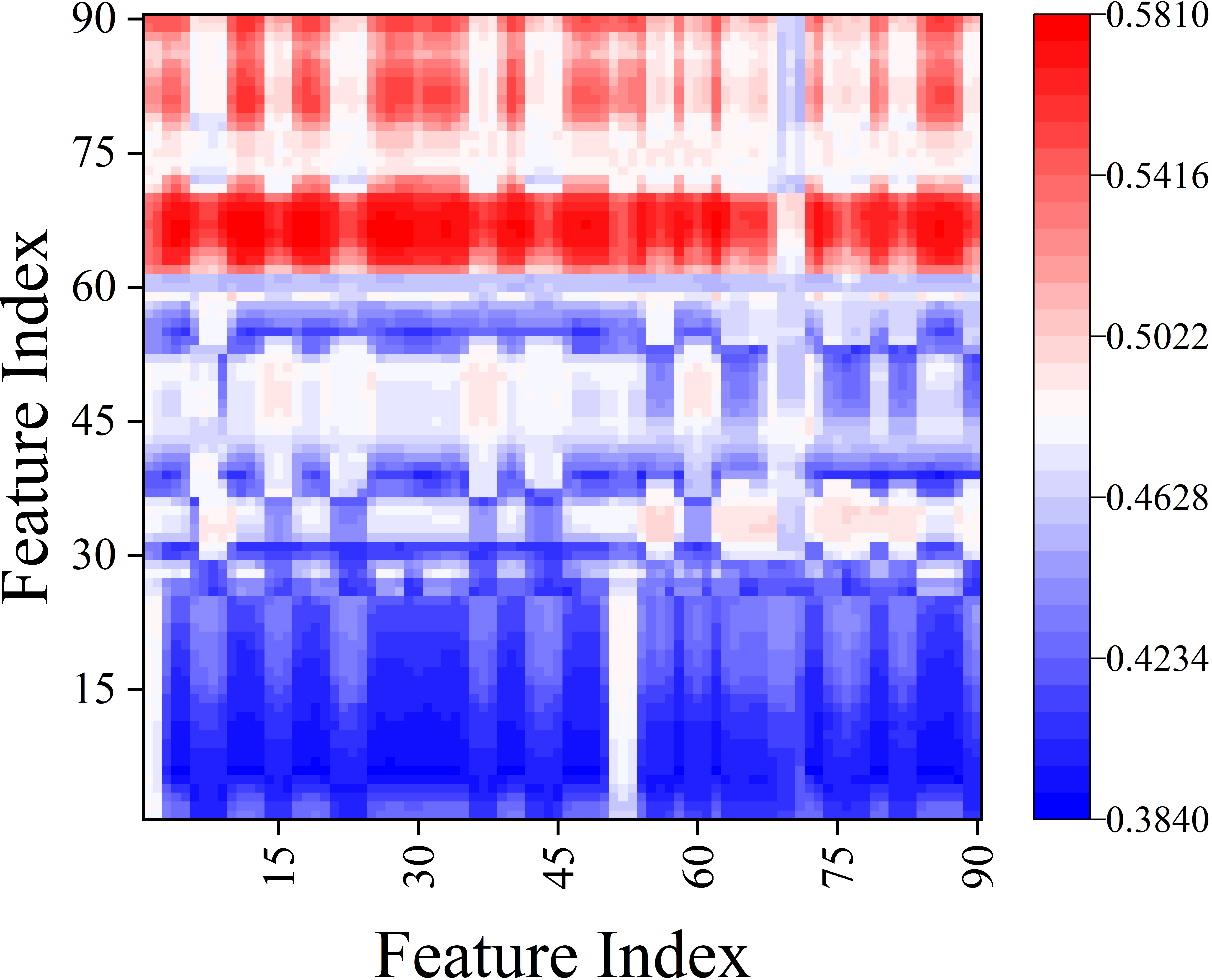}%
		\label{fig_Attension_first_case}}
	\hfil
	\subfigure[]{\includegraphics[width=1.5in]{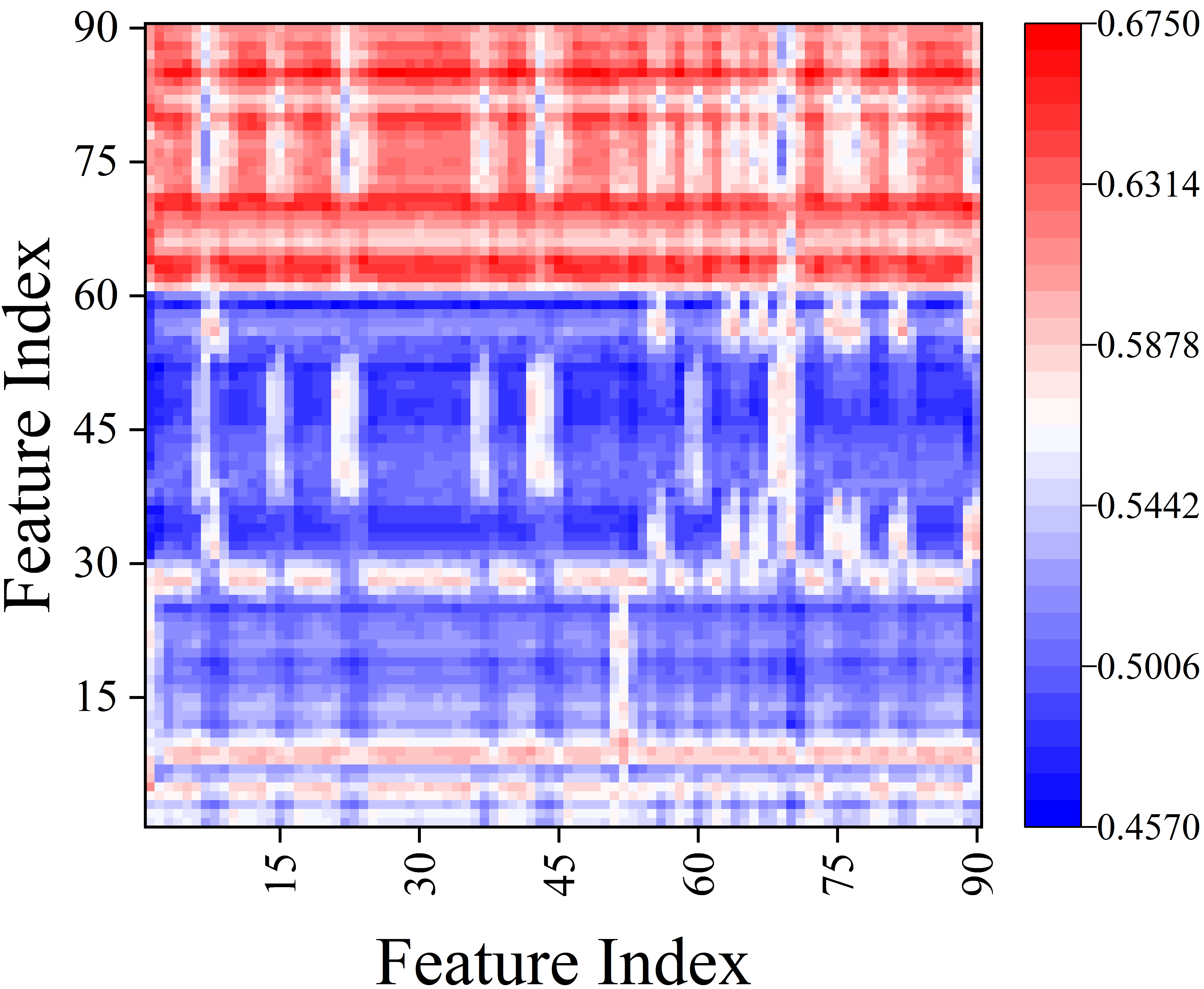}%
		\label{fig_Attension_second_case}} 
	\caption{Visualization of attention weights. (a) Contextual attention weights from the CNN output. (b) Attention weights from the dual information stream fusion.}
	\label{fig_Attension}
\end{figure}

\subsection{Performance Metrics}
To assess the precision and consistency of indoor localization, a comprehensive set of ML evaluation metrics is employed. The Mean Euclidean Distance (MED) serves as the principal performance indicator, defined as
\begin{equation}
\text{MED}=\frac{1}{N}\sum_{i=1}^{N}{\lVert \hat{L}\left( x\left( i \right) ,y\left( i \right) \right) -L\left( x\left( i \right) ,y\left( i \right) \right) \rVert _2}
\end{equation} 
where $ N $ denotes the total number of online testing samples, and $ \hat{L}\left( x\left( i \right) ,y\left( i \right) \right) $ and $ L\left( x\left( i \right) ,y\left( i \right) \right) $ represent the estimated and true 2-D coordinates for the $ i $-th sample, respectively. Additionally, the cumulative distribution function (CDF) is utilized for the statistical analysis of fingerprint positioning errors. Specifically, the $ 1-\sigma  $ (68.27\%) and $ 2-\sigma $ (95.45\%) intervals correspond to the proportions of positioning errors falling within these respective error margins. The subsequent Sections will evaluate the results based on these performance metrics.

\begin{figure*}[!t]
	\centering
	\subfigure[]{\includegraphics[width=1.55in]{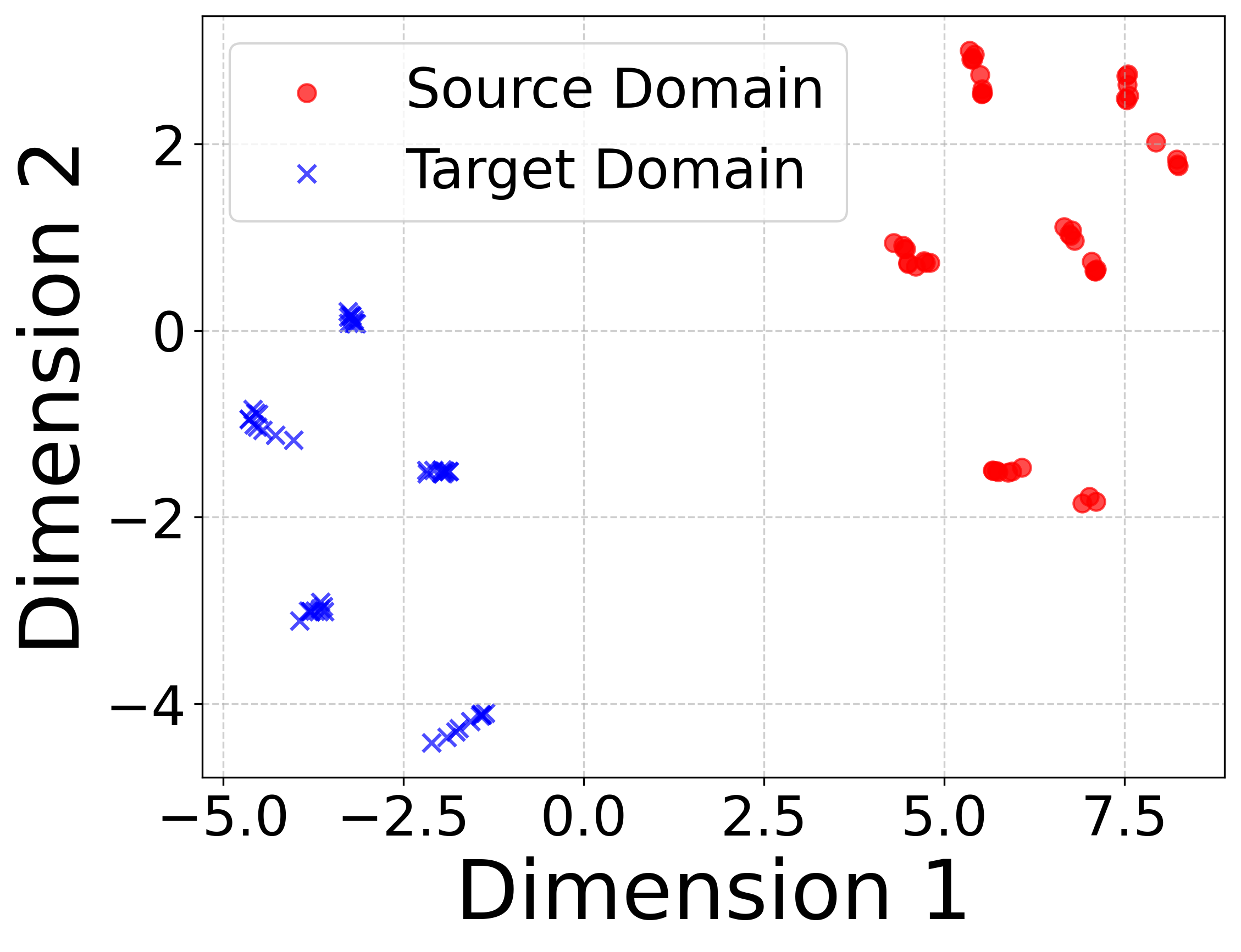}%
		\label{fig_adapted_None_case}}
	\hfil
	\subfigure[]{\includegraphics[width=1.55in]{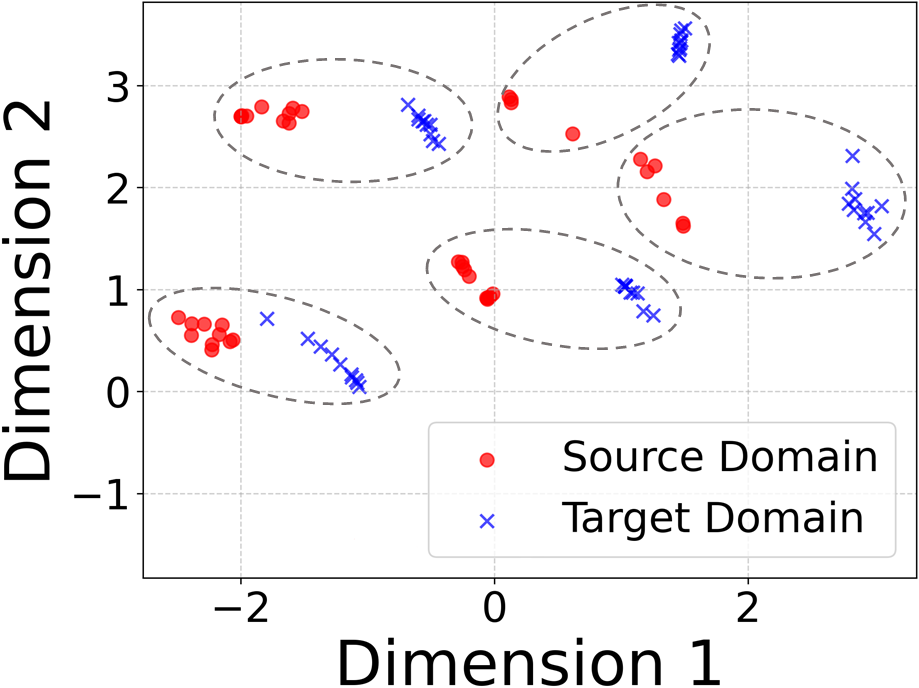}%
		\label{fig_adapted_first_case}}
	\hfil
	\subfigure[]{\includegraphics[width=1.55in]{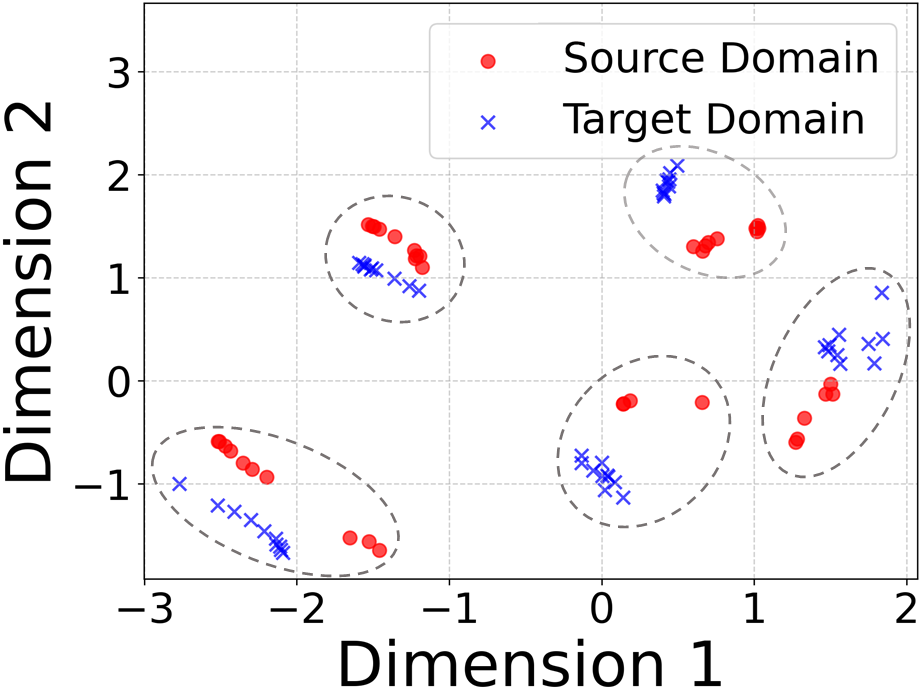}%
		\label{fig_adapted_second_case}} 
	\hfil
	\subfigure[]{\includegraphics[width=1.55in]{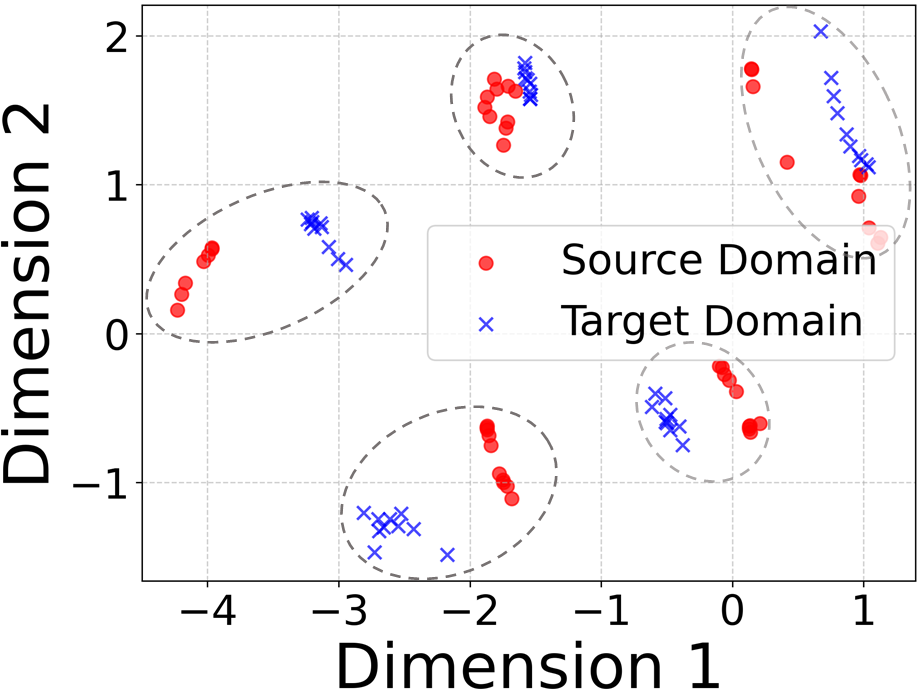}%
		\label{fig_adapted_third_case}}
	\caption{Illustration of DF-Loc for Domain Adaptation (Different ellipses represent different categories). (a) No domain adaptation. (b) Domain adaptation was performed using data from source 1.(c) Domain adaptation was performed using data from source 2.(d) Domain adaptation was performed using data from source 3.}
	\label{fig_adapted}
\end{figure*} 
\begin{figure}[!t]
	\centering
	\subfigure[]{\includegraphics[width=1.42in]{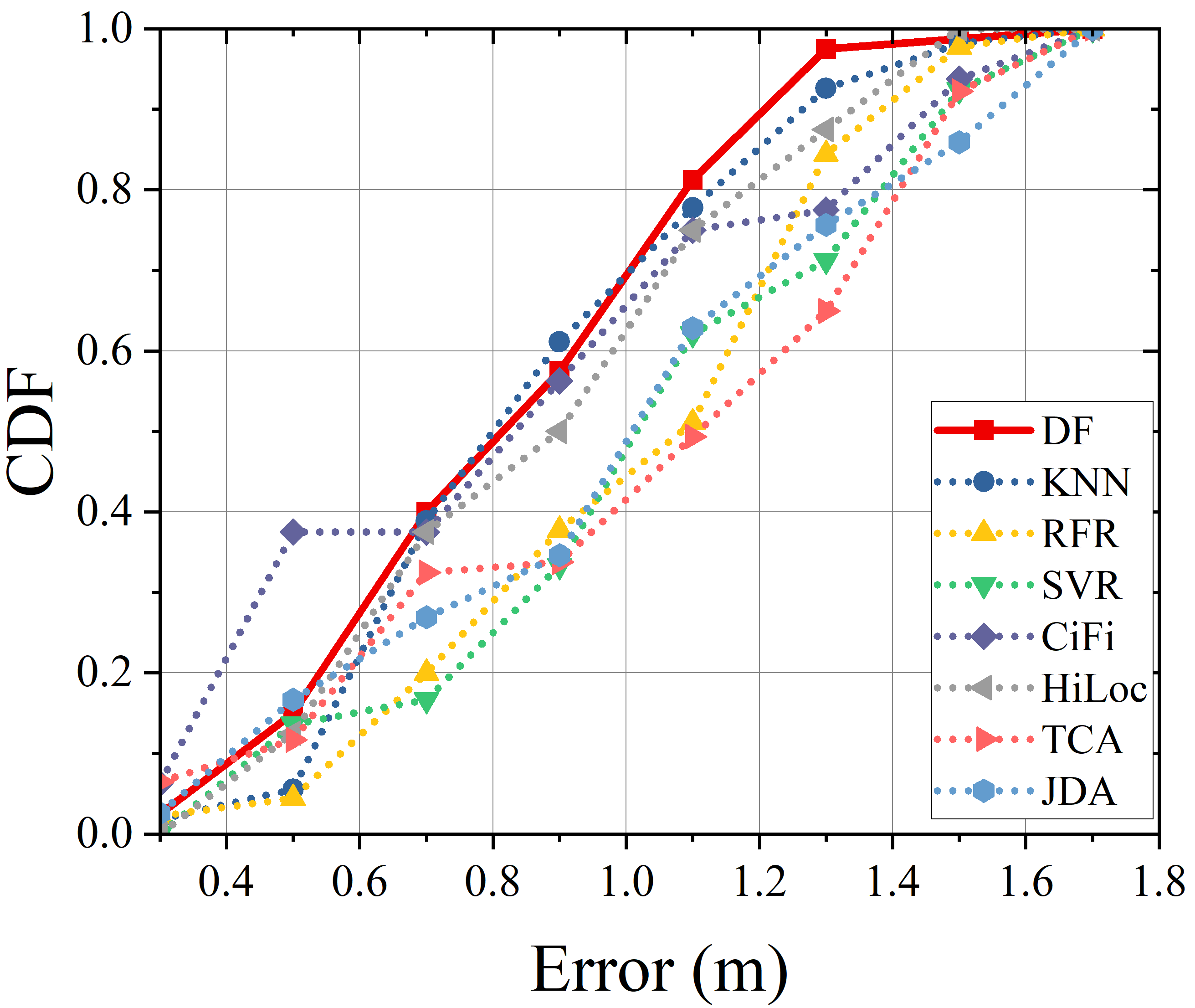}
		\label{fig_compare7_second_case}}
	\hfil 
	\subfigure[]{\includegraphics[width=1.4in]{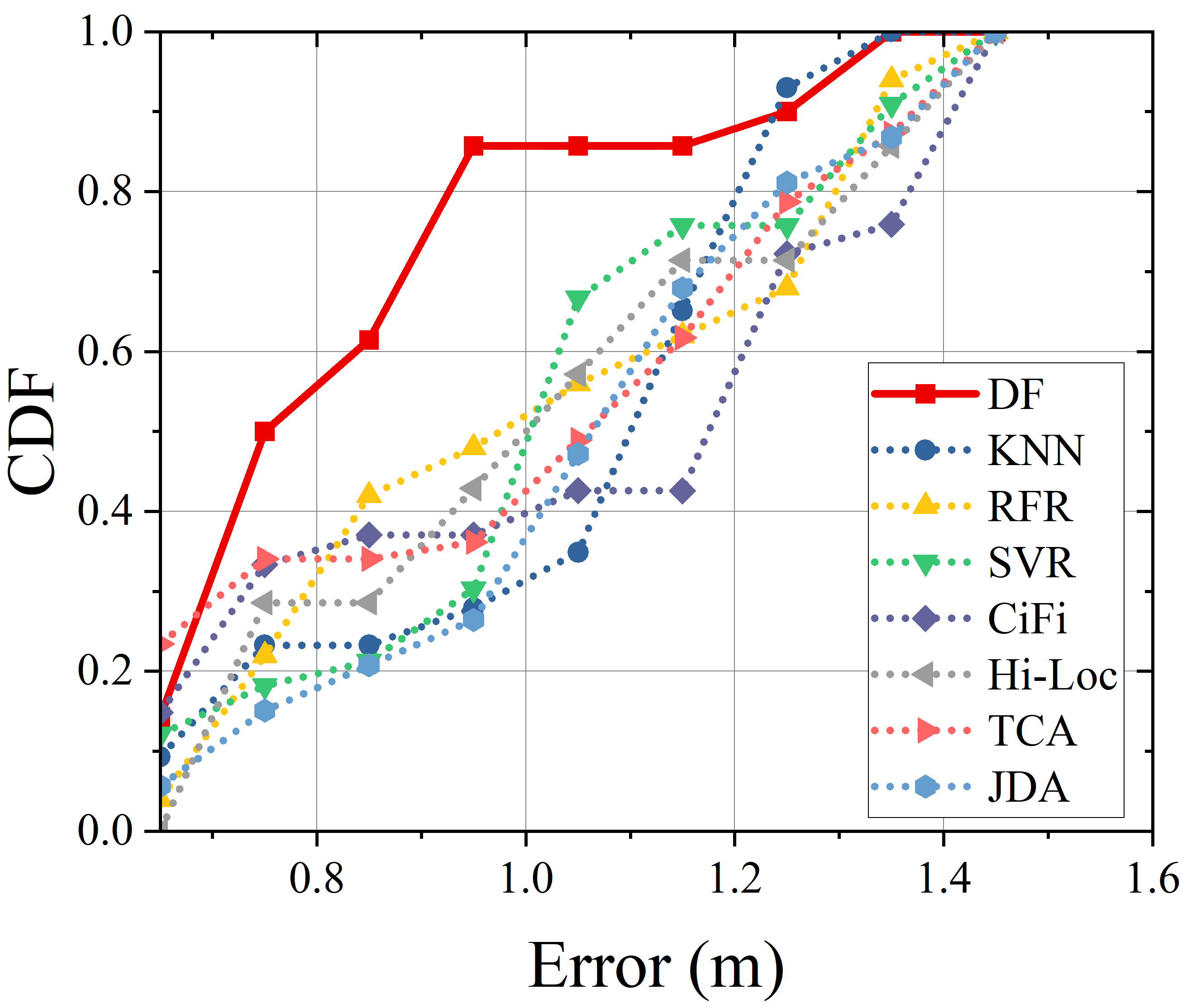}%
		\label{fig_compare7_third_case}}
	\hfil  \\
	\subfigure[]{\includegraphics[width=1.42in]{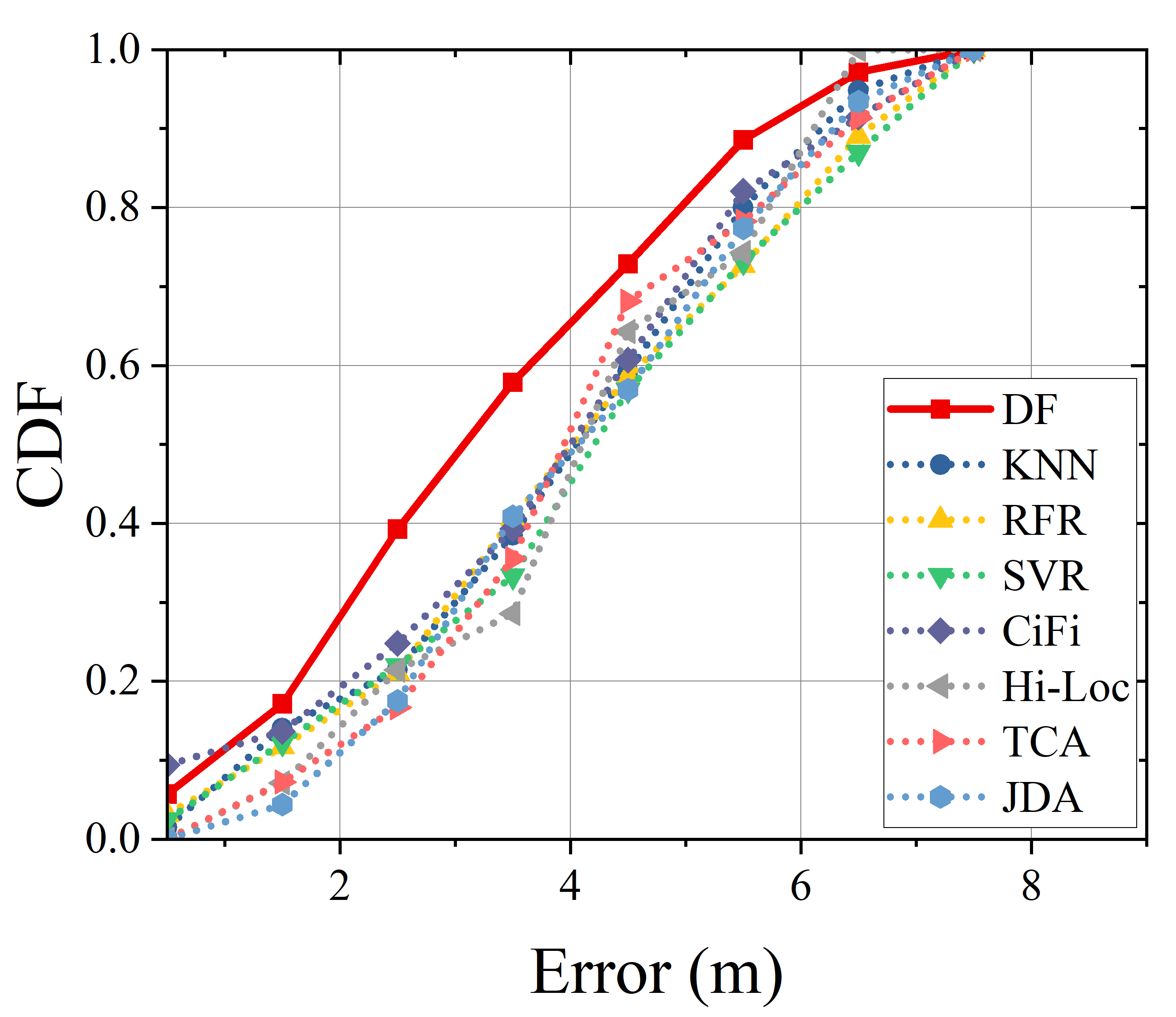}
		\label{fig_compare7_fourth_case}}
	\hfil
	\subfigure[]{\includegraphics[width=1.4in]{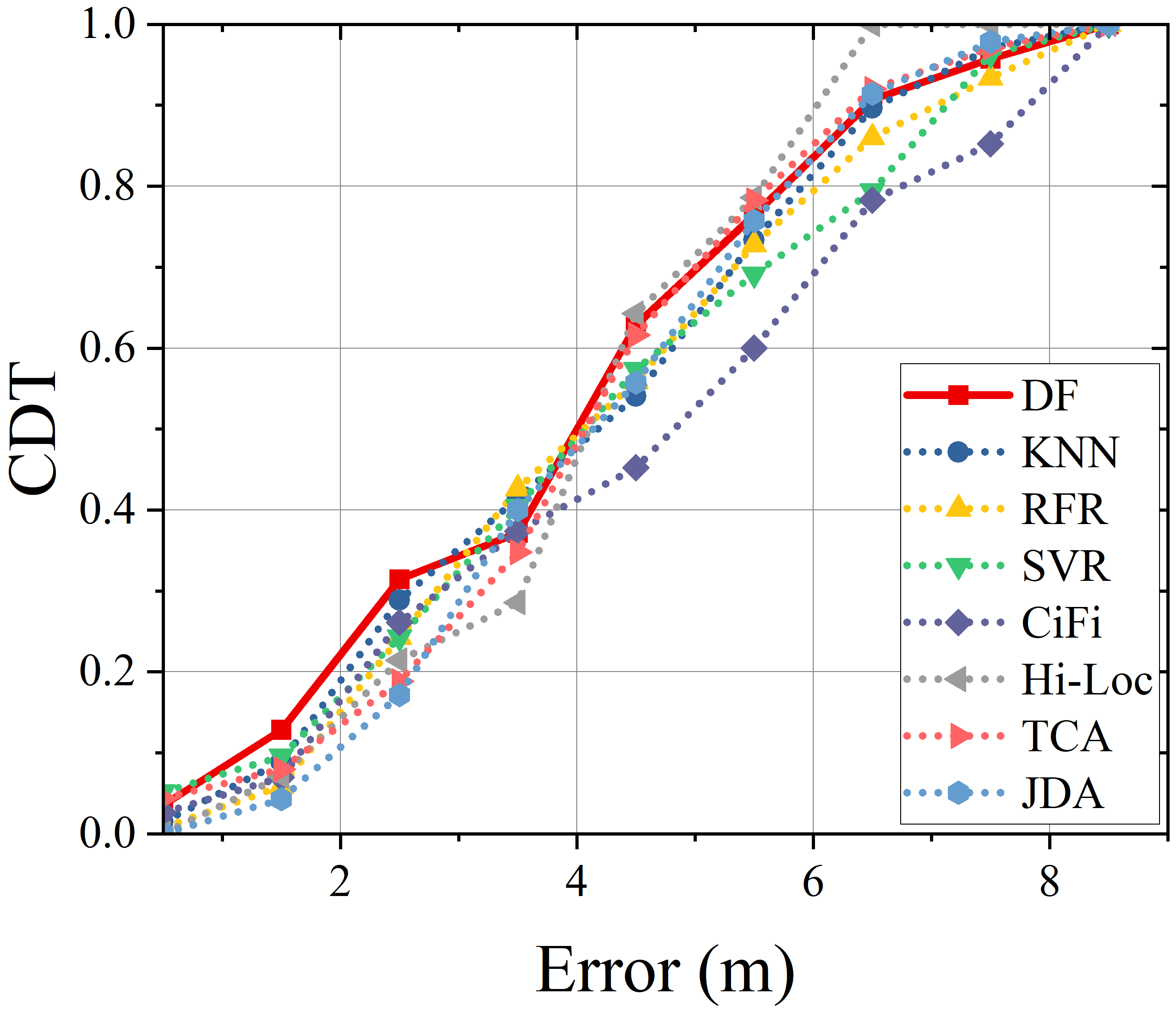}%
		\label{fig_compare7_fifth_case}}
	\caption{CDF of Localization errors with different algorithms. (a) OST case. (b) ODT case. (c) CST case. (d) CDT case.}
	\label{fig_compare7}
\end{figure}

\subsection{Experimental Results} \label{Experimental Results}
To demonstrate the effectiveness of the proposed method, experiments were conducted on OST, ODT, CST, and CDT test cases. The same trained model was used for both OST and ODT in the office scenario, and likewise for CST and CDT in the classroom scenario.
\subsubsection{Training Loss}Figure \ref{fig_Training loss} illustrates the change in the loss function with respect to the epoch during the training of DF-Loc in both the office and classroom scenarios. To prevent overfitting, the maximum epoch was set to 100, and early stopping was implemented with a patience of 10 epochs when the loss ceased to decrease. As depicted in Figure \ref{fig_Training loss}, training in the classroom scenario stopped early at epoch 13 with a loss of 3.5 m. In the four office areas, training stopped early at epochs 12, 31, 25, and 22, with corresponding losses of 2 m, 4.5 m, 1.8 m, and 2.6 m, respectively. The training times for the classroom and office scenarios were 74 s, 68 s, 175 s, 141 s, and 124 s, respectively, with corresponding sample sizes of 7280, 3840, 9920, 8000, and 7040. The testing time per sample in the classroom and office scenarios was 0.05 ms, 0.0381 ms, 0.037 ms, 0.036 ms, and 0.0372 ms, respectively. 

\subsubsection{Attention Weights}Figure \ref{fig_Attension} visualizes the weights of the MS-CAM attention mechanism and the AFF mechanism. In a single iteration, the MS-CAM attention mechanism focuses on the CNN output with larger weights, such as feature indices 60-70 in Figure \ref{fig_Attension_first_case}, representing more important contextual and local information. Similarly, the AFF mechanism focuses on the fused information input with larger weights, such as feature indices 60-90 in Figure \ref{fig_Attension_second_case}. During the iterative training process of the model, DF-Loc focuses more on features with larger weight values, indicating that our attention mechanism facilitates localization performance through weight updates and allocation. 

\subsubsection{Domain Adaptation}To enhance the model's generalization performance on the target domain, our method aims to reduce the distribution discrepancy between the source and target domains and learn domain-invariant feature representations. As illustrated in the figure, after applying our method, the distributions of source and target domain samples in the feature space become more aligned (as observed in Figure \ref{fig_adapted}, the overall overlap of samples from the two domains increases in the three specific feature spaces), indicating that our method effectively achieves marginal distribution alignment. Moreover, the feature representations of the same class in both domains become more similar (as observed in Figure \ref{fig_adapted}, the feature distributions of the same class in both domains become more consistent), demonstrating that our method also achieves effective conditional distribution alignment.

\subsubsection{Positioning Accuracy} Figure \ref{fig_compare7_second_case} shows the CDF of localization errors in the OST case. The $ 1-\sigma $ error of DF-Loc is 0.98 m, while that of the comparison algorithms ranges from 0.971 to 1.322 m. The $ 2-\sigma $ error of DF-Loc is 1.27 m, while that of the comparison algorithms ranges from 1.42 to 1.625 m. In contrast, when the localization error is between 0.69 and 1.46 m, DF-Loc exhibits good performance in terms of accuracy and stability. Several possible explanations exist for this result. On the one hand, MS-ConvBlock, MS-CAM, and AFF make better use of contextual information by extracting richer spatial information and fusing features from different levels, thereby improving localization robustness. On the other hand, the designed preprocessing module eliminates interference data and strengthens the matching degree between effective fingerprints and locations. Although DF-Loc demonstrates excellent performance in most cases, its accuracy is slightly lower than CiFi and KNN when the localization error is less than 0.69 m and around 0.9 m, respectively. This may be because the fingerprint features within a small range are not distinct, making it difficult for DF-Loc to achieve precise localization, which poses a significant challenge.

Figure \ref{fig_compare7_third_case} presents the CDF of localization errors in the ODT case.  DF-Loc achieves a $ 1-\sigma $ error of 0.87 m, while the comparison algorithms exhibit errors ranging from 1.05 to 1.275 m. The $ 2-\sigma $ error of DF-Loc is 1.31 m, compared to 1.29 to 1.425 m for the other algorithms. Differences below 5 cm are considered negligible. Overall, DF-Loc demonstrates superior performance in the ODT case with respect to accuracy and stability, particularly for larger localization errors. This result can be attributed to several factors. Firstly, the multi-source domain adaptation process in DF-Loc enhances the learning of domain-invariant feature representations, leading to improved performance in the target domain. Secondly, MS-ConvBlock, MS-CAM, and AFF enhance the extraction of transferable features. Finally, the preprocessing module plays a positive role in fingerprint matching across different locations.

\begin{figure}[!t]
	\centering
	\subfigure[]{\includegraphics[width=1.6in]{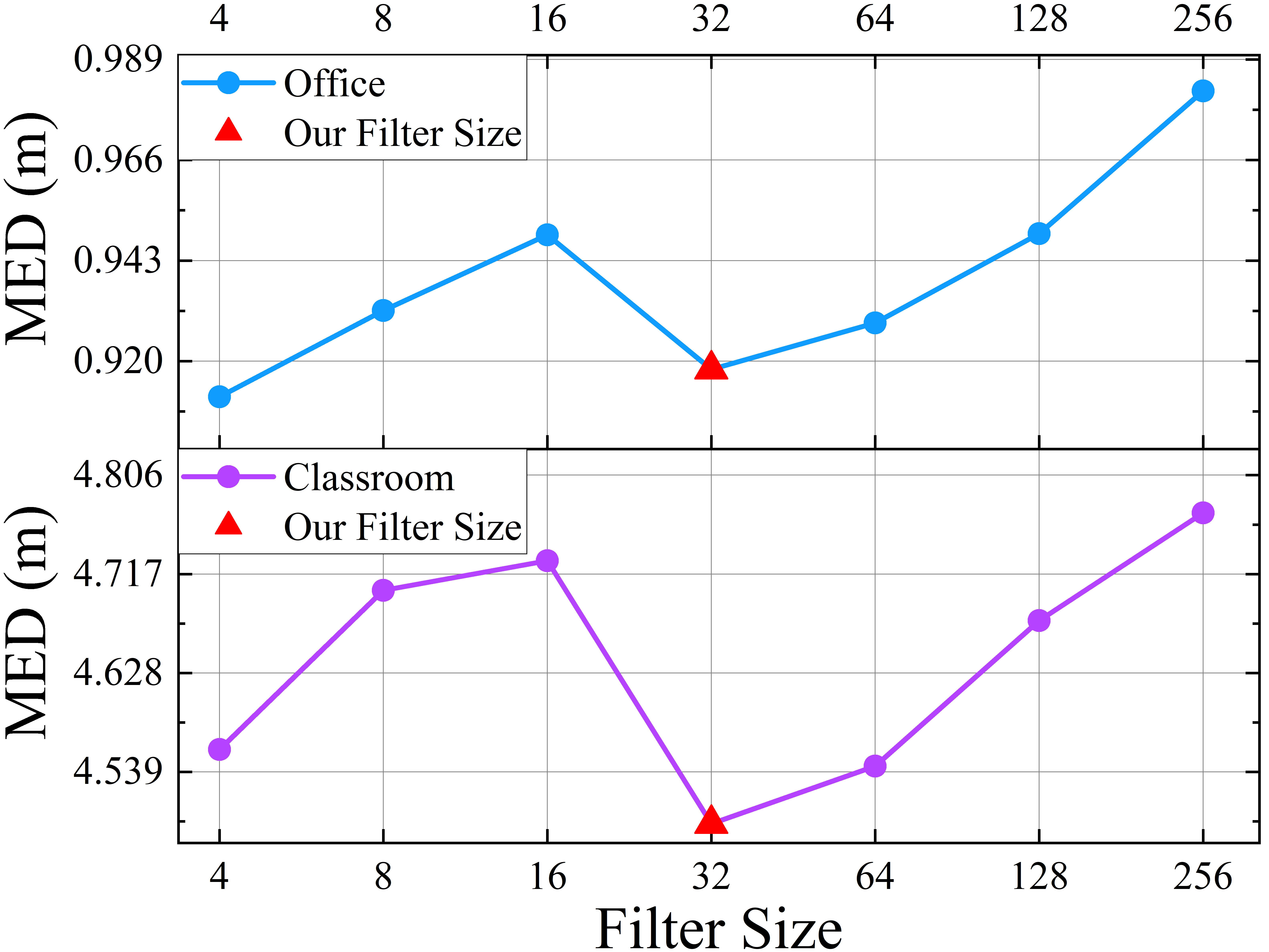}%
		\label{fig_Design_Parameters_2_case}}
	\hfil 
	\subfigure[]{\includegraphics[width=1.75in]{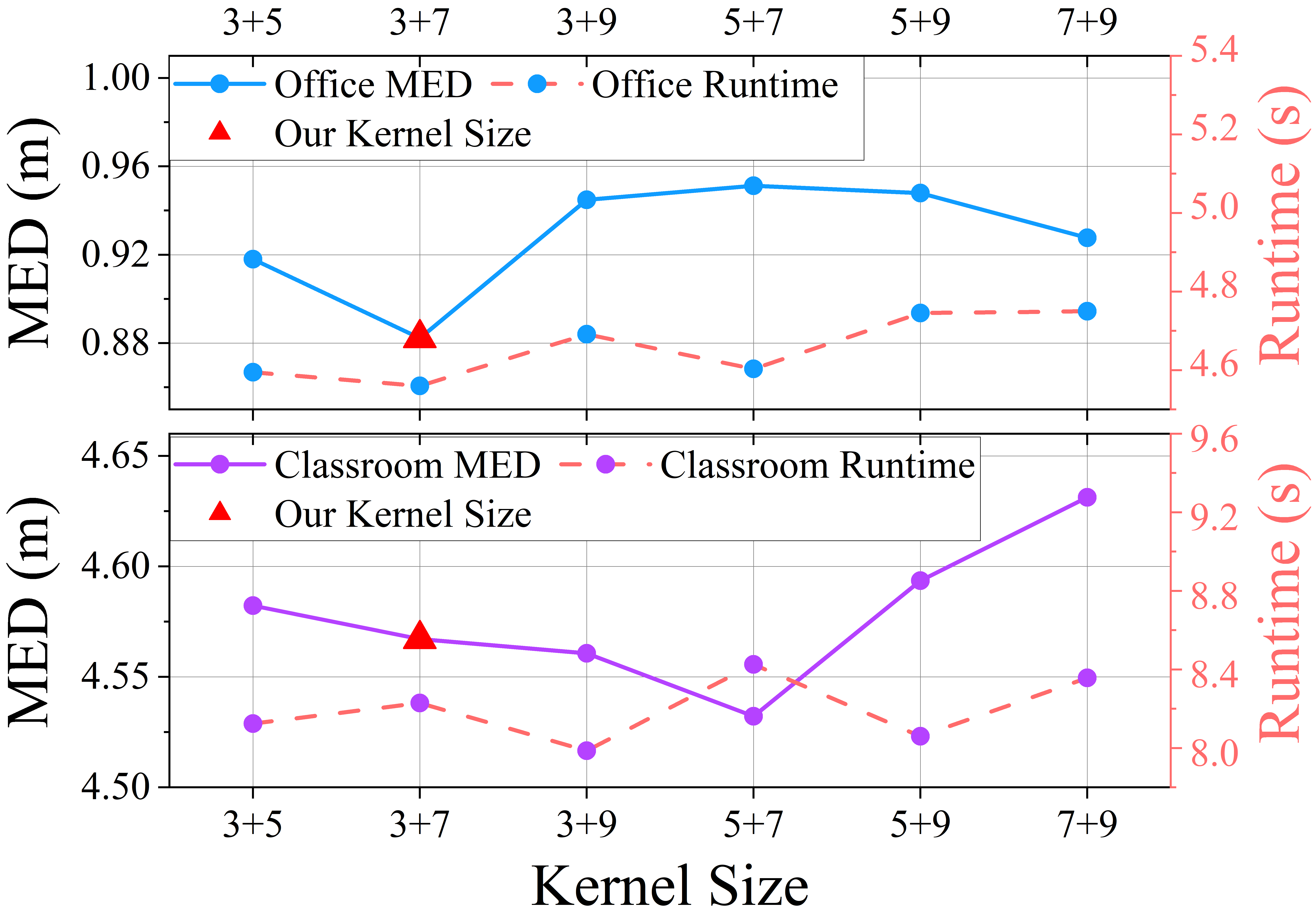}%
		\label{fig_Design_Parameters_3_case}}
	\hfil   
	\subfigure[]{\includegraphics[width=1.6in]{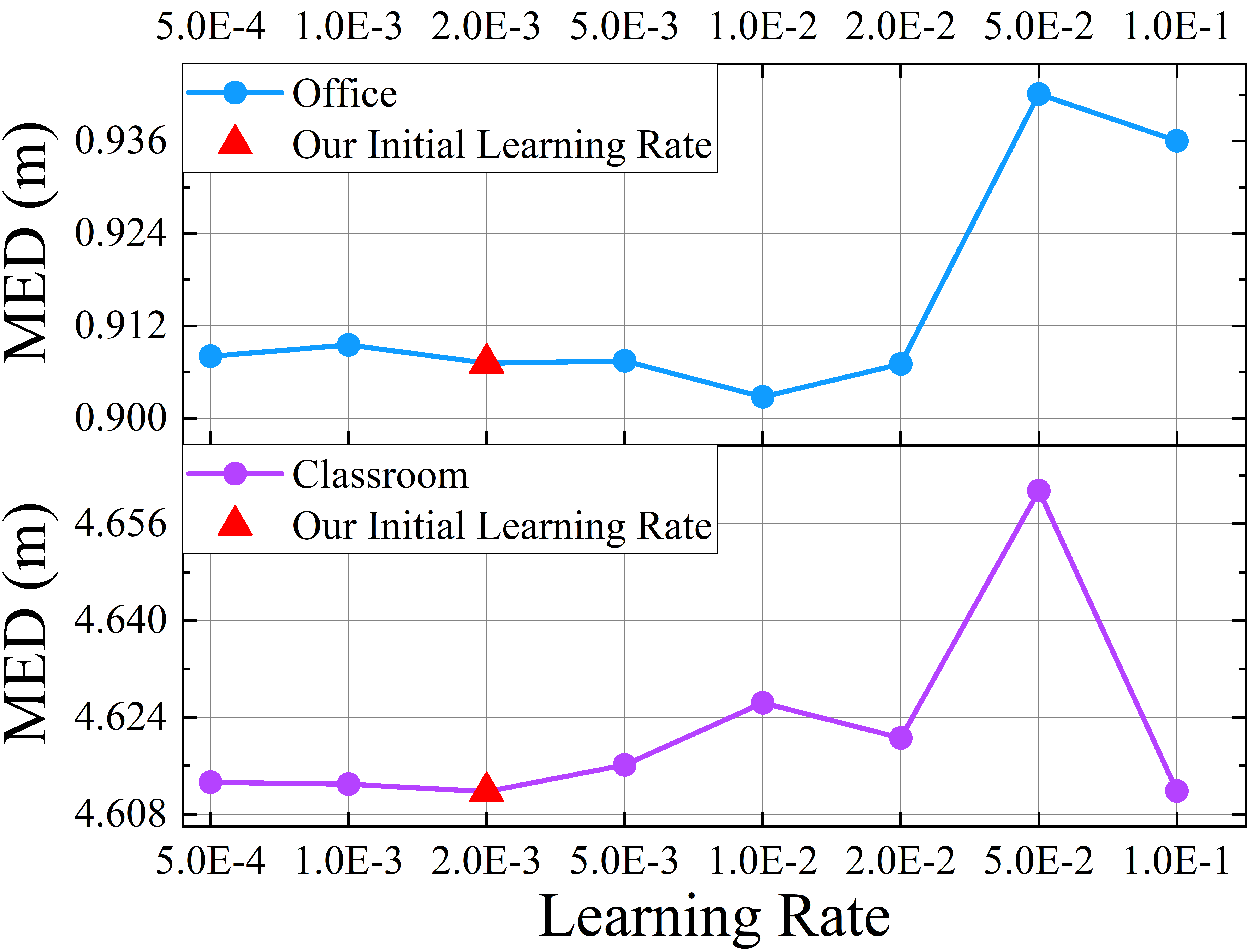}%
		\label{fig_Design_Parameters_5_case}}
	\hfil 
	\subfigure[]{\includegraphics[width=1.75in]{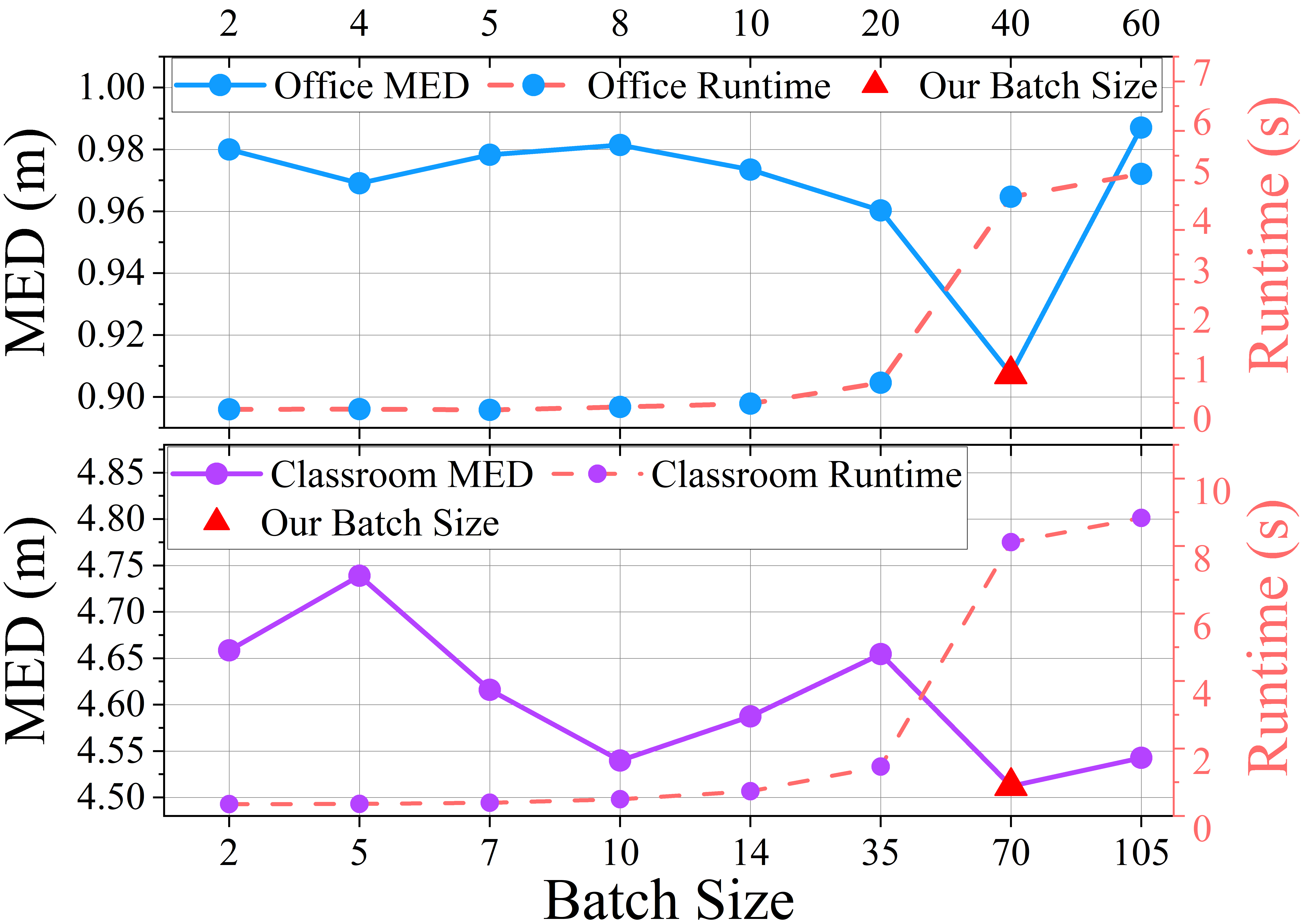}%
		\label{fig_Design_Parameters_6_case}}
	
	\caption{Impact of Design Parameters ($ K = 15, \gamma = 0.4 $).}
	\label{fig_Design_Parameters}
\end{figure}

Figure \ref{fig_compare7_fourth_case} presents the CDF of localization errors in the CST case.  DF-Loc achieves a $ 1-\sigma $ error of 4.15 m, while the comparison algorithms exhibit errors ranging from 4.4 to 5.2 m. The $ 2-\sigma $ error of Hi-Loc is 6.25 m, compared to 6.31 to 7.13 m for the other algorithms.  DF-Loc demonstrates superior localization accuracy compared to the other algorithms. This trend is similar to the OST case, primarily because the training and test sets in both the OST and CST cases have similar data distributions.

Figure \ref{fig_compare7_fifth_case} presents the CDF of localization errors in the CDT case. DF-Loc achieves a $ 1-\sigma $ error of 4.8 m, while the comparison algorithms exhibit errors ranging from 4.7 to 5.93 m. The $ 2-\sigma $ error of DF-Loc is 7.5 m, compared to 6.25 to 8.25 m for the other algorithms. Due to the larger spacing between RPs in this scenario, errors below 0.2 meters are considered negligible. Although DF-Loc performs slightly worse than Hi-Loc at larger errors, it demonstrates superior accuracy and stability within a smaller error range. Several factors may contribute to this result. First, the larger spacing in the classroom scenario leads to lower correlation between fingerprints of different RPs, making fingerprint matching more challenging for distant locations. Second, the target domain data may have a significantly different distribution compared to the source domain data, necessitating a more complex domain adaptation process, which poses a significant challenge for DF-Loc.

Furthermore, the localization accuracy of the OST and ODT test cases in the indoor office scenario is higher than that of the CST and CDT test cases in the classroom scenario. This is primarily attributed to the larger grid spacing and sparser location distribution in the classroom scenario. The trend in results is similar for the OST, CST, and CDT cases, mainly because the training and test sets in these cases are from different locations and exhibit similarities in their training and testing modalities. The ODT case shows a different trend due to the distinct distribution of the target domain data. Additionally, the ODT case exhibits significantly higher accuracy than the OST case, while the CDT case shows slightly lower accuracy than the CST case. This can be attributed to two primary factors: 1) sparser locations lead to reduced fingerprint correlation, resulting in a more negative impact on fingerprint matching in dynamic environments \cite{ruan2022ipos}; and 2) the classroom scenario has a more complex topological spatial structure and target domain data distribution, leading to a more complex domain alignment process.

DF-Loc was compared with several state-of-the-art methods, including KNN, Hi-Loc, and transfer learning approaches such as TCA. DF-Loc consistently outperformed these methods, achieving lower localization errors in both "ST" and "DT" cases. In the "DT" case, transfer learning-based methods exhibited a more pronounced advantage, demonstrating their potential for localization in dynamic environments. Further analysis reveals that the primary sources of error include signal interference from obstacles and variations in human posture, which can affect the stability of fingerprint features. A detailed analysis is provided in the next section.

\begin{figure}[!t]
	\centering
	\subfigure[]{\includegraphics[width=1.4in]{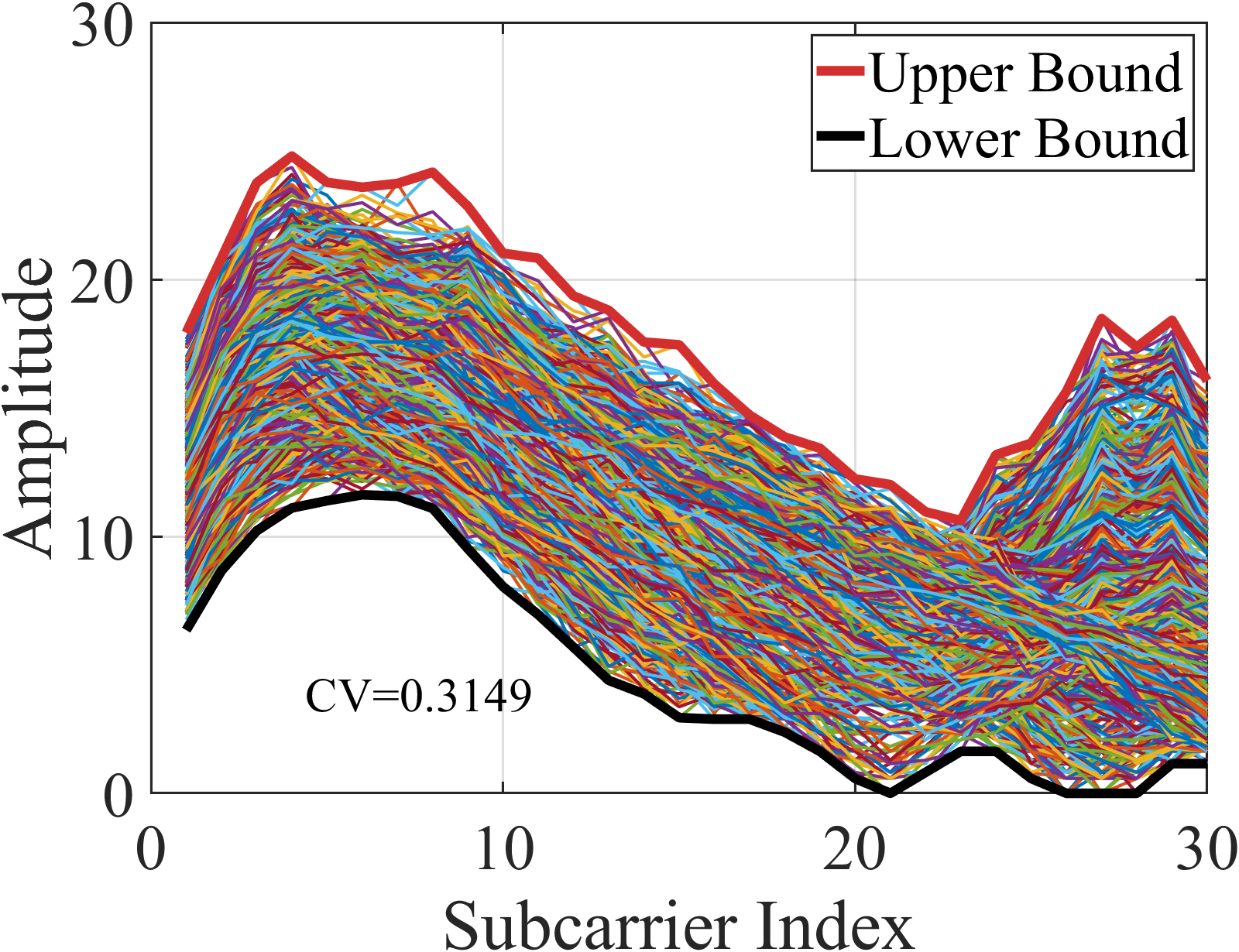}%
		\label{fig_Preprocessing_first_case}}
	\hfil
	\subfigure[]{\includegraphics[width=1.4in]{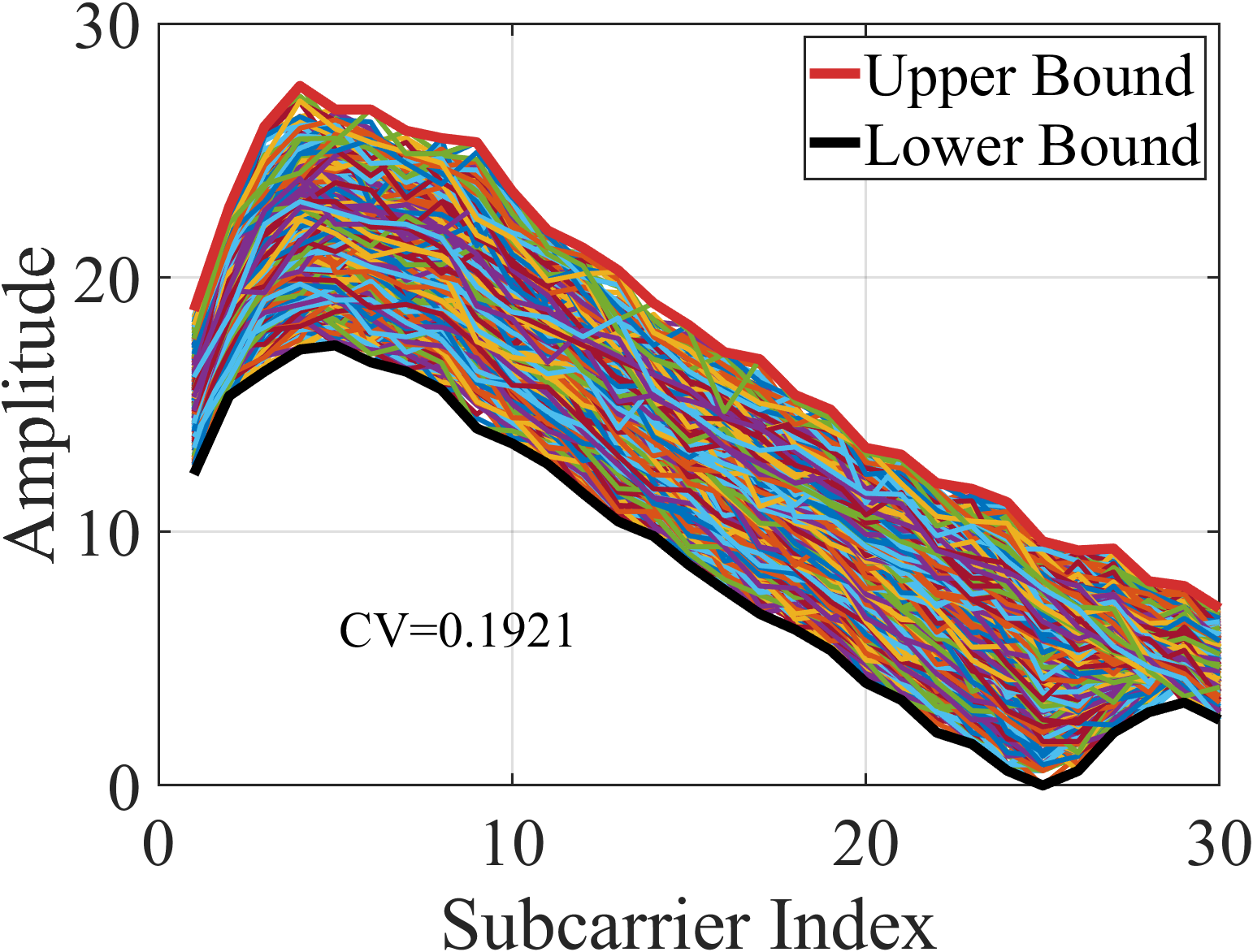}%
		\label{fig_Preprocessing_second_case}}
	\hfil \\
	\subfigure[]{\includegraphics[width=1.41in]{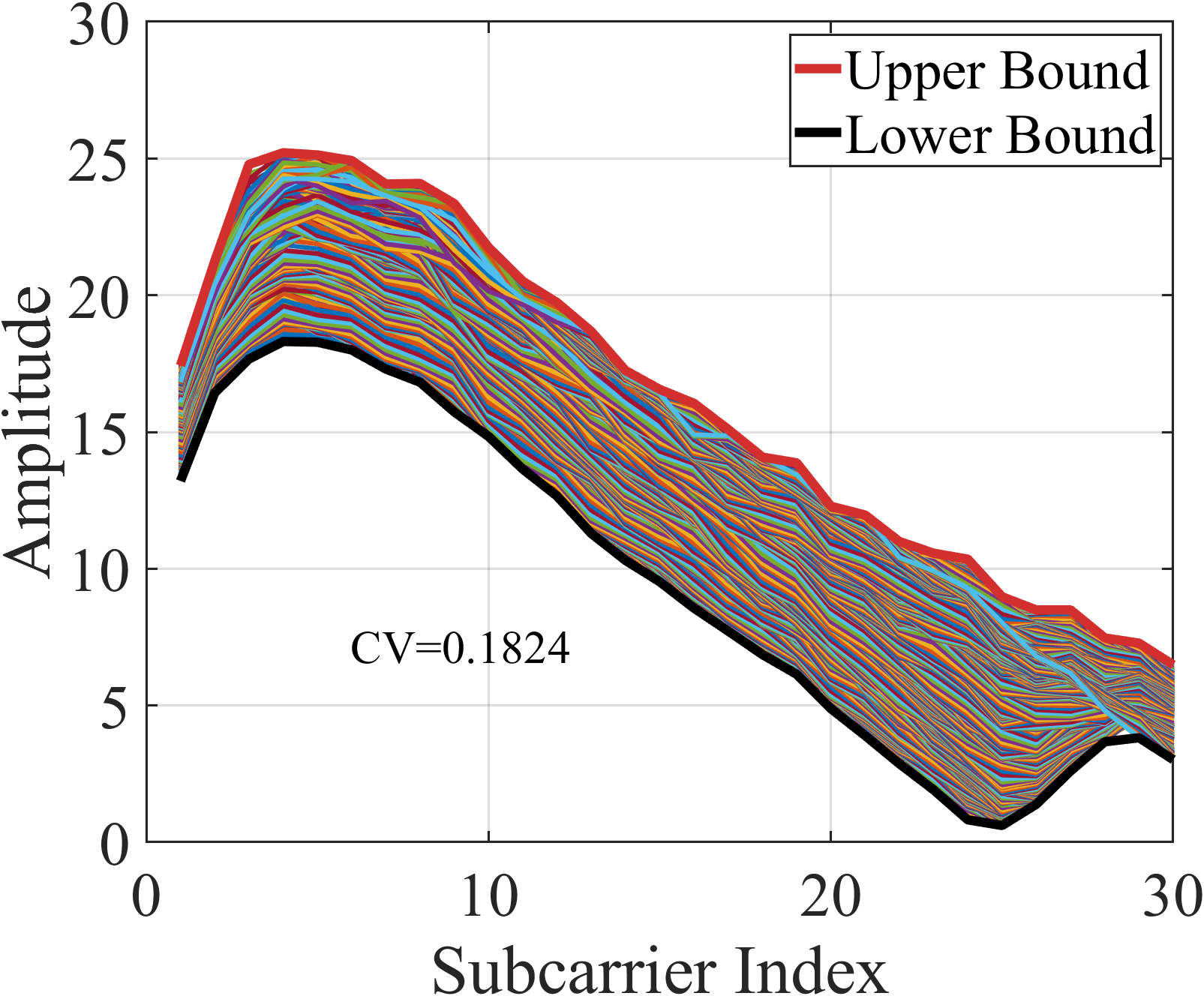}%
		\label{fig_Preprocessing_third_case}}
	\hfil
	\subfigure[]{\includegraphics[width=1.4in]{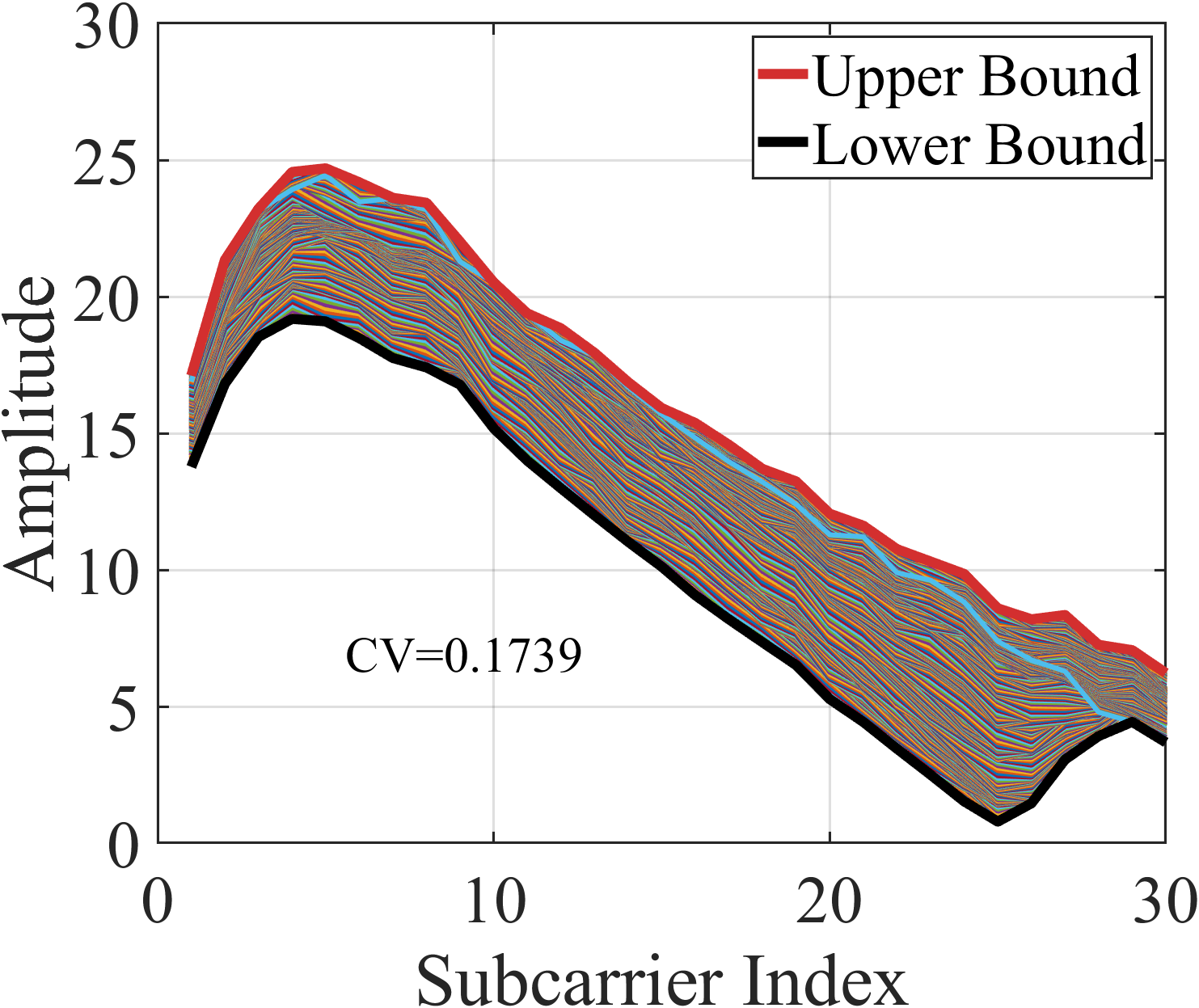}%
		\label{fig_Preprocessing_fourth_case}}
	\caption{Performance of CSI amplitude fingerprint preprocessing based on HWF (1000 packets). (a) Original amplitude. (b) After Hampel filtering. (c) After wavelet filtering. (d) After Butterworth low-pass filtering.}
	\label{fig_Preprocessing}
\end{figure}
\begin{figure}[!t]
	\centering
	\subfigure[]{\includegraphics[width=1.5in]{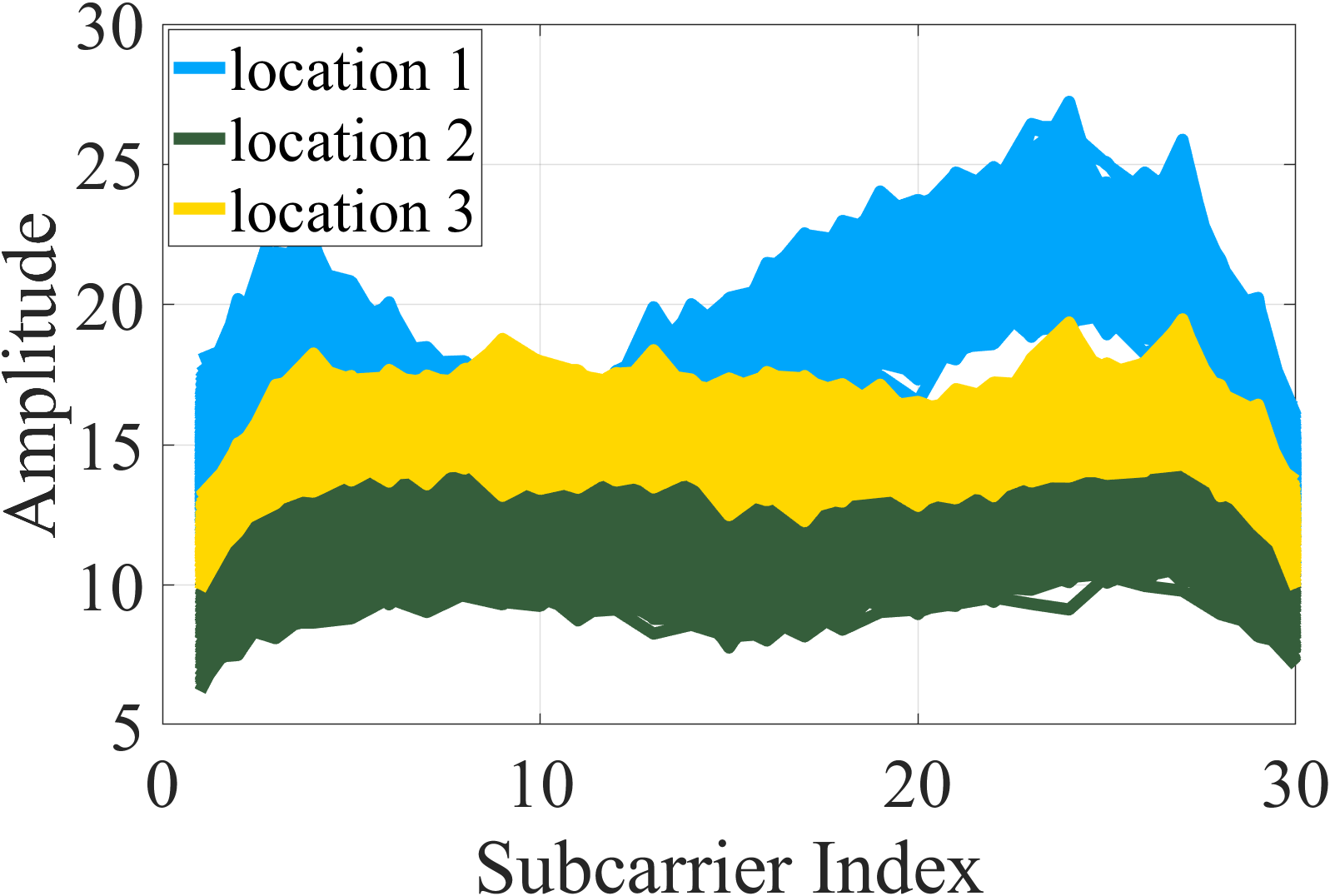}%
		\label{fig_compare_first_case}}
	\hfil
	\subfigure[]{\includegraphics[width=1.55in]{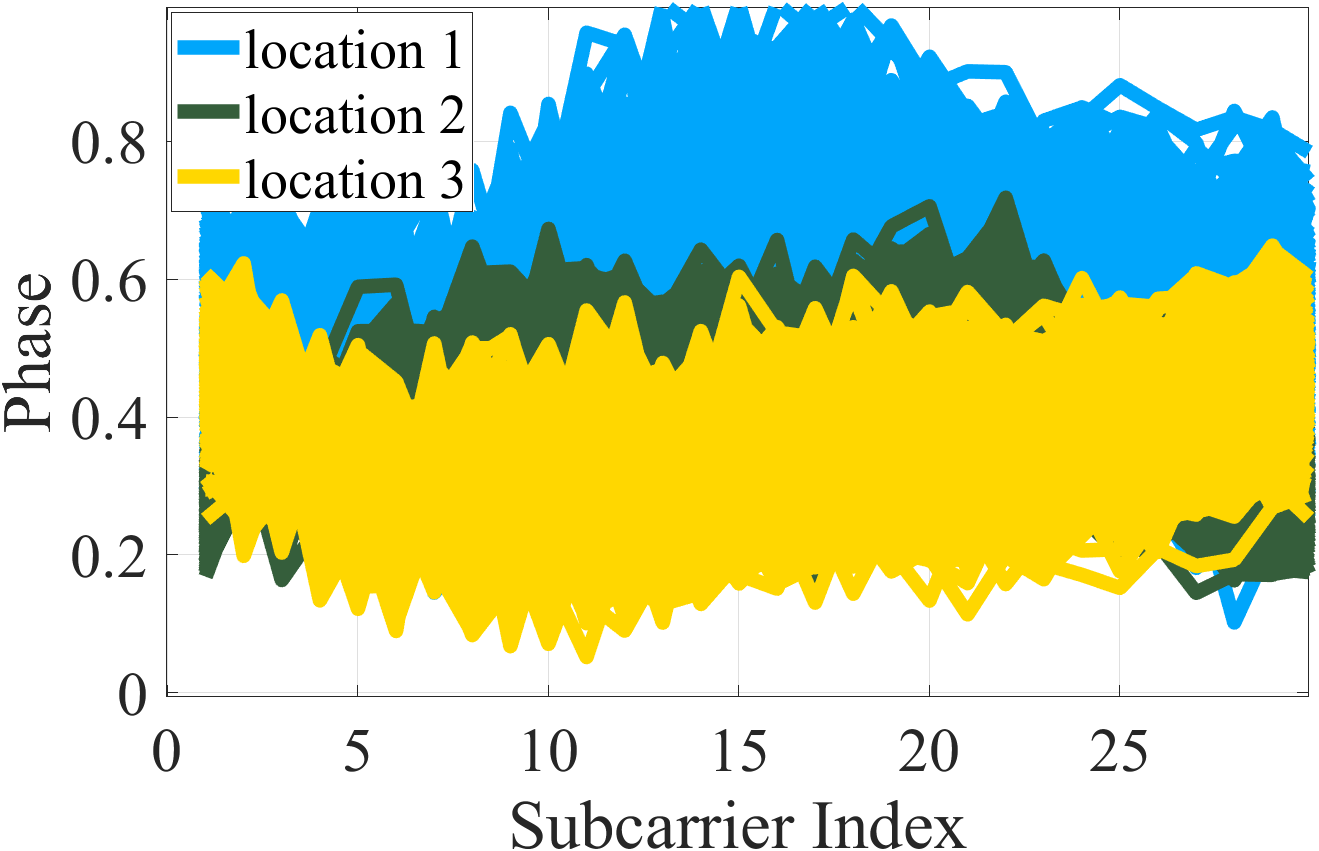}%
		\label{fig_compare_second_case}}
	\hfil \\
	\subfigure[]{\includegraphics[width=1.5in]{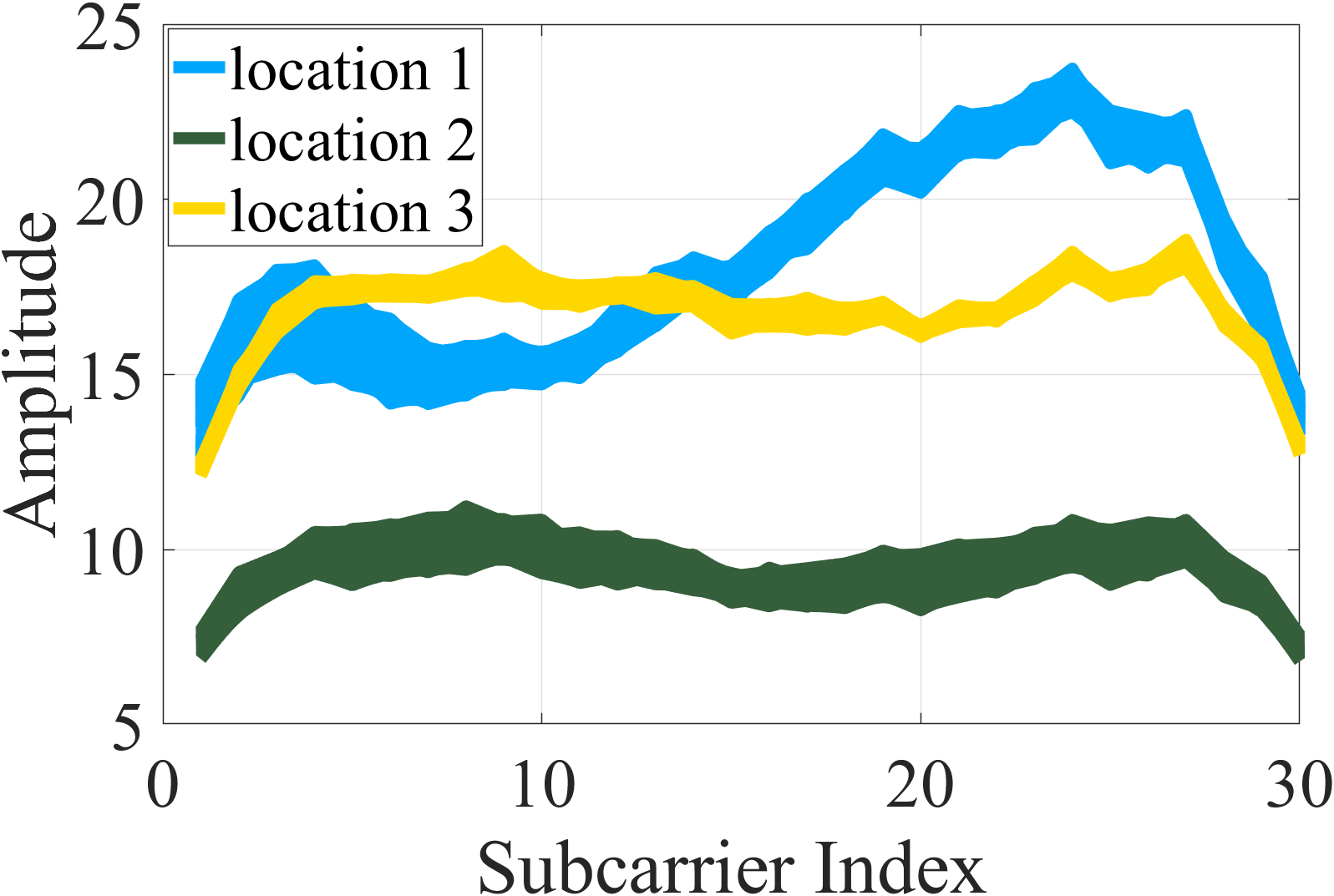}%
		\label{fig_compare_third_case}}
	\hfil
	\subfigure[]{\includegraphics[width=1.55in]{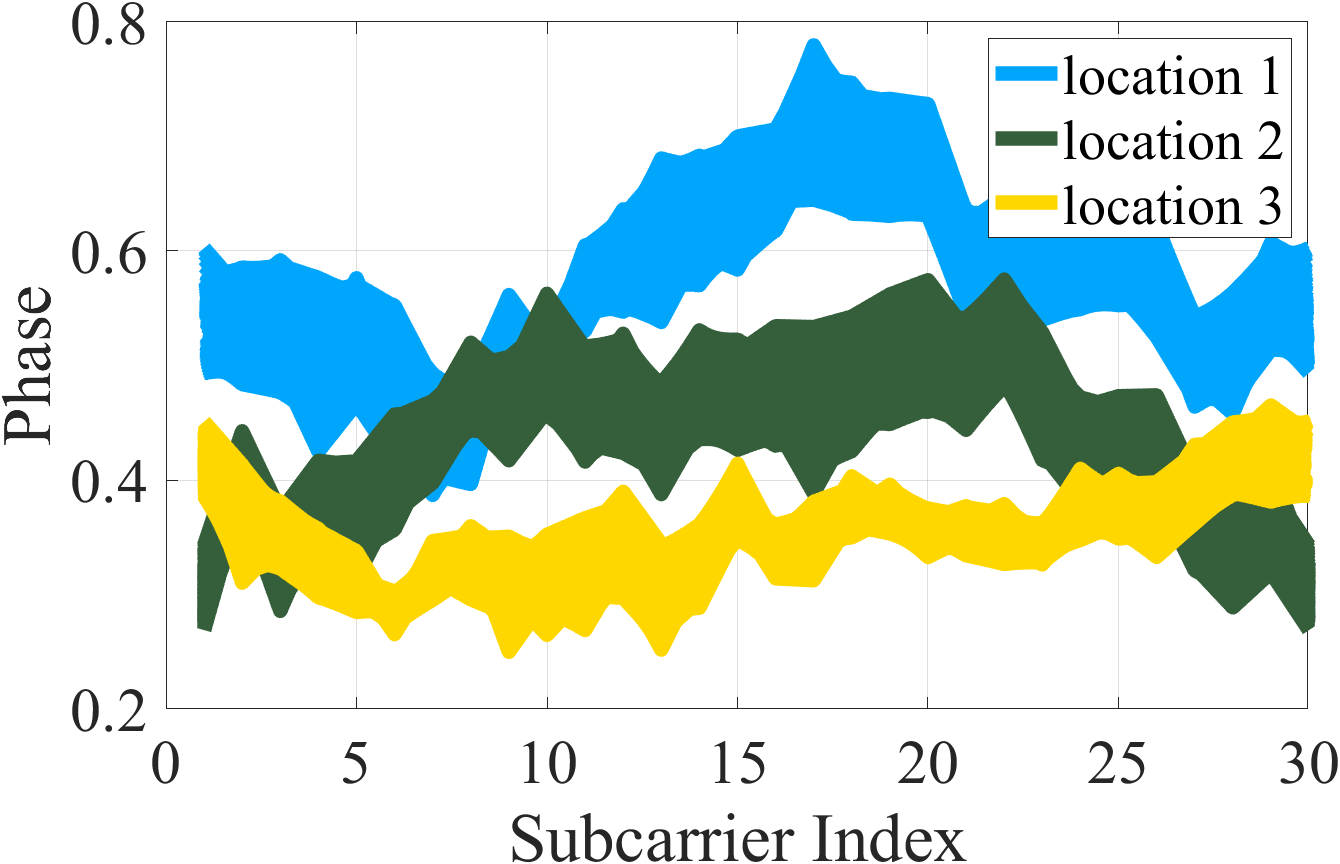}%
		\label{fig_compare_fourth_case}}
	\caption{CSI amplitude and phase fingerprint features of 1000 samples at three different locations in a classroom scenario. (a) Raw amplitude. (b) Raw phase after LC. (c) Amplitude after HWF. (d) Phase after LC and HWF.}
	\label{fig_compare}
\end{figure}

\section{COMMENTS AND DISCUSSION}

\subsection{Impact of Design Parameters} \label{Parameters}
To evaluate the impact of different DF-LocNet model parameters on indoor localization, we focus on the OST and CST test cases. The results presented for each parameter are averaged over 20 trials to minimize errors from individual runs, with consistent training and testing RPs maintained throughout. A controlled variable approach is employed for parameter analysis and discussion.
\subsubsection{Impact of the Filter Size}The filter size determines the number of feature maps in the CNN, which in turn affects the types of features learned and the expressive capacity of the model. To identify a suitable filter size, we trained and tested the model with filter sizes of $ N_f = 2^n $, where $ n\in \left[ 2,8 \right] $. As shown in Figure \ref{fig_Design_Parameters_2_case}, the MED initially increases, then decreases, and finally increases again as $ N_f $ increases, reaching its lowest point at $ N_f $ = 32.  The difference in MED between $ N_f $ = 32 and $ N_f $ = 4 is considered negligible in the office scenario. These results indicate that when the filter size is too small, features relevant for localization are not fully extracted. Conversely, when the filter size is too large, redundant features are extracted, and the improvement in localization performance is limited.

\subsubsection{Impact of the Kernel Size}In a CNN, the kernel size determines the receptive field, which in turn affects the scale and level of detail of the captured features, as well as the network's runtime.  As illustrated in Figure \ref{fig_Design_Parameters_3_case}, the training time gradually increases with increasing $ N_k $, although this trend is not pronounced. However, the MED exhibits a rebound trend. This demonstrates that the MS-ConvBlock design can achieve excellent accuracy while reducing network depth. Considering both MED and runtime, a combination of $ N_k = 3 $ and $ N_k = 7 $ is selected as the kernel size for both scenarios.

\subsubsection{Impact of the Learning Rate}The learning rate controls the step size of model parameter updates, thereby influencing the convergence speed and ultimate performance of the model. As depicted in Figure \ref{fig_Design_Parameters_5_case}, the MED increases rapidly when the initial learning rate $ \alpha $ exceeds 0.02, indicating significant network instability. To balance training stability and efficiency, we select $ \alpha = 0.002 $ as the initial learning rate.

\subsubsection{Impact of the Batch Size}Batch size determines the number of samples used in each parameter update, influencing the stability, speed, and generalization ability of model training. Considering the model's computation rules and the division of training and testing RPs, the batch size $ B $ is preset as a common divisor of the number of training and testing samples. As illustrated in Figure \ref{fig_Design_Parameters_6_case}, the training time increases slowly with increasing B in both scenarios but exhibits rapid growth at $ B = 40 $ and $ B = 70 $, respectively.  The MED shows a decreasing trend with increasing $ B $ in both scenarios, with a rebound observed at $ B = 40 $ and $ B = 70 $, respectively. Therefore, considering both training time and localization accuracy, $ B = 40 $ and $ B = 70 $ are selected as the batch sizes for model training in the two scenarios.

\subsection{Performance of the QC module, Attention mechanism and MUDA}
\subsubsection{Performance of the QC}To illustrate the performance of the HWF-based preprocessing, we use amplitude as an example and randomly select a RP to demonstrate its impact on fingerprint features.  We use the change in the coefficient of variation (CV) to quantify the change in feature stability.  Figure \ref{fig_Preprocessing} shows the HWF module preprocessing procedure at coordinates (4,5) in the classroom scenario. As depicted in Figures \ref{fig_Preprocessing_first_case}-\ref{fig_Preprocessing_fourth_case}, the CV of the amplitude decreases from 0.3149 to 0.1739, a reduction of approximately 44.78\%. Similar performance is observed at other locations, indicating that HWF can effectively ensure the stability of fingerprint features.

Figure \ref{fig_compare} presents examples of CSI fingerprint features for 1000 samples at three different locations in the classroom. Compared to Figures \ref{fig_compare_first_case} and \ref{fig_compare_second_case}, the amplitude and phase after HWF processing exhibit greater stability, as shown in Figures \ref{fig_compare_third_case} and \ref{fig_compare_fourth_case}, respectively. Furthermore, the distinct shapes and positions of the curves demonstrate that the amplitude and phase at different locations possess unique fingerprint characteristics. This highlights that HWF can enhance the discriminability of fingerprints across different locations.

\begin{table}[!t]
	\centering
	\caption{LOCALIZATION PERFORMANCE COMPARISON MODES}
	\label{tab1tab4}
	\begin{tabular}{c|c} 
		\toprule
		Mode  & Description \\
		\midrule
		
		DF &  DF-loc  \\ 
		M1 &  DF-loc only without preprocessing\\ 
		M2 &  DF-loc only without MS-CAM \& AFF\\ 
		M3 &  DF-loc only without MUDA\\
		M4 &  DF-loc without MS-CAM \& AFF and MUDA\\
		M5 &  DF-loc without QC, MS-CAM \& AFF and MUDA \\
		\bottomrule
	\end{tabular}
\end{table}

\begin{figure}[!t]
	\centering
	
	\subfigure[]{\includegraphics[width=1.5in]{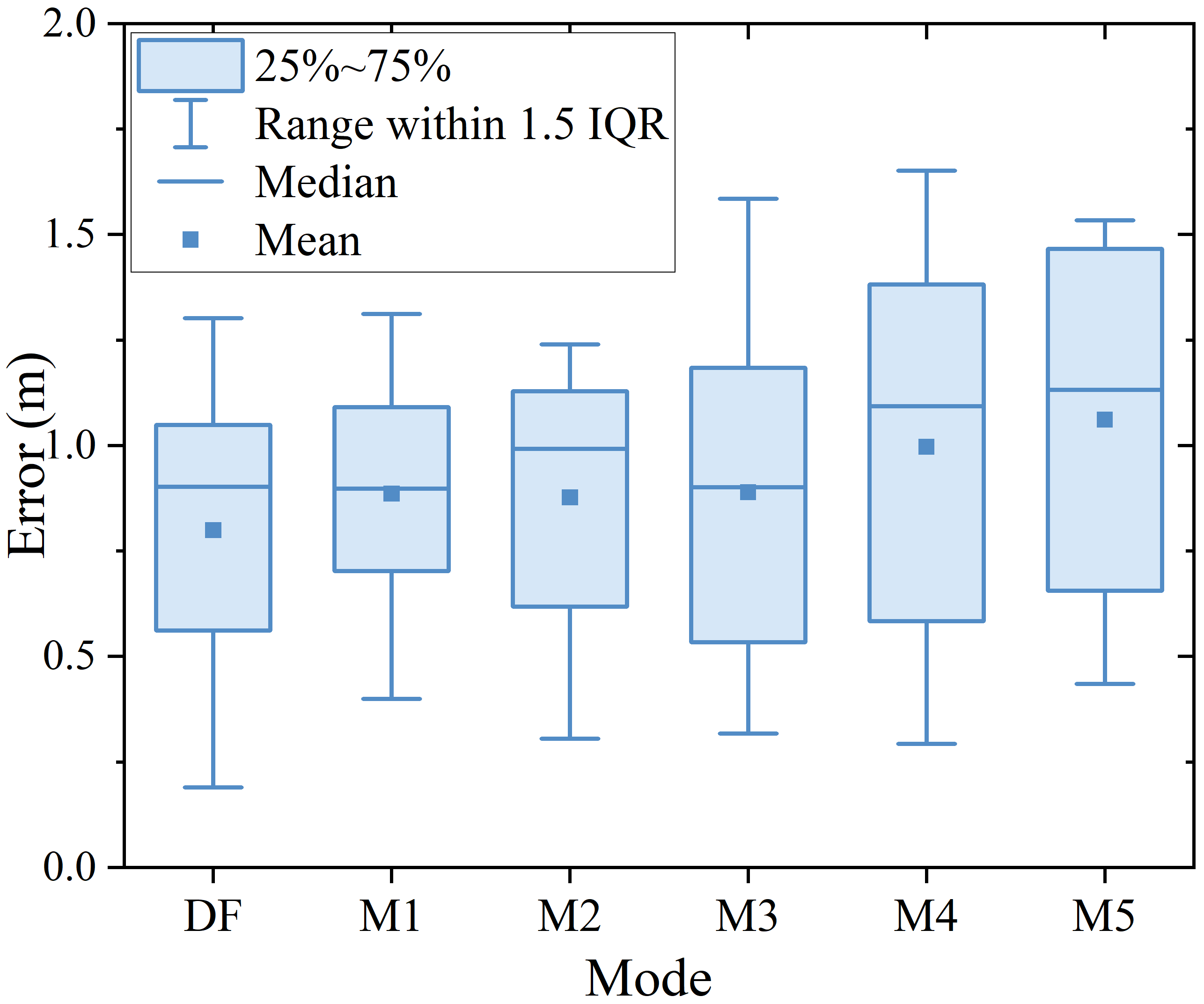}%
		\label{figBox_Plot_1_case}}
	\hfil 
	\subfigure[]{\includegraphics[width=1.55in]{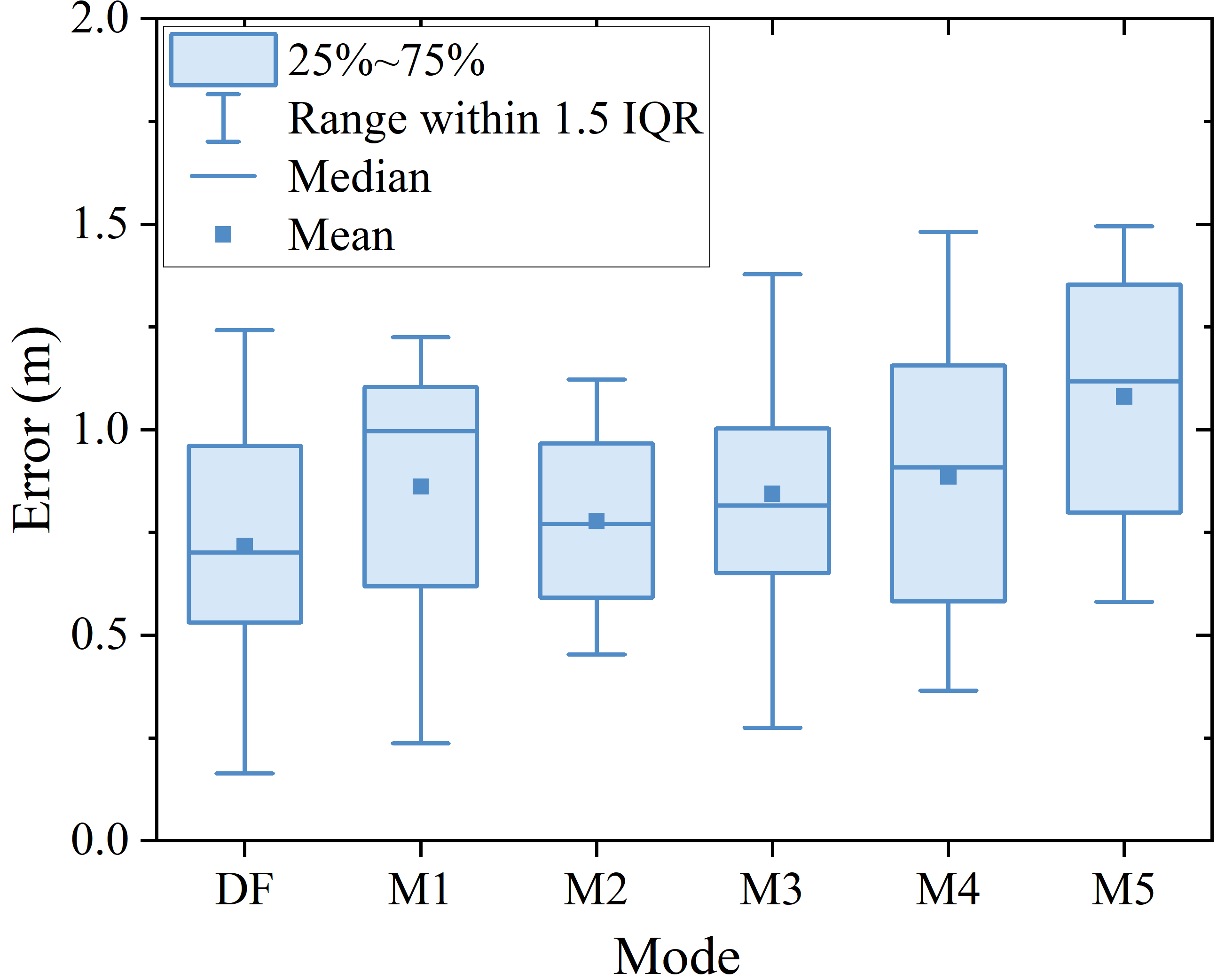}%
		\label{fig_Box_Plot_2_case}}
	\hfil \\
	\subfigure[]{\includegraphics[width=1.5in]{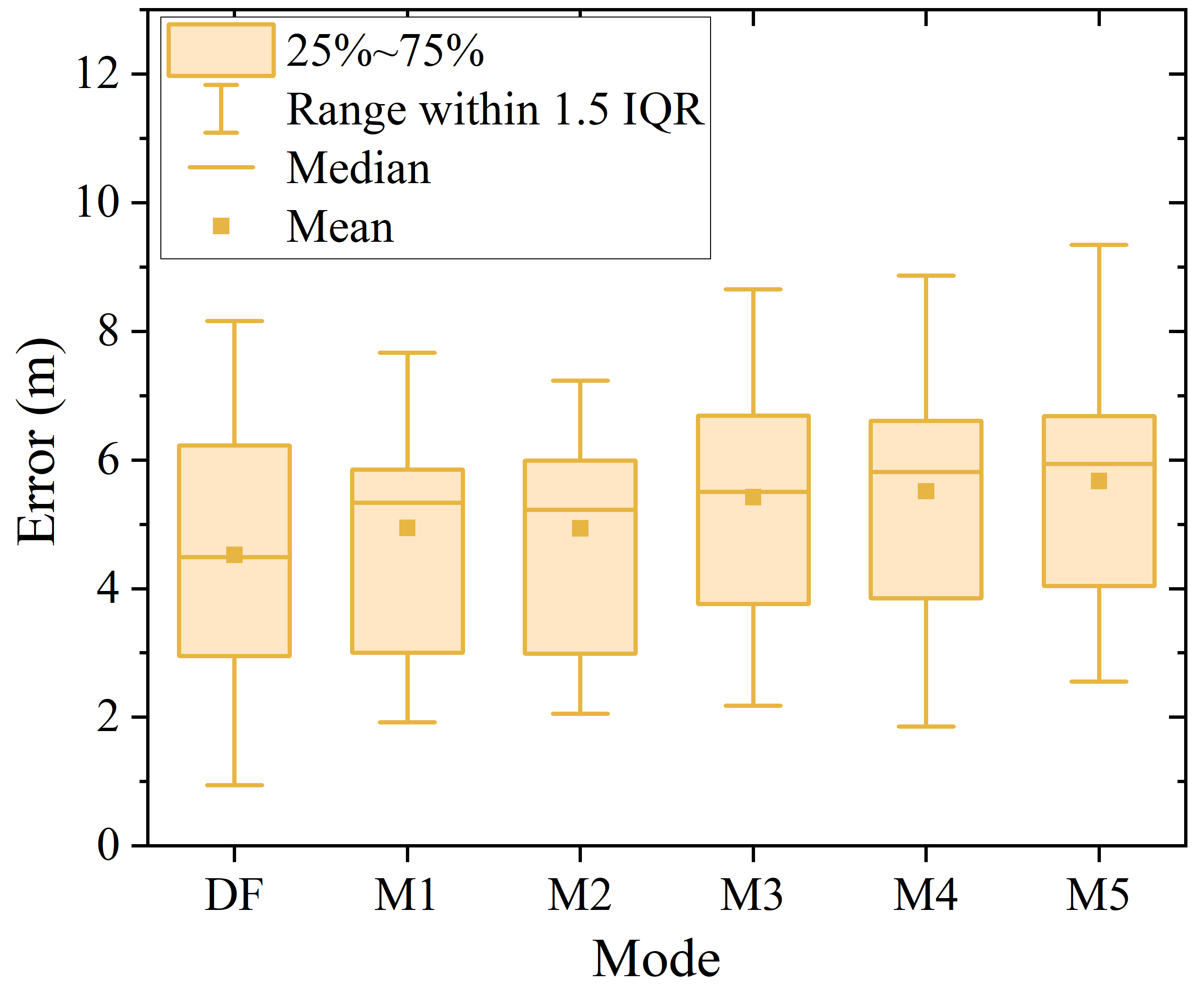}%
		\label{figBox_Plot_3_case}}
	\hfil 
	\subfigure[]{\includegraphics[width=1.5in]{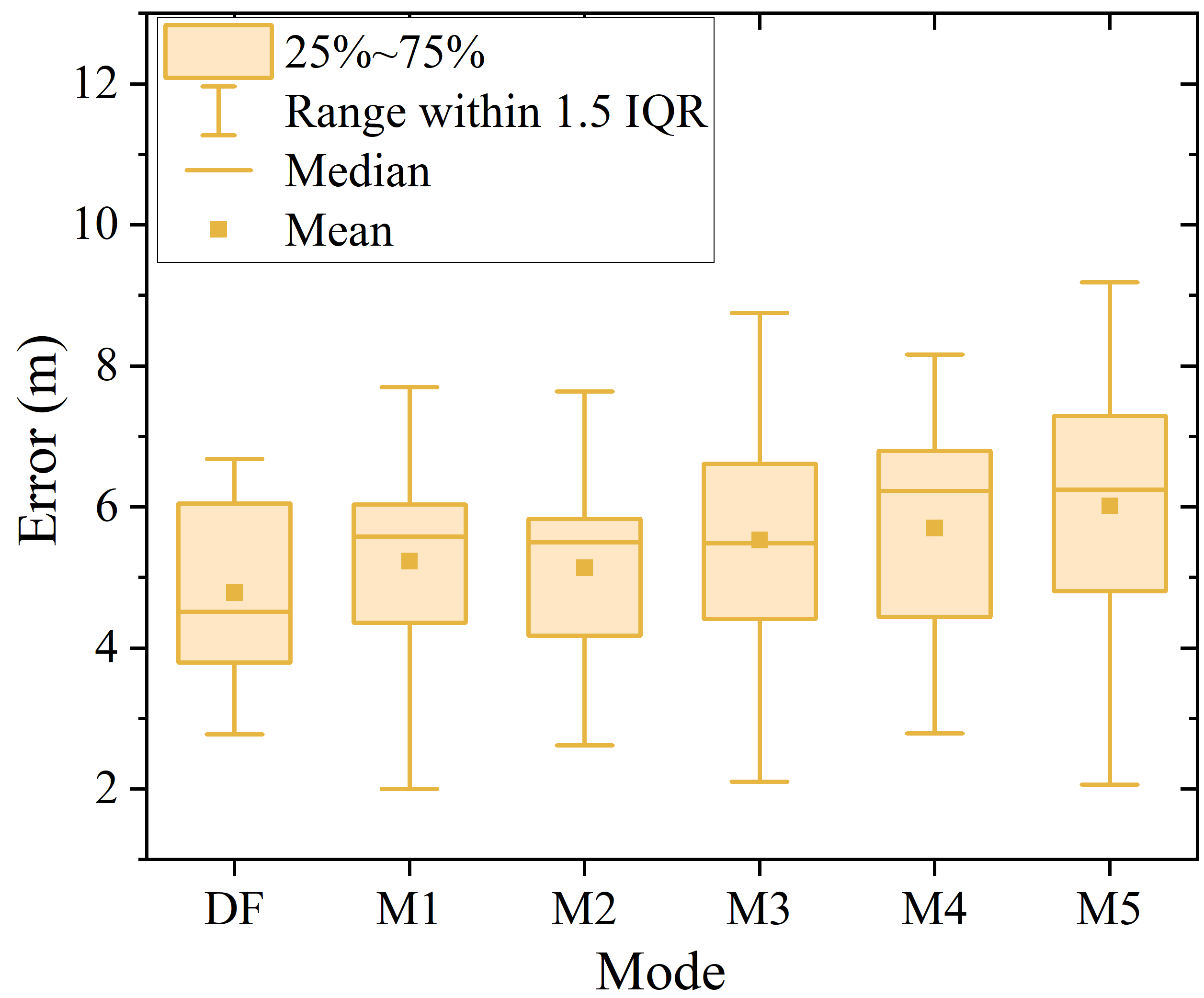}%
		\label{fig_Box_Plot_4_case}}
	
	\caption{Comparison of localization accuracy across the six modes  for different test cases ($ K = 15, \gamma = 0.4 $). (a) OST case. (b) ODT case. (c) CST case. (d) CDT case.}
	\label{Box_Plot}
\end{figure}

\subsubsection{Performance of the Attention and MUDA}To investigate the impact of QC, the attention mechanism, and MUDA on localization performance, we define six comparison modes, as detailed in Table \ref{tab1tab4}. Based on these modes, we conduct performance comparisons across four test cases: OST, ODT, CST, and CDT. The results are presented in Figure \ref{Box_Plot}.

Figures \ref{figBox_Plot_1_case} and \ref{figBox_Plot_3_case} present the localization results for the six modes in the OST and CDT test cases, respectively.  The DF mode achieves the smallest median, quartile, and average error in both cases, outperforming the other modes. (The difference in quartile between DF and M3 is negligible.) Similarly, Figures \ref{fig_Box_Plot_2_case} and \ref{fig_Box_Plot_4_case} compare the localization results in the ODT and CST test cases.  Again, the DF mode exhibits the smallest median, quartile, and average error. These results demonstrate the effectiveness of the QC module, attention mechanism, and MUDA in these test cases. A more detailed comparison and analysis follows.

The DF mode exhibits a greater performance advantage in the "different" test cases (e.g., ODT and CDT) compared to the "same" test cases (e.g., OST and CST). This is primarily attributed to the enhanced multi-source knowledge transfer and improved data utilization efficiency in the "different" test cases, further demonstrating the feasibility of DF-Loc for localization in dynamic environments.

Comparing modes M1, M2, M3, and M4, the impact of QC on localization results is greater than that of the attention mechanism across all test cases. This is primarily because QC enhances fingerprint stability and reduces interference during feature learning, thereby improving the regression characteristics of certain locations. The attention mechanism has a greater impact on localization results in the "different" test cases compared to the "same" test cases. This indicates that DF-Loc can extract transferable features from all domains, demonstrating the importance of the attention mechanism in assigning different weights to fingerprint features. Similarly, MUDA has a greater impact on localization results in the "different" test cases compared to the "same" test cases, demonstrating that MUDA-based knowledge transfer can enhance the usability and generalizability of DF-Loc in dynamic localization scenarios.

Overall, Figure \ref{Box_Plot} illustrates that the improvement in localization performance is a result of the combined effects of QC, the attention mechanism, and MUDA.

%
%
%

\begin{figure}[!t]
	\centering
	
	\subfigure[]{\includegraphics[width=1.5in]{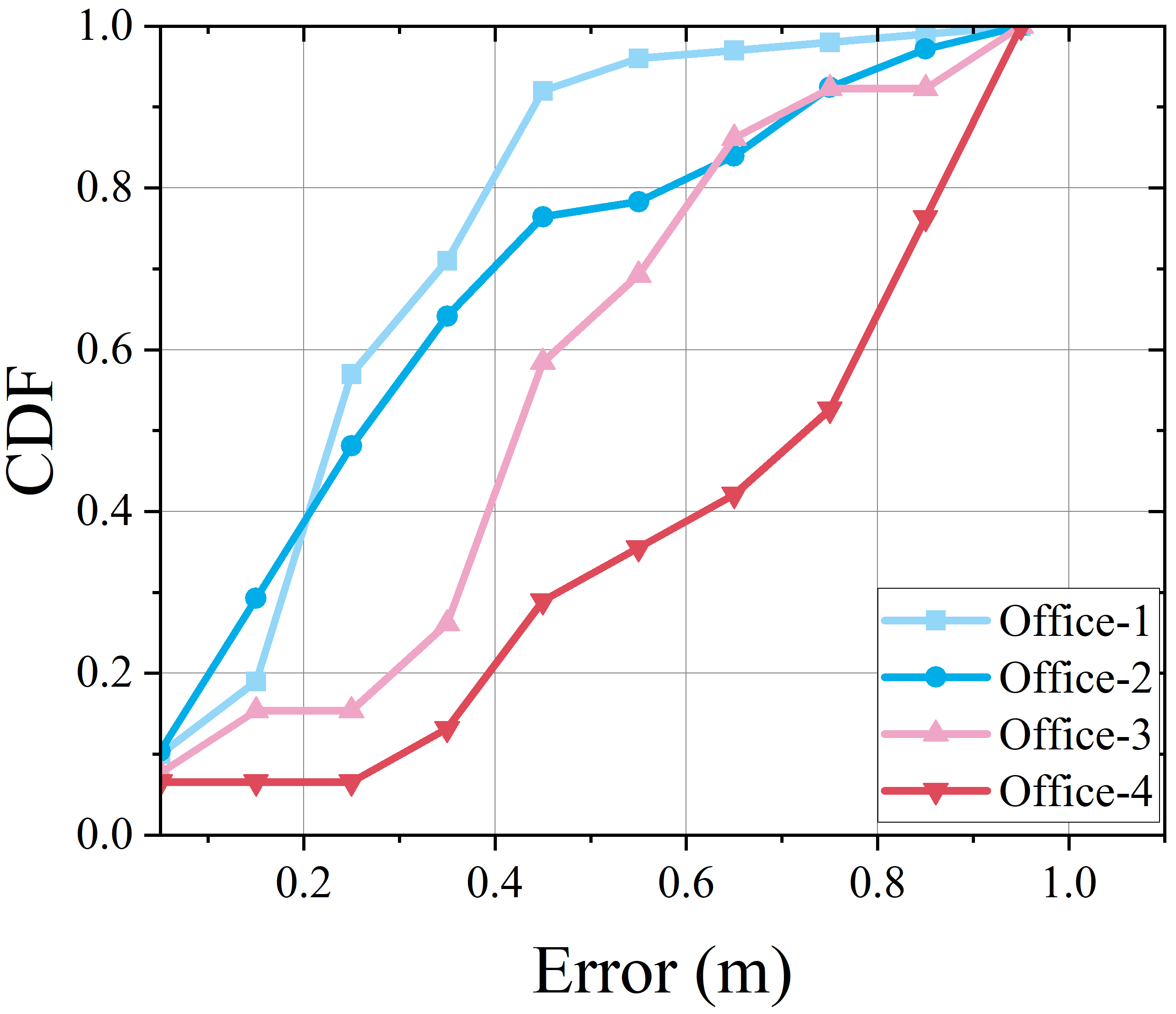}%
		\label{fig_cdf_Plot_1_case}}
	\hfil 
	\subfigure[]{\includegraphics[width=1.54in]{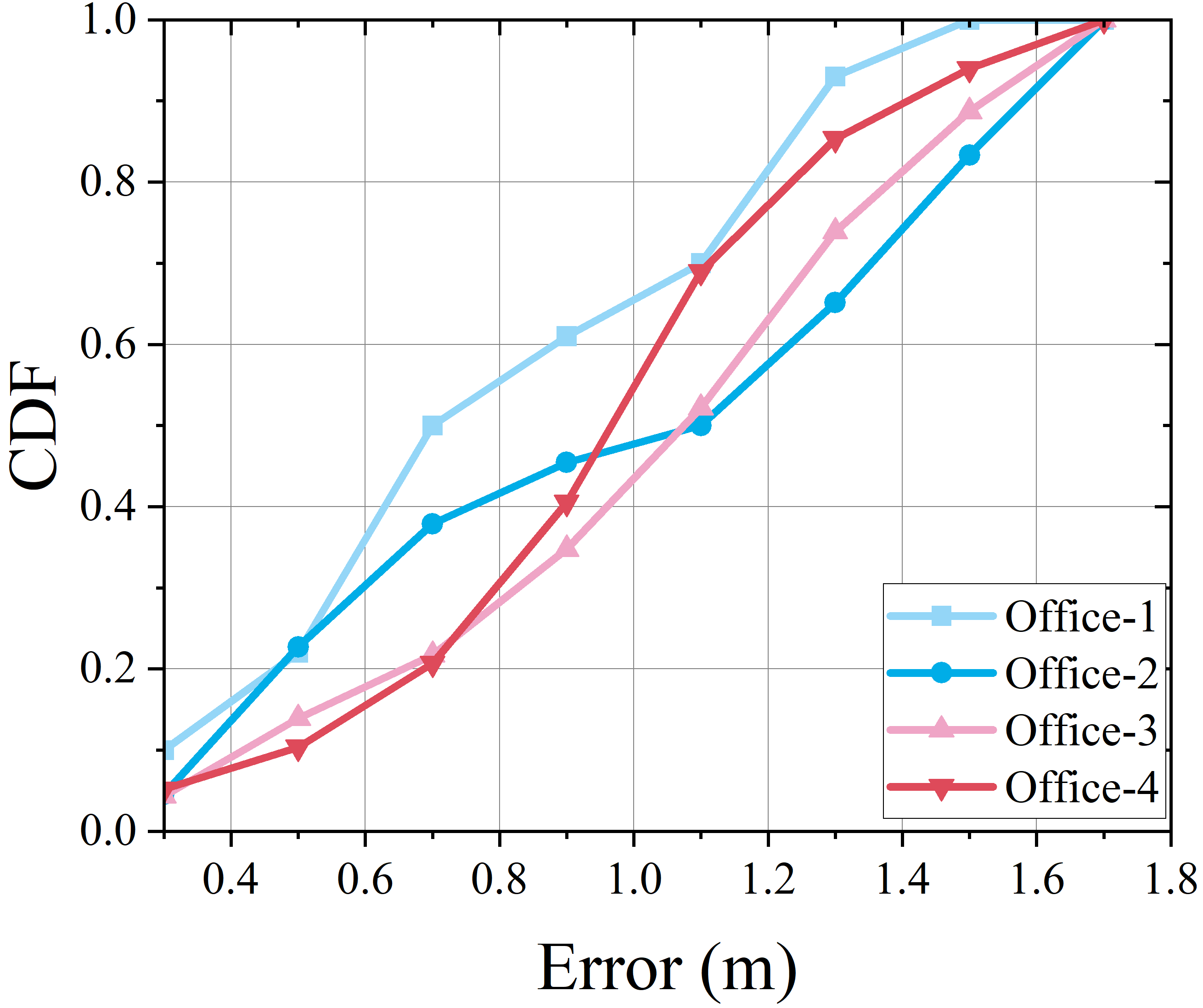}%
		\label{fig_cdf_Plot_2_case}}
	\hfil 
	
	\caption{Comparison of localization performance in the different area. (a) OWT-1 case. (b) OWT-2 case.}
	\label{fig_Box_Plot}
\end{figure}

\subsection{Robustness of DF-Loc} \label{Robustness}
To evaluate the robustness of DF-Loc, we conducted localization tests in four different office areas under test cases OWT-1 and OWT-2. As depicted in Figure \ref{fig_cdf_Plot_1_case}, Office-1 exhibits the highest localization accuracy, achieving a high localization success rate within a small error range. Conversely, Office-4 shows the lowest accuracy, with a relatively flat CDF curve, indicating a lower success rate even at larger error ranges. This suggests that this environment poses significant challenges to the localization system, potentially due to factors like pedestrian movement, environmental layout, and signal interference. The accuracy of Office-2 and Office-3 lies between these two extremes and is relatively similar, indicating comparable challenges in these environments.Compared to Figure \ref{fig_cdf_Plot_1_case}, the localization accuracy in all environments decreases in Figure \ref{fig_cdf_Plot_2_case}, as evidenced by a rightward shift in the CDF curves, indicating larger localization errors. This highlights that the performance of the localization system is affected when the feature distributions of the test and training data differ. Notably, Office-1 maintains the highest accuracy, while Office-4 remains the most challenging environment.

\begin{figure}[!t]
	\centering
	\subfigure[]{\includegraphics[width=1.52in]{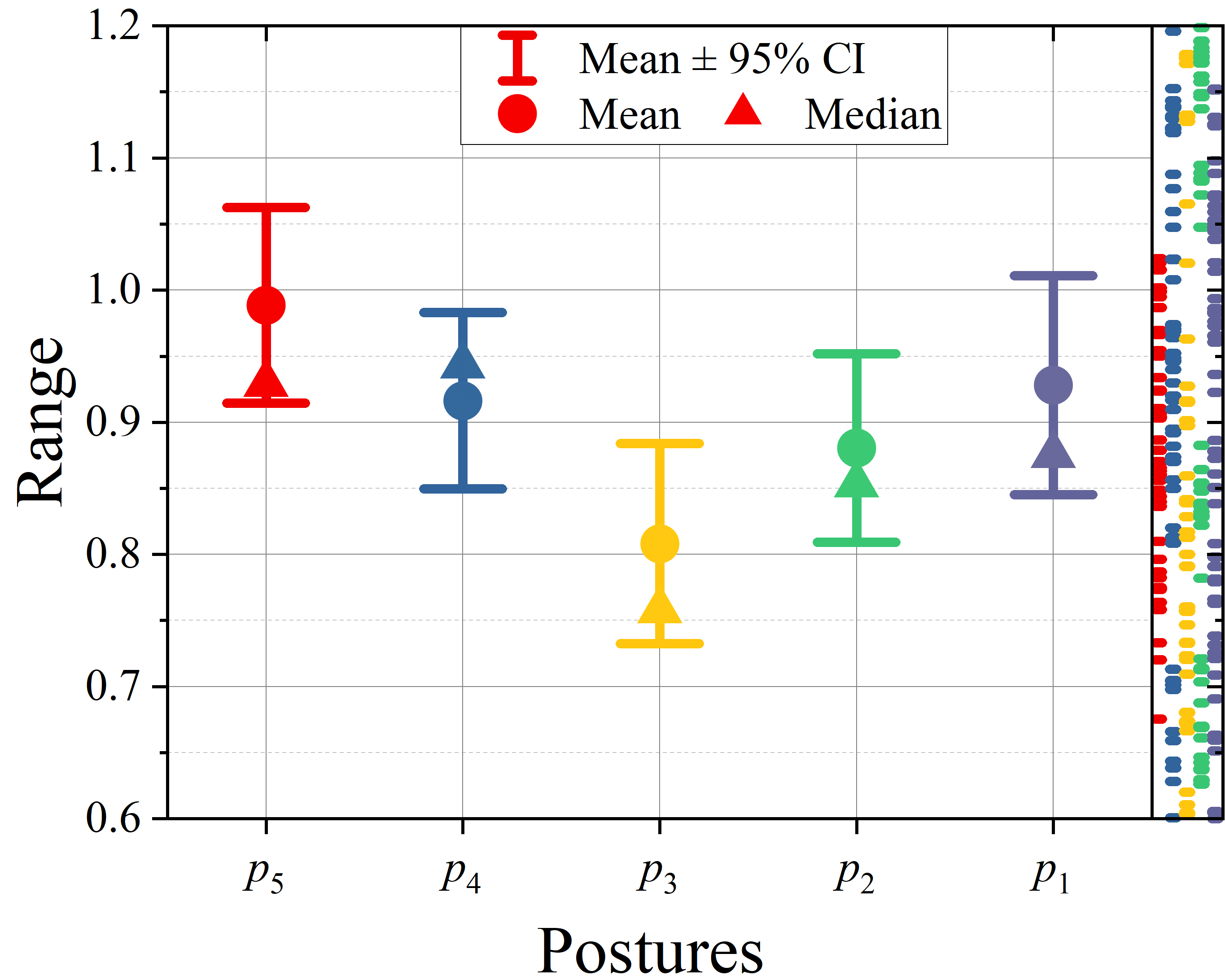}%
		\label{fig_Robustness_1_case}}
	\hfil 
	\subfigure[]{\includegraphics[width=1.52in]{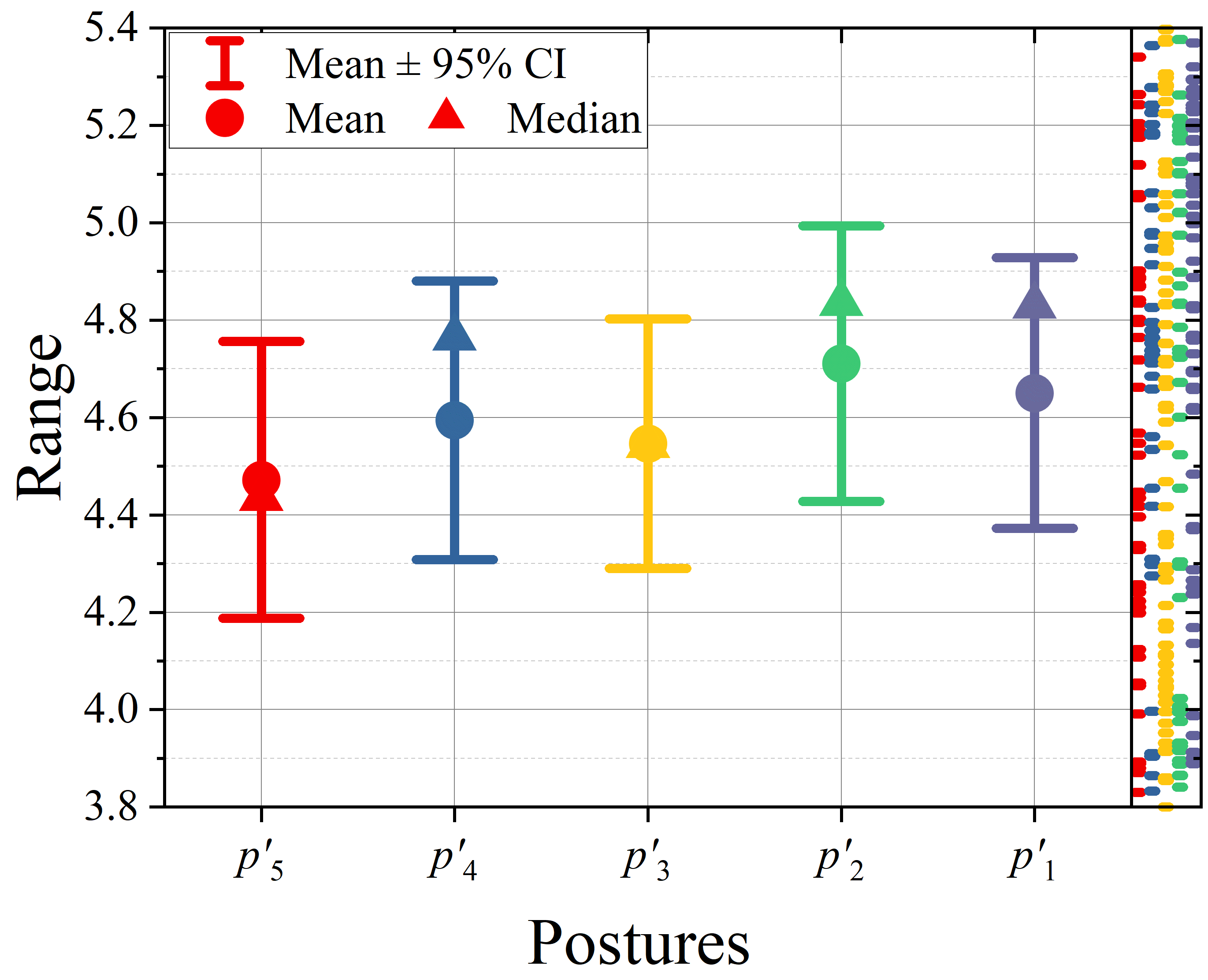}%
		\label{fig_Robustness_2_case}}
	\caption{Comparison of localization performance under different postures. (a) OWT-2 case.(b) CWT-2 case.}
	\label{fig_Robustness}
\end{figure}

Additionally, we investigate the impact of different body postures on localization performance under test cases OWT-2 and CWT-2, as illustrated in Figure \ref{fig_Robustness}.  In the office scenario (Figure \ref{fig_Robustness_1_case}), posture $ p_5 $ exhibits the largest range of localization errors, with the highest mean and median values, indicating poorer stability. Conversely, $ p_3 $ shows the smallest error range and the lowest mean and median values, suggesting optimal accuracy and stability. The remaining postures ($ p_1 $, $ p_2 $, and $ p_4 $) exhibit similar error ranges.In the classroom scenario (Figure \ref{fig_Robustness_1_case}), posture $ p'_2 $ demonstrates the lowest accuracy and stability, while $ p'_5 $ shows the highest. The other postures ($ p'_1 $, $ p'_3 $, and $ p'_4 $) have similar error ranges. This variation across postures suggests that body postures differentially affect signal blockage and reflection, leading to variations in the quality of the received signal. Postures with more limb movements introduce additional errors. Notably, the overall localization performance in the classroom scenario is lower than in the office scenario across all postures. This is primarily attributed to the sparser grid in the classroom, which reduces fingerprint correlation between locations and increases mismatches in dynamic environments. Figures \ref{fig_Box_Plot} and \ref{fig_Robustness} demonstrate that DF-Loc maintains good performance across various environments and conditions, even in the most challenging scenarios, highlighting its robustness.


\subsection{Generalization of DF-Loc} \label{Generalization}
\begin{figure}[!t]
	\centering
	\subfigure[]{\includegraphics[width=1.8in]{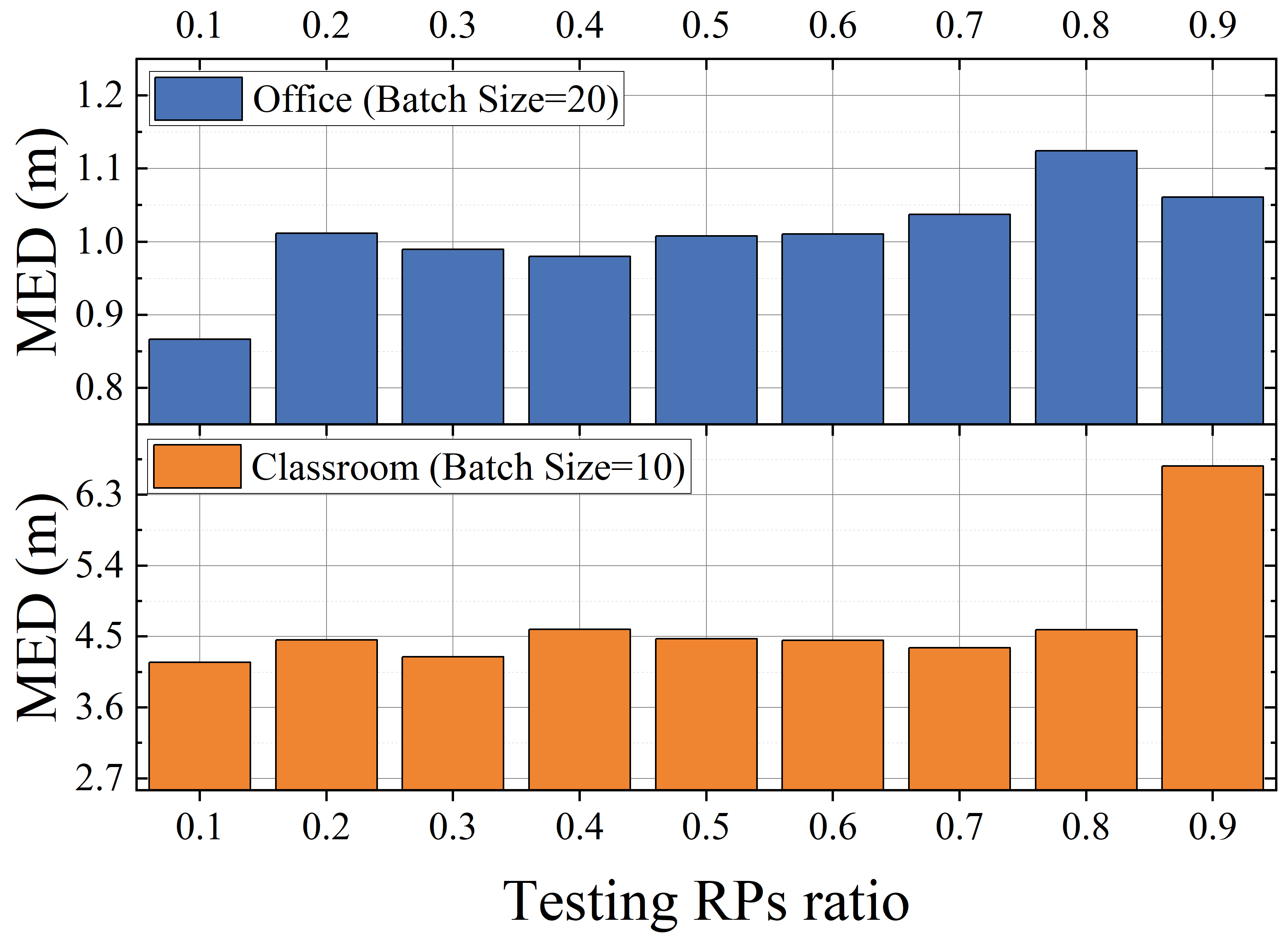}%
		\label{fig_generalization_2_case}}
	\hfil 
	\subfigure[]{\includegraphics[width=1.6in]{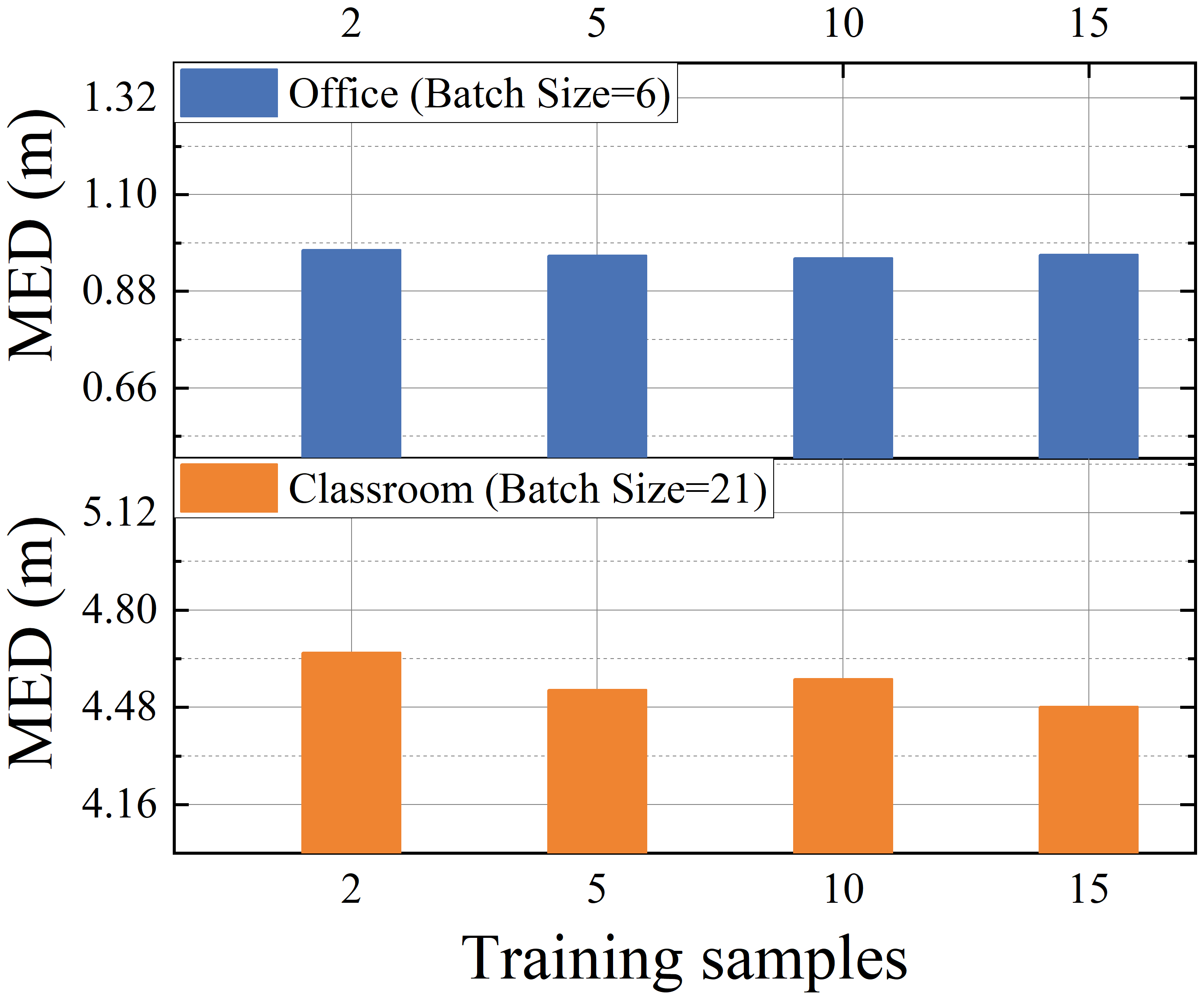}%
		\label{fig_generalization_3_case}}
	
	\caption{Generalization of DF-Loc in positioning MED. (a) Different testing RPs ratios $ \epsilon  $. (b) Different training sample size $ K $}
	\label{fig_generalization_DF-loc}
\end{figure}

To validate the generalizability of DF-Loc, we conducted two sets of experiments in two different scenarios (office and classroom). As illustrated in Figure \ref{fig_generalization_DF-loc}, the first set investigates the impact of varying testing RP ratios on localization accuracy, while the second focuses on the impact of different training sample sizes. Figure \ref{fig_generalization_2_case} shows that the MED increases with increasing $ \epsilon  $ in both scenarios, reaching a minimum at $ \epsilon = 0.1  $. This indicates that lower testing RP ratios yield higher localization accuracy. Notably, the MED values exhibit relatively small fluctuations in the middle portion of the bar chart in both scenarios, suggesting that DF-Loc is robust to variations in the testing RP ratio and maintains acceptable performance even at larger ratios, demonstrating good generalizability. Figure \ref{fig_generalization_3_case} shows that the MED values decrease slightly with increasing $ K $ in both scenarios, with relatively small overall changes. This indicates that DF-Loc can achieve good localization performance with limited training samples, significantly improving data utilization efficiency and further demonstrating its strong generalization ability.

\begin{figure}[!t]
	\centering
	\subfigure[]{\includegraphics[width=1.7in]{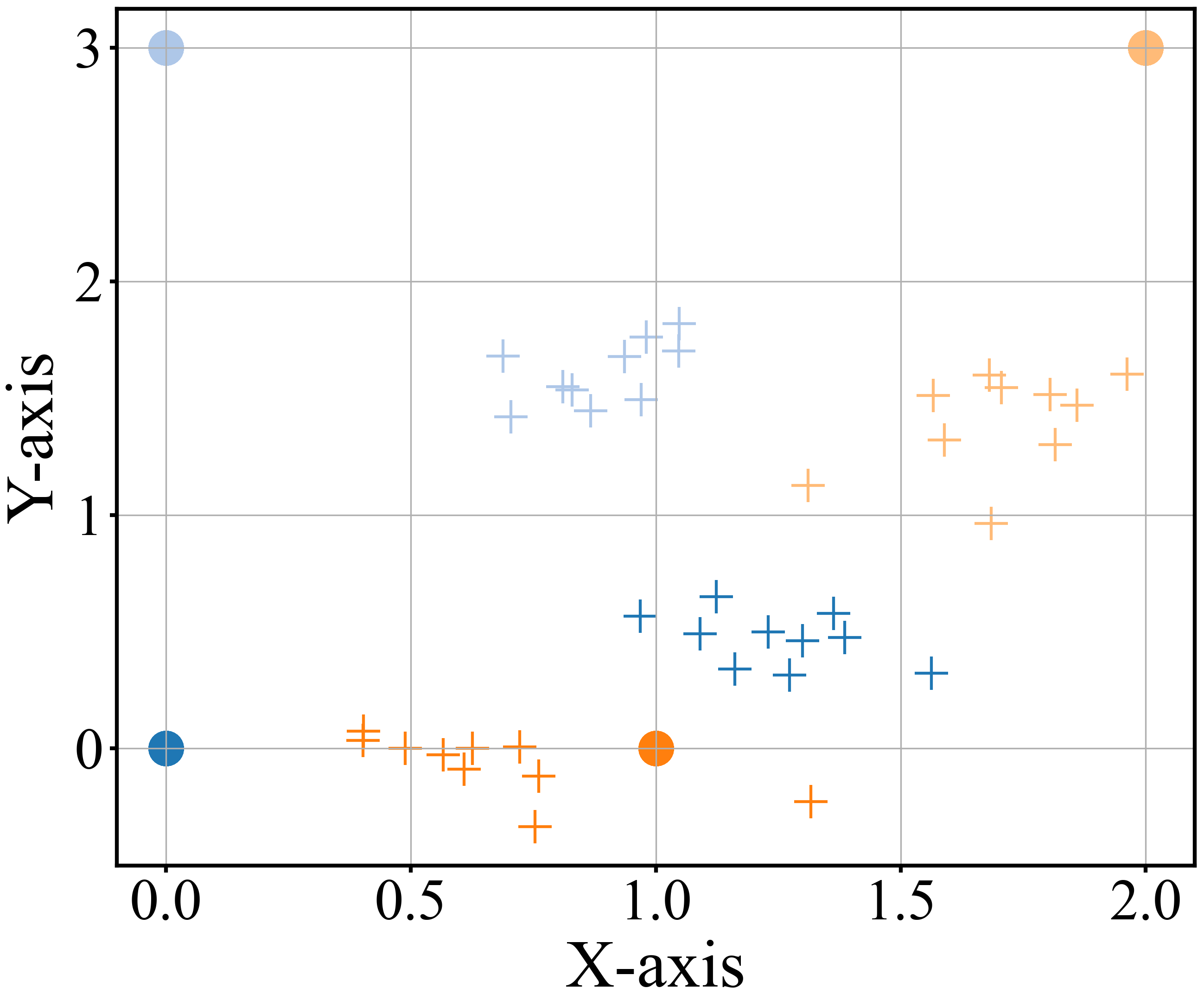}%
		\label{fig_Challenge_1_case}}
	\hfil 
	\subfigure[]{\includegraphics[width=1.7in]{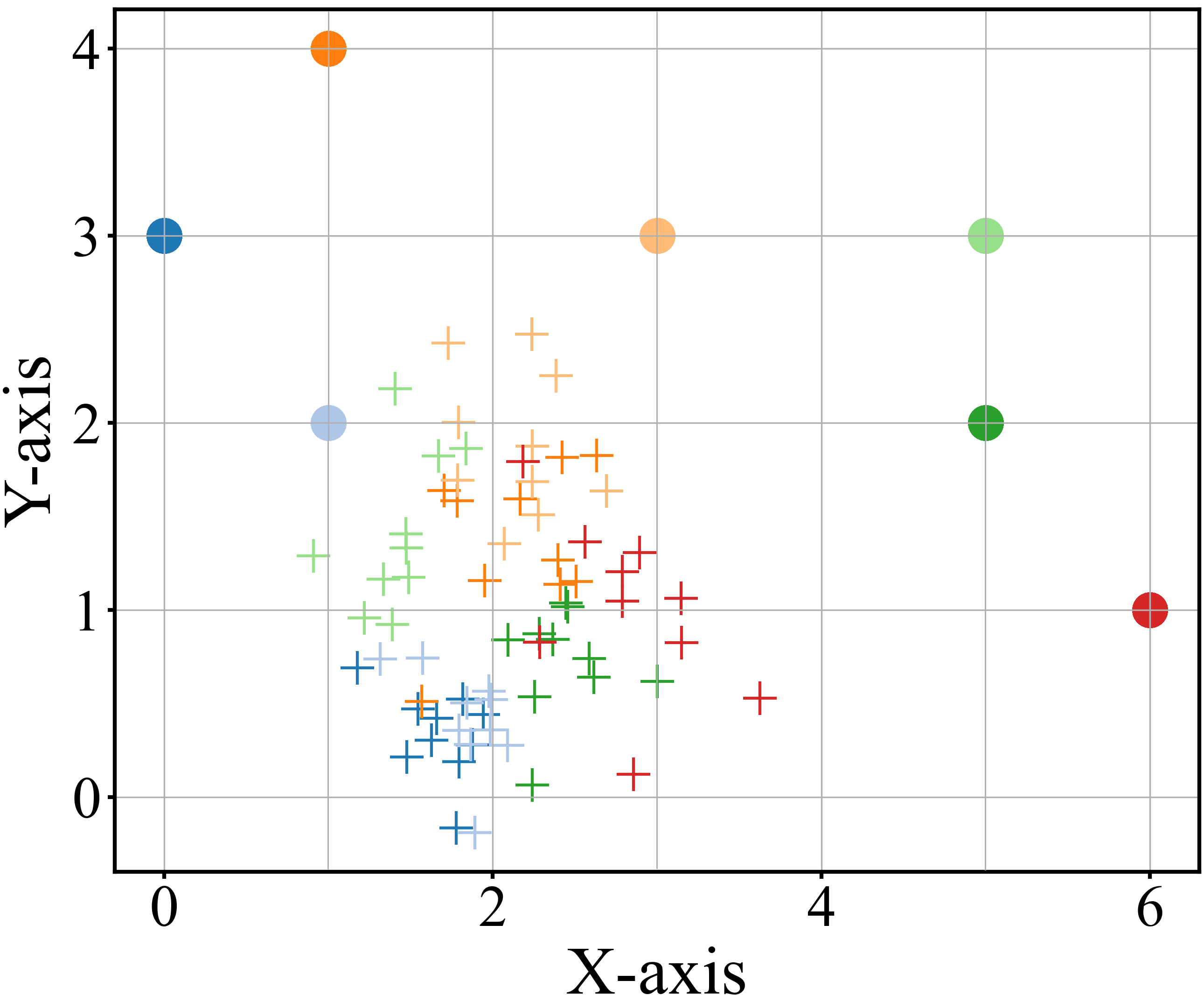}%
		\label{fig_Challenge_2_case}}
	
	\caption{Challenge of DF-Loc in positioning MAE. (a) Office-4 with $ p_5 $. (b) Classroom with $ p'_2 $. }
	\label{fig_Challenge_DF-loc}
\end{figure}

\subsection{Challenge for DF-Loc}
Sections \ref{Experimental Results}, \ref{Robustness}, and \ref{Generalization} have established the superiority of DF-Loc compared to other algorithms, its robustness to different locations and conditions, and its strong generalization performance on the data. However, DF-Loc still faces challenges related to the regression characteristics between features at different locations due to environmental layouts and human behavior. To illustrate these challenges, we conducted tests in two environments, office-4 and the classroom, using data from postures $ p_5 $ and $ p'_2 $, respectively. The prediction results are visualized as scatter plots in Figures \ref{fig_Challenge_1_case} and \ref{fig_Challenge_2_case}.

As shown in Figure \ref{fig_Challenge_1_case}, the predicted locations for coordinates (0, 3) and (0, 0) deviate significantly from the actual coordinates. This is primarily attributed to the presence of numerous obstacles in the office-4 environment, leading to NLOS signal propagation and reduced fingerprint stability.  Figure \ref{fig_Challenge_2_case} shows that the predicted locations for coordinates (0, 3), (5, 2), (5, 3), and (6, 1) also deviate significantly from the actual coordinates. Although the predicted results for the same location are relatively concentrated, their distributions overlap with those of other locations. This is mainly because human movement and occlusion introduce complexities in the received signal distribution, making it difficult for DF-Loc to extract domain-invariant feature representations. Additionally, the current feature extraction method may not effectively capture subtle differences between locations, resulting in high similarity between feature vectors of different locations, thereby affecting localization accuracy.

To address these challenges, our future work will focus on the following aspects: 1) Integrating environmental layout information into the localization algorithm, such as by constructing environment maps and utilizing semantic information, to assist localization and improve accuracy. 2) Investigating more robust localization algorithms, such as multi-station localization and multi-sensor fusion, to mitigate the impact of NLOS environments and human behavior and improve localization accuracy and stability.


\section{CONCLUSION AND FUTURE WORK}
This paper presents DF-Loc, an innovative indoor localization system based on MUDA, specifically designed for dynamic environments. DF-Loc integrates a HWF module and a LC module for CSI preprocessing, a multi-scale attention-based feature fusion network for enhanced feature extraction, and a dual-stage alignment model to align the distributions of multiple source-target domain pairs. Extensive experiments conducted in typical indoor environments demonstrate that DF-Loc achieves high localization accuracy and robustness across diverse conditions, including varying environments and human postures. Future research will focus on incorporating environmental information and exploring more robust localization algorithms to further improve performance in complex indoor scenarios.



\bibliographystyle{IEEEtran}
\bibliography{Ref.bib}
%
%
%
%

\newpage

%
%


\vspace{11pt}

\vfill

\end{document}